%% file: main.tex
\pdfoutput=1
\documentclass{article}

\PassOptionsToPackage{numbers, compress}{natbib}
\usepackage[preprint]{neurips_2022}

\usepackage[utf8]{inputenc} % allow utf-8 input
\usepackage[T1]{fontenc}    % use 8-bit T1 fonts
\usepackage{hyperref}       % hyperlinks
\usepackage{url}            % simple URL typesetting
\usepackage{booktabs}       % professional-quality tables
\usepackage{amsfonts,amsmath}       % blackboard math symbols
\usepackage{mathtools}
\usepackage{nicefrac}       % compact symbols for 1/2, etc.
\usepackage{microtype}      % microtypography
\usepackage[dvipsnames]{xcolor}  % more colors
\usepackage{xcolor}         % colors
\usepackage{siunitx}        % For nicer numbers e.g. \num{1e-10}
\usepackage{pifont}         % For some dingbats etc
\usepackage{graphicx}       % Figures
\usepackage{caption}        % For...
\usepackage{subcaption}     % ...subfigures
\usepackage{float}          % for H placement.
\usepackage{enumitem}
\usepackage{wrapfig}        % for wrapped figures
\usepackage{adjustbox}
\usepackage{changepage}
\usepackage{multirow}       % Multirow tables
\usepackage{longtable}      % Silly long tables

\newcolumntype{P}[1]{>{\centering\arraybackslash}p{#1}}

\newlength{\offsetpage}
\newenvironment{widepage}{\begin{adjustwidth}{-\offsetpage}{-\offsetpage}%
    \addtolength{\textwidth}{2\offsetpage}}%
{\end{adjustwidth}}

\DeclareCaptionFormat{smallcaption}
{%
    \small{\textbf{#1#2}}\textit{\small #3}
}
\newcommand{\cmark}{\ding{51}}%
\newcommand{\xmark}{\ding{55}}%

\usepackage{selectp} % For outputting a subset of pages
\definecolor{darkgray}{rgb}{0.25, 0.25, 0.25}
\newcommand{\ci}[2]{\textcolor{darkgray}{$\scriptstyle{_{#1}^{#2}}$}}

\newcommand{\mmoe}{\texttt{LIMoE}}

\newcommand{\rocketacc}{84.1}%

\usepackage[english]{babel} % bibliography

\def\mA{{\mathbf{A}}}
\def\mX{{\mathbf{X}}}

\def\mG{{\mathbf{G}}}

\def\mW{{\mathbf{W}}}
\def\mG{{\mathbf{G}}}
\def\mI{{\mathbf{I}}}
\def \calZ{{\mathcal{Z}}}
\def \calN{{\mathcal{N}}}
\def\vx{{\mathbf{x}}}
\def\vi{{\mathbf{i}}}
\def\vt{{\mathbf{t}}}
\def\vz{{\mathbf{z}}}

\def\vzero{{\mathbf{0}}}

\newcommand{\R}{\mathbb{R}}
\title{Multimodal Contrastive Learning with LIMoE: \\ the Language-Image Mixture of Experts}

\author{%
  Basil Mustafa\thanks{Authors contributed equally.} , Carlos Riquelme\textsuperscript{*}, Joan Puigcerver\textsuperscript{*}, Rodolphe Jenatton, Neil Houlsby \\
  Google Brain \\
  \texttt{\{basilm, rikel, jpuigcerver, rjenatton, neilhoulsby\}@google.com}
}

\begin{document}

\maketitle

\input{sections/abstract}
\input{sections/introduction}
\input{sections/method}
\input{sections/main_experimental}
\input{sections/ablations}
\input{sections/analysis}
\input{sections/literature_review}
\input{sections/conclusions}

\section{Acknowledgements}
We first thank Andreas Steiner, Xiao Wang and Xiaohua Zhai, who led early explorations into dense single-tower models for contrastive multimodal learning, and also were instrumental in providing data access. We also thank Andreas Steiner, and Douglas Eck, for early feedback on the paper. We thank Andr\'e Susano Pinto, Maxim Neumann, Barret Zoph, Liam Fedus, Wei Han and Josip Djolonga for useful discussions, and Erica Moreira and Victor Gomes for help scaling up to \mmoe{}-H/14.
\newpage

{
\small  % Allowed in neurips style guide.
\bibliographystyle{unsrt}
\bibliography{main}
}

% \newpage

%%%%%%%%%%%%%%%%%%%%%%%%%%%%%%%%%%%%%%%%%%%%%%%%%%%%%%%%%%%%
% \input{sections/checklist}
%%%%%%%%%%%%%%%%%%%%%%%%%%%%%%%%%%%%%%%%%%%%%%%%%%%%%%%%%%%%

\newpage
\appendix

\input{appendix/1_training_setup} \clearpage
\input{appendix/2_auxiliary_losses} \clearpage
\input{appendix/3_tabular_results} \clearpage
\input{appendix/4_practical_aspects} \clearpage
\input{appendix/5_experiment_overflow} \clearpage
\input{appendix/6_analysis} \clearpage
\input{appendix/7_rocket}

\end{document}

%% file: sections/abstract.tex
\begin{abstract}
\looseness=-1
Large sparsely-activated models have obtained excellent performance in multiple domains.
However, such models are typically trained on a single modality at a time.
We present the Language-Image MoE, \mmoe{}, a sparse mixture of experts model capable of multimodal learning.
\mmoe{} accepts both images and text simultaneously, while being trained using a contrastive loss.
MoEs are a natural fit for a multimodal backbone, since expert layers can learn an appropriate partitioning of modalities.
However, new challenges arise; in particular, training stability and balanced expert utilization, for which we propose an entropy-based regularization scheme.
Across multiple scales, we demonstrate remarkable performance improvement over dense models of equivalent computational cost.
{\mmoe{}-L/16} trained comparably to {CLIP-L/14} achieves 78.6\% zero-shot ImageNet accuracy (vs. 76.2\%), and when further scaled to H/14 (with additional data) it achieves \rocketacc\%, comparable to state-of-the-art methods which use larger custom per-modality backbones and pre-training schemes.
We analyse the quantitative and qualitative behavior of \mmoe{}, and demonstrate phenomena such as differing treatment of the modalities and the organic emergence of modality-specific experts.
\end{abstract}

%% file: sections/introduction.tex
\section{Introduction}
\label{sec:intro}
Sparsely activated mixture of expert (MoE) models have recently been used with great effect to scale up both vision~\cite{riquelme2021vmoe,lou2022mixermoe} and text models~\cite{lepikhin2021gshard,zoph2022stmoe}. The primary motivation for using MoEs is to scale model parameters while keeping compute costs under control. These models however have other benefits; for example, the sparsity protects against catastrophic forgetting in continual learning~\cite{collier2020routing} and can improve performance for multitask learning~\cite{ma2018mmoe} by offering a convenient inductive bias.

Given success in each individual domain, and the intuition that sparse models may better handle distinct tasks, we explore the application of MoEs to multimodal modelling. 
We take the first step in this direction, and study models that process both images and text.
In particular, we train a single multimodal architecture that aligns image and text representations via contrastive learning~\cite{radford2021clip}.

When using a setup proposed in prior unimodal models~\cite{fedus2022switch,riquelme2021vmoe}, we find that feeding multiple modalities to a single architecture leads to new failure modes unique to MoEs.
To overcome these, we present a set of \textit{entropy based regularisers} which stabilise training and improve performance.
We call the resulting model \mmoe{} (Language-Image MoE).

We train a range of \mmoe{} models
%Using \mmoe{}, we train a range of large scale models 
which significantly outperform compute-matched dense baselines. 
We scale this up to a large 5.6B parameter \mmoe-H/14, which applies 675M parameters per token. When evaluated zero-shot~\cite{radford2021clip} on ImageNet-2012~\cite{deng2009imagenet} it achieves an accuracy of \rocketacc\%, competitive with two-tower models that make use of modality-specific pre-training and feature extractors, and apply 3-4x more parameters per token.

% Sorry, wrapfig is annoying, and this will need to be tuned as we move text around.
\begin{wrapfigure}[17]{r}{0.37\textwidth}
  \begin{center}
  \vspace{-0.9cm}
  \includegraphics[scale=0.88, bb = 0.2in 0 2.2in 3.3in]{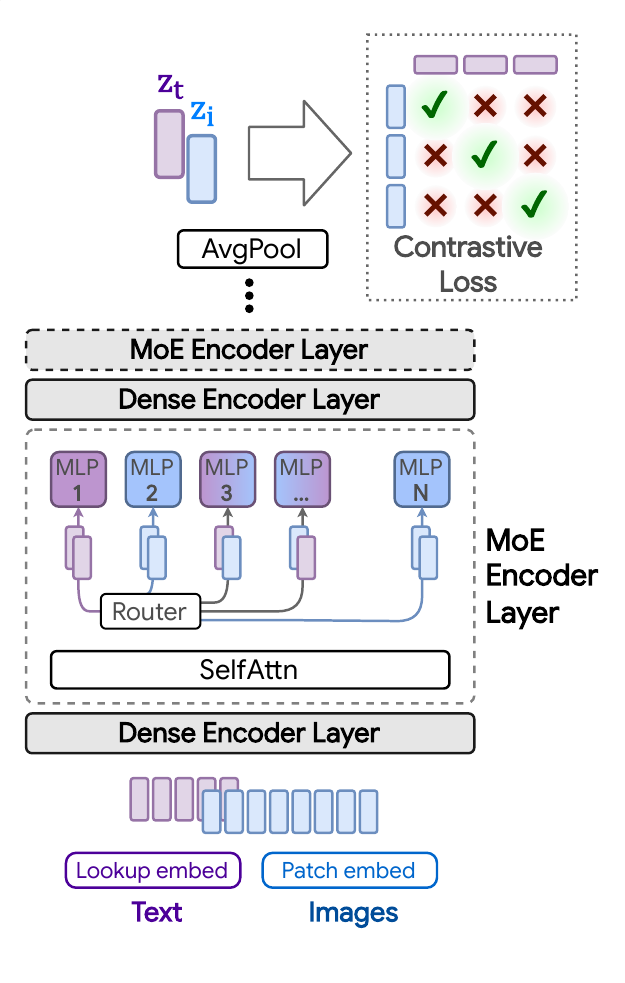}
  \vspace{-0.6cm}
  \caption{\mmoe{}, a sparsely activated multimodal model, processes both images and texts, utilising conditional computation to allocate tokens in a modality-agnostic fashion.}
  \label{fig:model_diagram}
  \end{center}
\end{wrapfigure}

In summary, our contributions are as follows.

\begin{itemize}[leftmargin=1.5em]
    \item We propose \mmoe, the first large-scale multimodal mixture of experts models.
    \item We demonstrate in detail how prior approaches to regularising mixture of experts models fall short for multimodal learning, and propose a new entropy-based regularisation scheme to stabilise training.
    \item We show that \mmoe{} generalises across architecture scales, with relative improvements in zero-shot ImageNet accuracy ranging from 7\% to 13\% over equivalent dense models. Scaled further, \mmoe-H/14 achieves \rocketacc\% zero-shot ImageNet accuracy, comparable to SOTA contrastive models with per-modality backbones and pre-training.
    \item Lastly, we present ablations and analysis to understand the model's behavior and our design decisions.
\end{itemize}

%% file: sections/method.tex
\section{Multimodal Mixture of Experts}
\label{sec:method}
Multimodal contrastive learning typically works with \emph{independent} per-modality encodings~\cite{radford2021clip,jia2021align}.
%
% That is, separate models $\vtheta_m$ are trained for each modality $m$ in order to provide a final representation for each input $\vi_m$ from that modality: $\vz_{\vi_m} := f_{\vtheta_m}(\vi_m)$. 
%
%% JOAN: The following notation is confusing, since we first say that the model is \vtheta_m, but then we use f_{\vtheta_m}.
% That is, separate models $\vtheta_m$ are trained for each modality $m$ in order to provide a final representation for each input from that modality. In the example of some image and text inputs $\vi$ and $\vt$, we have $\vz_{\vi} = g_{\vtheta_\text{image}}(\vi)$ and $\vz_{\vt} = f_{\vtheta_\text{text}}(\vt)$.
That is, separate models $f_m$ are trained to provide a final representation for every input from the corresponding modality, $m$.
In the case of some image and text inputs, $\vi$ and $\vt$, we have $\vz_{\vi} = f_{\text{image}}(\vi)$ and $\vz_{\vt} = f_{\text{text}}(\vt)$.
%
% \rj{I have the feeling that the notation for $\vi_m$ is overloaded, which could be confusing. Later, $\vi_j$ represents the $j$-th image, while here $\vi_m$ stands for a generic input with modality $m$. I would propose:
% ``
% \dots training a different model $\vtheta_m$ for each modality $m$ in order to provide a final representation $\vz$ for each input from that modality. In the example of some image and text inputs $\vi$ and $\vt$, we have $\vz_{\vi} = g_{\vtheta_\text{image}}(\vi)$ and $\vz_{\vt} = f_{\vtheta_\text{text}}(\vt)$.''}
%
For contrastive learning with images and text, this approach results in a ``two-tower''
% \rj{We should be consistent, either ``\dots'' or `\dots'; both are used now}
architecture, one for each modality. We study a one-tower setup instead, where a \emph{single} model is shared for all modalities, as shown in Figure~\ref{fig:model_diagram}.
The one-tower design offers increased generality and scalability, and the potential for cross-modal and cross-task knowledge transfer.
We next describe the \mmoe~architecture and training routine.

\subsection{Multimodal contrastive learning}
\looseness=-1
Given $n$ pairs of images and text captions $\{(\vi_j, \vt_j)\}_{j=1}^n$, the model learns representations $\calZ_n\!=~\!\!\{(\vz_{\vi_j}, \vz_{\vt_j})\}_{j=1}^n$
such that those corresponding to paired inputs are closer in feature space than those of unpaired inputs.
The contrastive training objective~\cite{radford2021clip,zhang2020convirt}, with learned temperature $T$, is:
% JOAN: Colors are not nice for colorblind folks and people printing the article.
% \begin{equation}\label{eq:contrastive_loss}
% \mathcal{L}_j(\calZ_n) =
% {\color{violet}
% -\frac{1}{2}\log\frac{e^{\langle \vz_{\vi_j}, \vz_{\vt_j} \rangle /T}}{\sum_{k=1}^n e^{\langle \vz_{\vi_k}, \vz_{\vt_j} \rangle /T}}}
% {\color{CornflowerBlue}
% -\frac{1}{2}\log\frac{e^{\langle \vz_{\vi_j}, \vz_{\vt_j} \rangle /T}}{\sum_{k=1}^n e^{\langle \vz_{\vi_j}, \vz_{\vt_k} \rangle /T}}
% }
% \end{equation}
% consisting of an {\color{violet}image-to-text loss}, a {\color{CornflowerBlue}text-to-image loss} and a learned temperature $T$.
%
\begin{equation}\label{eq:contrastive_loss}
\mathcal{L}_j(\calZ_n) =
\underbrace{%\color{CornflowerBlue}
-\frac{1}{2}\log\frac{e^{\langle \vz_{\vi_j}, \vz_{\vt_j} \rangle /T}}{\sum_{k=1}^n e^{\langle \vz_{\vi_j}, \vz_{\vt_k} \rangle /T}}
}_{\text{image-to-text loss}}
\underbrace{%\color{violet}
-\frac{1}{2}\log\frac{e^{\langle \vz_{\vi_j}, \vz_{\vt_j} \rangle /T}}{\sum_{k=1}^n e^{\langle \vz_{\vi_k}, \vz_{\vt_j} \rangle /T}}}_{\text{text-to-image loss}}.
\end{equation}
%where $T$ is a learned temperature.

\begin{figure}[t]
  \centering
  \includegraphics[width=\textwidth, bb=0 0 1769 718]{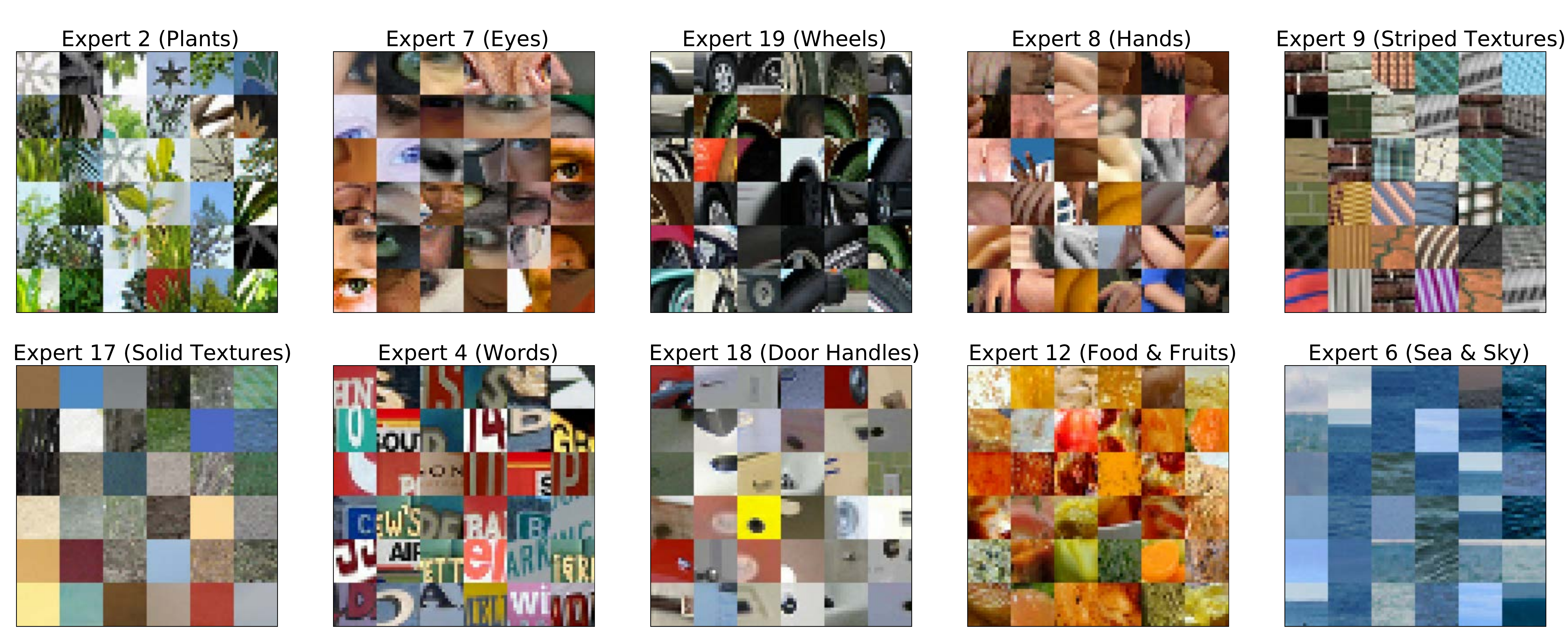}
  \caption{\textbf{Token routing examples for Coco.} Image examples of how patches are routed at the MoE layer placed in the 18-th encoder block --i.e. middle of the network-- for the \mmoe{}-H/14 model.}
  \label{fig:limoe_h14_examples_l17_main}
\end{figure}

\subsection{The \mmoe{} Architecture}
We use a single Transformer-based architecture for both image and text modalities. 
The model uses a linear layer per modality to project the intrinsic data dimension to the desired width:
for text, a standard one-hot sentencepiece encoding and learned vocabulary~\cite{kado2018spm}, and for images,  ViT-style patch-based embeddings~\cite{dosovitskiy2020vit}.
% It is otherwise modality-agnostic: % Joan: the previous is not true, since we average-pool per-modality and have final per-modality linear projection
Then all tokens are processed by a shared transformer encoder, which is not explicitly conditioned on modality.
The token representations from the final layer are average-pooled to produce a single representation vector $\vz_m$ for each modality.
To compute the training loss in~\eqref{eq:contrastive_loss},
the paired image and text representations
are then linearly projected using per-modality weight matrices~$\mW_m$'s and $\mathcal{L}_j$ is applied to $\{(\mW_\text{image} \ \vz_{\vi_k}, \mW_\text{text} \ \vz_{\vt_k})\}_{k=1}^n$.

This one-tower setup can be implemented with a standard dense Transformer (and we train many such models as baselines).
Next, we describe how we introduce MoEs to this setup for \mmoe.

\textbf{Sparse MoE backbone:}
% JOan: Removed, this suggests that sparse parts are not shared across modalities or are conditioned on modalities.
%Similarly, \mmoe{} processes a sequence of image or text tokens.
%Dense components of the model (self attention layers and some MLP layers) are still shared across modalities.
%
Sparse MoE layers are introduced following the architectural design of~\cite{riquelme2021vmoe, lepikhin2021gshard}.
The \textit{experts}---parts of the model activated in an input-dependent fashion---are  MLPs.
\mmoe{} contains multiple MoE layers. In those layers, each token $\vx \in\mathbb{R}^D$ is processed sparsely by $K$ out of $E$ available experts. To choose which $K$, a lightweight router predicts the gating weights \textit{per token}: $g(\vx) = \texttt{softmax}(\mW_g\vx) \in \R^E$ with learned $\mW_g \in \R^{D\times E}$. The outputs of the $K$ activated experts are linearly combined according to the gating weights: $\texttt{MoE}(\vx) = \sum_{e=1}^K g(\vx)_e \cdot \texttt{MLP}_e(\vx)$.

Note that, for computational efficiency and implementation constraints, experts have a \textit{fixed buffer capacity}. The number of tokens each expert can process is fixed in advance, and typically assumes that tokens are roughly balanced across experts. If capacity is exceeded, some tokens are ``dropped''; they are not processed by the expert, and the expert output is all zeros for those tokens. The rate at which tokens are successfully processed (that is, not dropped) is referred to as the ``success rate''. It is an important indicator of healthy and balanced routing and often indicative of training stability.

We discovered that routing with tokens from multiple modalities introduces new failure modes; in the next sections we demonstrate this phenomenon, and describe our techniques to address it.

\subsubsection{Challenges for multimodal MoEs}
As mentioned, experts have a fixed buffer capacity. 
Without intervention, Top-$K$ MoEs tend to ``collapse'', thus using only one expert. This causes most tokens to be dropped and leads to poor performance~\cite{shazeer2017outrageously}. Prior works therefore use auxiliary losses to encourage balanced routing~\cite{riquelme2021vmoe,lepikhin2021gshard,fedus2022switch}.

In multimodal settings, new challenges arise; one is modality misbalance. In realistic setups, there will likely be more of one data type than another.
Accordingly, we do not assume or enforce balanced data across modalities, and our experiments have $3-17\times$ more image tokens than text tokens.

Modality-specific experts tend to emerge naturally. In this imbalanced context, this leads to a scenario where all of the tokens from the minority modality get assigned to a single expert, which runs out of capacity.
On a global level, routing still appears balanced: tokens from the majority modality are nicely distributed across experts, thereby satisfying modality-agnostic auxiliary losses. For example, in our standard B/16 setup, the router can optimize the importance loss~\cite{shazeer2017outrageously} to within 0.5\% of its minimum value by perfectly balancing image tokens but dropping all text tokens. This however leads to unstable training and unperforming models.

% Second, text and images are significantly distinct modalities and we observe them to behave very differently - note that shared capacity constraints lead to batch-wise interactions between routings, and their routing distributions can possibly cause interference. \rj{I am not sure to fully understand the second part of the sentence, after ``note''. Moreover, given the way the observation is phrased, the reader feels like willing to know what happens if the capacity constraints were not shared; is that feasible? Perhaps we can discuss that option and why it was a bad/not practical idea.} \neil{I feel this needs a little more elaboration, it feels like an instance of the first problem?} % Basil: In hindsight this made no sense. We know that image & text tokens like to be routed differently, but why?

% % \carlos{can we add one plot showing this? either here or in the appendix.}

\subsubsection{Auxiliary losses}\label{sec:entropy_aux_losses}
We refer to auxiliary losses used in V-MoE~\cite{riquelme2021vmoe} as the \textit{classic} auxiliary losses. We find that they do not yield stable and performant multimodal MoE models.
Therefore, we introduce two new losses: the \textit{local entropy loss} and the \textit{global entropy loss}, which are applied on a per-modality basis.
We combine these losses with the classic losses; see Appendix~\ref{app:aux_losses} for a summary of all auxiliary losses.

\textbf{Definition.} In each MoE layer, for each modality $m$, the router computes a gating matrix $\mG_m \in \R^{n_m \times E}$. Each row of $\mG_m$ represents the probability distribution over $E$ experts for one of the $n_m$ tokens of that modality in the batch.
For a token $\vx$ that corresponding row is $p_m(\texttt{experts} | \vx) \in \R^E$; this later dictates which experts process $\vx$.
The local and global entropy losses are defined by:
\begin{equation}\label{eq:local_and_global_ent_losses}
\!\!\Omega_\text{local}(\mG_m) \!:=\!
\frac{1}{n_m} \! \sum_{i=1}^{n_m} \mathcal{H}(p_m(\texttt{experts} | \vx_i))
\ \ \text{and}\ \
% \Omega_\text{global}(\mG_m) \!=\!
% -\mathcal{H}\!\left(\!\!
% \frac{1}{n_m} \! \sum_{i=1}^{n_m} p_m(\texttt{experts} | \vx_i)
% \!\!\right)\!,
\Omega_\text{global}(\mG_m) \!:=\!
-\mathcal{H}\!\left(
 \tilde{p}_m(\texttt{experts})
\right),
\end{equation}
where $\tilde{p}_m(\texttt{experts}) = \frac{1}{n_m} \sum_{i=1}^{n_m} p_m(\texttt{experts} | \vx_i)$
is the expert probability distribution averaged over the tokens and $\mathcal{H}(p) = -\sum_{e=1}^E p_e \log(p_e)$ denotes the entropy. Note that $\tilde{p}_m(\texttt{experts}) \approx p_m(\texttt{experts})$ since we approximate the true marginal from the tokens in the batch. We use the terminology \textit{local} vs.~\textit{global} to emphasise the fact that $\Omega_\text{local}$ applies the entropy \textit{locally} for each token while $\Omega_\text{global}$ applies the entropy \textit{globally} after having marginalized out the tokens. 

% \carlos{So far here we don't say anything about the thresholded loss for global entropy? i.e. $\text{loss} = \max(H_{\text{target}} - H_{P_m}, 0)$. I see it's introduced below, whereas it's already applied in Figure~\ref{fig:entropy_analysis}.} \rj{I did it on purpose, to streamline as much as possible the presentation. I would say it is fine since when we mention the threshold, we clearly refer to Fig 2 once more (``it is evident in Figure 2b'') I would be in favor to keep it that way.}

\begin{figure}[H]
    \begin{widepage}
    \begin{subfigure}[c]{0.25\textwidth}
        \vspace{-0.25cm}
        \centering
        \includegraphics[width=\textwidth, bb=0 0 304 518]{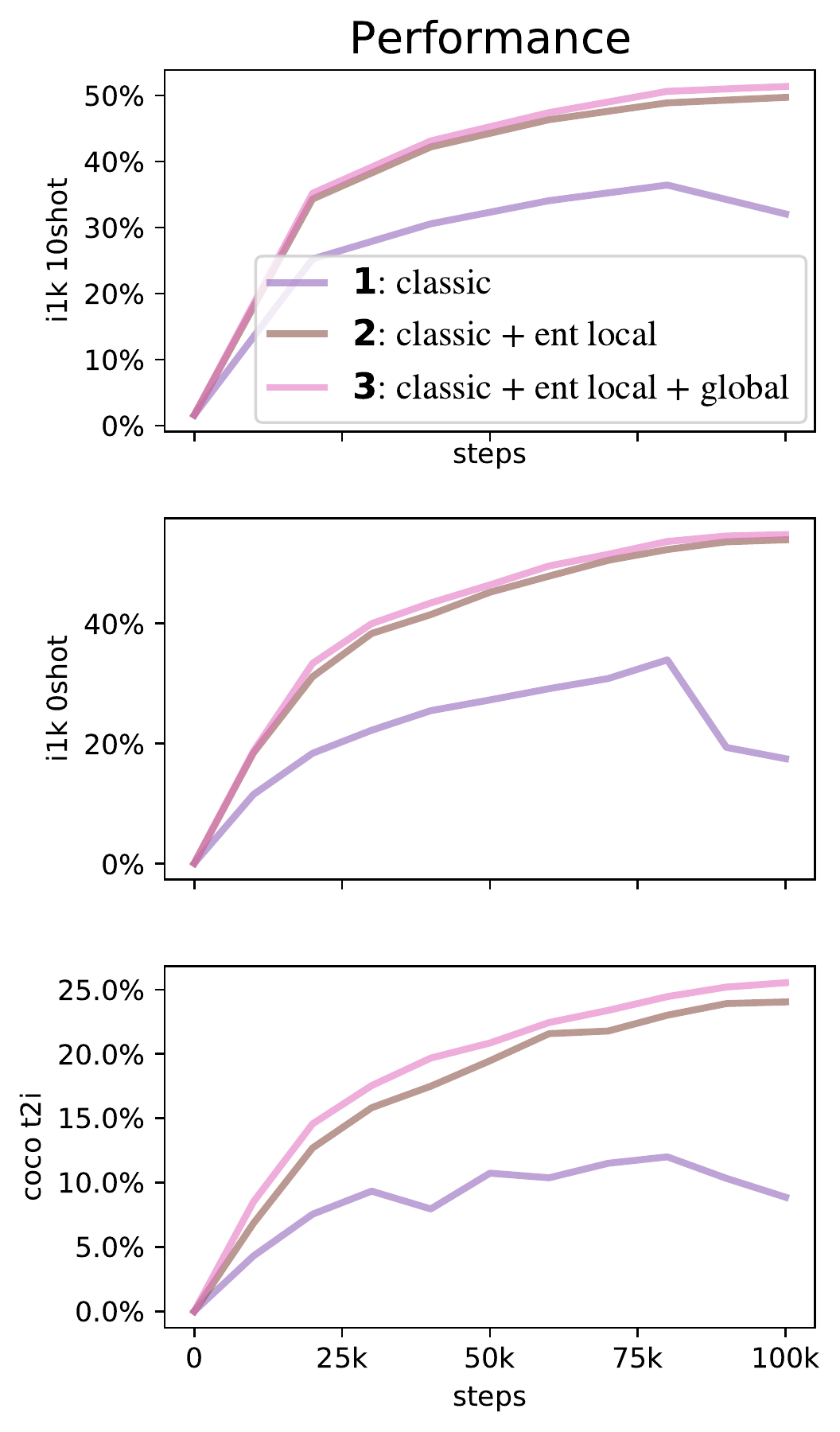}
        \caption{\tiny{Performance w.r.t. aux losses.}}
        \label{fig:ent_analysis_performance}
    \end{subfigure}
    \begin{subfigure}[c]{0.74\textwidth}
        \centering
        \includegraphics[width=\textwidth, bb=0 0 882 517]{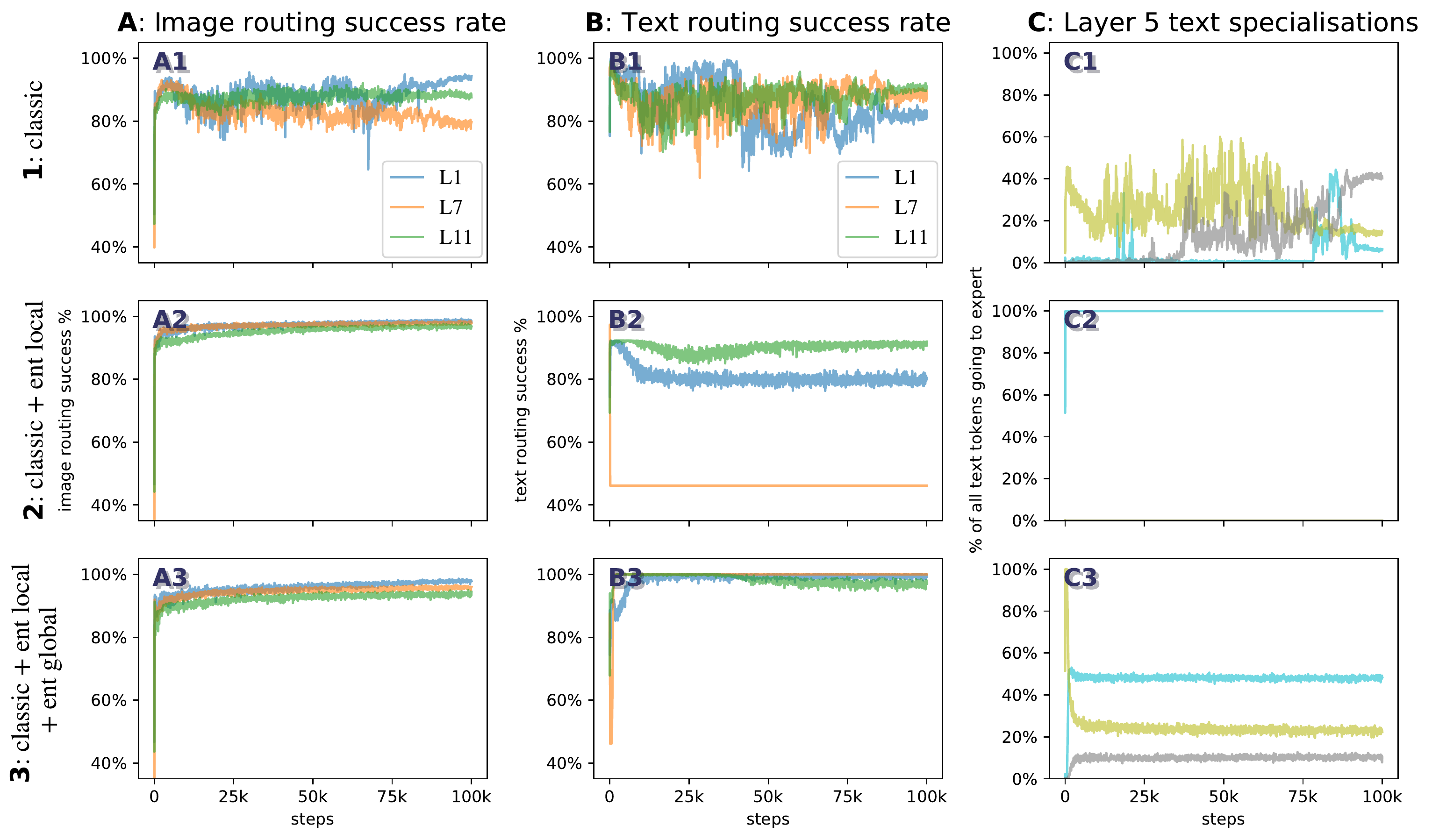}
        \caption{\tiny{Analysing routing behaviour of the auxiliary losses. \textit{First column}: Average success rate of image routing in layers 1/7/11. \textit{Second column}: Same, for text.
        \textit{Third column}: In some experts of layer 5, what fraction of all text tokens go to those experts}}         \label{fig:ent_analysis_routing}
    \end{subfigure}
	\caption{\textbf{What necessitates entropy losses?} \textit{Classic} refers to the standard formulation (importance + load losses~\cite{riquelme2021vmoe}). We add the local entropy loss to text tokens (middle row), followed by the global entropy loss (bottom row). \textbf{Left:} The ``classic'' setting is low-performing and unstable. \textbf{Right:} Analyzing the entropies shows us why: Without the local loss, the model is prone to unstable changes in expert preferences (C1), and routing success rates are low (A1, B1). The local loss fixes this but causes distributional collapse for one modality (C2), with all text tokens going to one expert (expert 11); this causes even poorer text success rates (B2). This is addressed by the global loss, which has stable expert allocations (C3) and consistently high success rates (A3, B3).
	}
	\label{fig:entropy_analysis}
	\end{widepage}
% 	\vspace{-1cm}
\end{figure}

\textbf{Effects of the losses.} Figure~\ref{fig:entropy_analysis} shows why these losses are necessary. With the default losses, modality-specific experts naturally emerge, but the router often changes its preference. This results in unstable training and poor success rate, particularly for the text modality.
The local entropy loss encourages concentrated router weights ($p_\text{text}(\texttt{experts} | \vx_i)$'s have low entropy), but at the expense of the \textit{diversity} of the text experts: the same expert is used for all text tokens (the marginal $\tilde{p}_\text{text}(\texttt{experts})$ also has low entropy), leading to dropping. In this setup, many layers have poor text success rates.

To address this, $\Omega_\text{global}$ encourages maximization of the marginal entropy, thus pushing $\tilde{p}_\text{text}(\texttt{experts})$ towards a more uniform expert distribution. The result is diverse expert usage, stable and confident routing, and high success rates. These are consequently the most performant models. 

Intuitively, it is desirable for text tokens to use multiple experts, but not all of them. In order to allow flexibility, we threshold the global entropy loss as $\Omega^\tau_\text{global}(\mG_m) = \max\{0, \tau + \Omega^\text{global}(\mG_m)\}$, such that the model is encouraged to have a certain minimum entropy, but after exceeding that, the loss is not applied. This avoids distributional collapse but does not apply overly restrictive priors on the routing distribution, as there are many optimal solutions. This can be thought of as a ``soft minimum'' $S$. With $\tau = \log(S)$, the model must use at least $S$ experts to minimize the loss (either a uniform distribution across $S$ experts -with entropy $\log(S)$-, or a non-uniform distribution using more than $S$). Figure~\ref{fig:ent_analysis_routing} shows the latter occurs; the empirical effect of these thresholds is analysed in Section~\ref{sec:design_choice_aux_losses}.

\textbf{Connection with mutual information.} The sum $\Omega_\text{local}(\mG_m) + \Omega_\text{global}(\mG_m)$ corresponds to the (negative) mutual information~\cite{cover1999elements} between experts and tokens, conditioned on the modality $m$, which we write $-\text{MI}_m(\texttt{experts}; \vx)$.
For each modality taken separately, we are effectively encouraging the knowledge of the token representation to reduce the uncertainty about the experts selection. We also tried other variants of the losses which exploit this connection, such as the mutual information between the experts and modalities, $-\text{MI}(\texttt{experts}; m)$, obtained by first marginalizing the tokens. %We found that variant to typically perform similarly or worse, especially as the architecture scale increases (see Section~\ref{sec:design_choice_aux_losses}), but believe it is a promising direction for a simple, performant auxiliary loss. % Can leave these details for the results!

\subsubsection{Priority routing}
\label{sec:bpr}
With Top-$K$ routing, some token dropping is virtually inevitable.
Batch Priority Routing (BPR)~\cite{riquelme2021vmoe} actively decides which tokens to skip based on their routing weights.
It assumes that tokens with a large routing weight are likely to be informative, and should be favored.
BPR was mostly used at inference time in \cite{riquelme2021vmoe}, allowing for smaller expert capacity buffers.
In this setup, one must take care not to systematically favor one modality over the other, for instance, by determining which token to drop based on their rank in the batch, which are usually grouped according to the token modality. 
BPR provides an essential stabilisation effect during training (Figure~\ref{fig:bpr}); we show that it does not trivially rank one modality over another, and it cannot be replaced by other methods of re-ordering the batch.
In the appendix we further show how routing priorities compare across text and images.

%% file: sections/main_experimental.tex
\section{Experiments}
\label{sec:experiments}
We study \mmoe{} in the context of multimodal contrastive learning.
We first perform a controlled comparison of \mmoe{} to an equivalent ``standard'' dense Transformer, across a range of model sizes.
We then show that when scaled up \mmoe{} can reach a high level of performance.
Finally, we ablate the various design decisions leading to \mmoe{} in Section~\ref{sec:ablations}.

\textbf{Training data.}
By default, all models are trained on paired image-text data used in~\cite{zhai2021lit}, consisting of 3.6B images and alt-texts scraped from the web.
For {large~\mmoe-H/14} experiment, we also co-train with JFT-4B~\cite{zhai2021scalingvit}.
We construct artificial text captions from JFT by comma-delimited concatenation of the class names~\cite{pham2022basic}.
Appendix~\ref{app:training_details} contains full details of our training setup.

\textbf{Evaluation.} 
Our main evaluation is ``zero-shot'': the model uses its text representations of the classes to make predictions on a new task without extra training data~\cite{socher2013zeroshot,radford2021clip}.
We focus on image classification accuracy on ImageNet~\cite{deng2009imagenet} and cross-modal retrieval on MS-COCO~\cite{lin2014coco}, following the protocol in~\cite{zhai2021lit}.
We also evaluate \mmoe's image representations via a linear adaptation protocol~\cite{dosovitskiy2020vit}, and report 10-shot accuracy on ImageNet accuracy accordingly.
Where ranges are given, they report 95\% confidence intervals across three trials.
%\rj{I am wondering if we should not explain (at least it was not trivial for me initially) something along those lines: ``While perhaps surprising at first sight, we do expect the zero-shot accuracy on ImageNet do be better than the corresponding 10-shot accuracy since we compute the latter with the simpler linear approach from~\cite{dosovitskiy2020vit}.''}%\bm{I think we have no really thorough way to show this, only handwavy logic :(

\subsection{Controlled study across scales}
We train a range of \mmoe~models at batch size $16$k for $781$k steps.
This matches the number of training examples used for CLIP~\cite{radford2021clip}. Due to use of different training data and additional tricks, a direct comparison is difficult; we therefore train dense one-tower models as baselines.
% We also provide a comparison to a two-tower setup in Section~\ref{sec:ablations}. % Is this necessary?
All models activate $k=1$ experts per token, similar to Switch Transformer~\cite{fedus2022switch}. 
% To some extent, this means performance improvements will not be due to simply scaling FLOPS applied per input, thus enabling an understanding of whether the sparse conditional computation is a good inductive bias for multimodal modelling.

\begin{figure}[h]
  \centering
  \includegraphics[width=1.0\textwidth, bb=0 0 684 224]{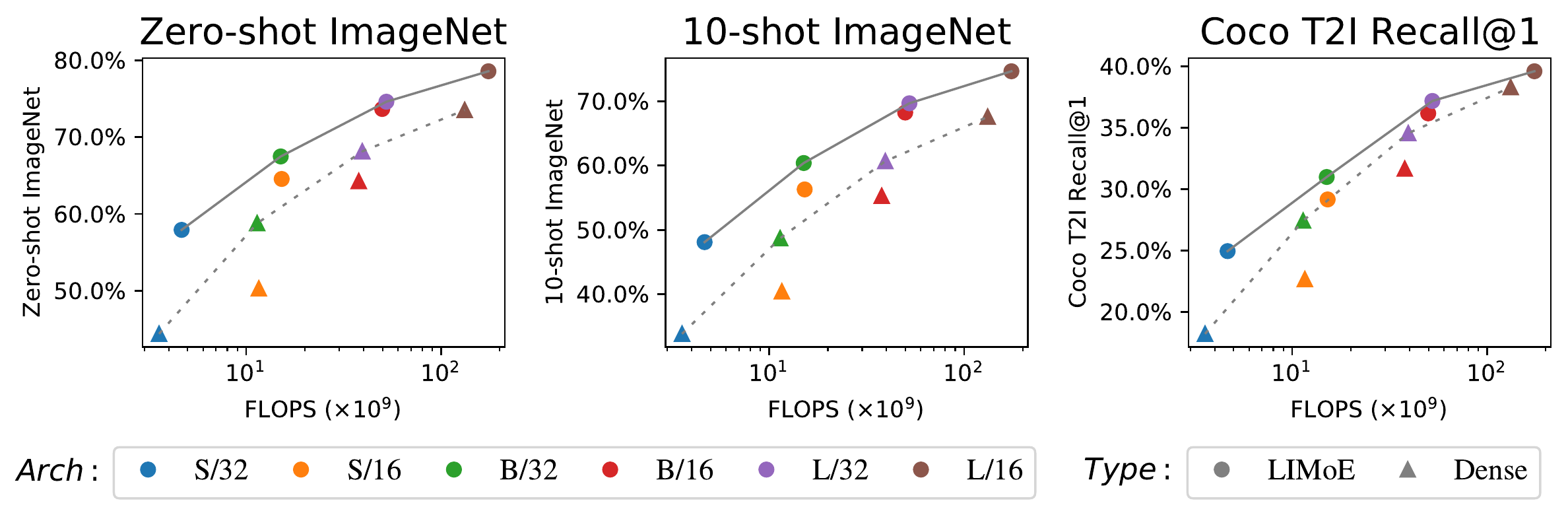}
  \caption{\mmoe{} scales well to large models, with consistent performance improvements.}
  \label{fig:headline}
\end{figure}

Figure~\ref{fig:headline} shows the performance of each model (dense and sparse) against forward-pass FLOPs (for step times and further discussion on compute costs, see Appendix~\ref{app:profiling_discussion}.).
The cost-performance Pareto frontier for \mmoe{} dominates the dense models by a wide margin, indicating that \mmoe{} offers strong improvements across all scales from S/32 , up to L/16.
The effect is particularly large on zero-shot and 10-shot ImageNet classification, with absolute performance improvements of \textbf{10.1\%} and \textbf{12.2\%} on average. % See below for mean (min - max) performance improvements across architectures, both absolute and relative, if someone wants to adjust this.
For text-to-image retrieval on COCO, \mmoe{} offers a strong boost at small scales, while at larger scales the gains are more modest but still significant. 
% Relative performance improvements
% 0shot absolute: 10.1% (4.4% - 14.3%)  -- relative: 18.9% (5.9% - 30.5%)
% 10shot absolute: 12.2% (6.4% - 15.9%)  -- relative: 27.7% (9.5% - 42.1%)
% coco_t2i absolute: 4.5% (1.1% - 6.7%)  -- relative: 19.2% (3.0% - 37.0%)
% coco_i2t absolute: 4.9% (0.2% - 8.5%)  -- relative: 13.5% (0.4% - 28.0%)

\subsection{Scaling up \mmoe{}}
We increase the architecture size, training duration, and data size to assess the performance of \mmoe{} in the large-scale regime. In particular, we train a 32-layer \mmoe-H/14 with 12 expert layers; these are non-uniformly distributed, with 32 experts per layer, and $K=1$ activated per token. It was trained at a batch size of $21$k, introducing 25\% JFT-4B images~\cite{zhai2021scalingvit} into each batch (with class names as texts). We average checkpoints towards the end of training~\cite{wortsman2022soup}; refer to Appendix~\ref{app:rocket_settings} for details.

The model contains 5.6B parameters in total, but only applies 675M parameters per token. All routers combined account for less than 0.5M parameters. Table~\ref{tab:sota_models} shows its performance alongside current state-of-the-art contrastive models.
\mmoe{} achieves \rocketacc\% zero-shot ImageNet classification accuracy with a comparably modest architecture size and training counts. \mmoe{} is fully trained from scratch, without any pre-trained components, and is the first competitive model with a shared backbone.

In light of its modality agnostic approach, this result is surprisingly strong.
Large models handling dozens of distinct tasks are increasingly popular~\cite{reed2022gato}, but do not yet approach the state-of-the-art in these tasks. We believe the ability to build a generalist model with specialist components, which can decide how different modalities or tasks should interact, will be key to creating truly multimodal multitask models which excel at everything they do. \mmoe{} is a promising first step in that direction.

\begin{table}[hbt]
\centering
\caption{Comparing state of the art zero-shot classification models. At a relatively modest scale, \mmoe-H/14 is comparable with the best two-tower models, and it is the first performant one-tower model at this scale.
T-\texttt{x} refers to a Transformer~\cite{vaswani2017attn} with the equivalent parameters of ViT-\texttt{x}~\cite{dosovitskiy2020vit}.\\
\tiny{\textit{
\textbf{Key: }$^*$ Pretrained $\qquad$ \textsuperscript{PT} Examples seen during pretraining $\qquad$ $^\dagger$ Uses FixRes~\cite{touvron2019FixRes} $\qquad$ 
$^\mathsection$ Other non-contrastive training objective
}}%
\label{tab:sota_models}}
\vspace{-6pt}
\begingroup
\setlength{\tabcolsep}{3pt}
\begin{tabular}{l@{\hskip 10pt}lllll@{\hskip 15pt}rrrr}
\toprule
& \multicolumn{2}{c}{Architecture} & Batch & Examples seen & Parameters& \multicolumn{4}{c}{ImageNet top-1 \%} \\
                    & Image & Text & size  &               & per token & Test & V2 & R & A \\
\midrule
COCA$^\mathsection$~\cite{yu2022coca} & ViT-g  & T-g & 65k & 32.8B & 2.1B & 86.3   & 80.7      & 96.5    & 90.2   \\
BASIC~\cite{pham2022basic}             & CoAtNet-7$^*$ & T-H$^{*}$ & 65k  & 19.7B\textsuperscript{PT} +32.8B & 3B  & 85.7 & 80.6      & 95.7    & 85.6   \\
LIT~\cite{zhai2021lit} & ViT-g$^*$ & T-g & 32k & 25.8B\textsuperscript{PT} + 18.2B & 2.1B & 84.5   & 78.7      & 93.9    & 79.4\\
ALIGN~\cite{jia2021align}  & EffNet-L2 & T-L$^*$ & 16k & 19.8B & $\sim$ 820M & 76.4        & 70.1       & 92.2       & 75.8        \\
CLIP~\cite{radford2021clip} & ViT-L/14$^\dagger$  & T-B & 32k & 12.8B & $\sim$ 400M & 76.2        & 70.1       & 88.9       & 77.2        \\
\midrule
\mmoe                                  & \multicolumn{2}{c}{H/14} & 21k & 23.3B & 675M & \rocketacc        & 77.7       & 94.9       & 78.7       \\
\bottomrule
\end{tabular}
\endgroup
\end{table}

%% file: sections/ablations.tex
\section{Ablations}
\label{sec:ablations}
We use a smaller setup to study various aspects of \mmoe{}. We train B/16 models at batch size 8096 for 100,000 steps (see Appendix~\ref{app:ablation_settings} for further details). Table~\ref{tab:ablations_base} shows the average over three trials of this setting alongside dense one-tower and two-tower baselines.
\mmoe{} greatly outperforms both dense models on ImageNet 0- and 10-shot, while confidence intervals overlap for retrieval with two towers.
The two-tower model is twice as large and expensive, and still falls behind the sparse one.
% In this section, we present a number of ablation and comparative studies for some key algorithmic aspects of \mmoe{}.
\begin{table}[h]
\caption{Baselines for ablations: B/16 with batch size 8096 trained for for 100,000 steps. \\ 0shot and 10shot columns show accuracy (\%), t2i and i2t show recall@1 (\%).}
\label{tab:ablations_base}
\centering
\begin{tabular}{lrrrr}
\toprule
           Model &                  i1k 0shot &                 i1k 10shot &                   coco t2i &                   coco i2t \\
\midrule
 dense one-tower &  49.8 \ci{49.2}{50.4} &  43.8 \ci{43.3}{44.3} &  23.7 \ci{23.4}{24.0} &  36.7 \ci{34.6}{38.9} \\
 dense two-tower &  54.7 \ci{54.1}{55.2} &  47.1 \ci{46.7}{47.6} &  26.6 \ci{26.2}{27.1} &  41.3 \ci{40.6}{42.0} \\
           \mmoe &  56.9 \ci{56.7}{57.1} &  50.5 \ci{50.2}{50.8} &  25.6 \ci{23.9}{27.3} &  39.7 \ci{37.1}{42.2} \\
\bottomrule
\end{tabular}
\end{table}

\subsection{Routing and auxiliary losses}
\label{sec:design_choice_aux_losses}

\textbf{Choice of auxiliary losses.}
With the introduction of the entropy based losses in addition to classic ones, there are 7 possible auxiliary losses. We aimed to find the simplest combination of these which obtains good performance. 
To study this, we performed a large sweep of auxiliary losses: for $N\in [2,\ldots, 5]$, we considered all $\binom{7}{N}$ possible loss combinations. Table~\ref{tab:aux_losses_sweep} shows, for each loss, the highest performing model with and without that loss.
Some conclusions stand out: Both entropy losses are important for text, but for images, the global loss is not impactful and the local loss is harmful. The final combination of losses was chosen based on validation accuracy alongside qualitative observations around training stability and routing success rate.

\begin{table}
\caption{Across 121 combinations, each row shows the best accuracy (\%) of all combinations that \textit{included} the auxiliary loss (\cmark) vs. those that did not (\xmark). Bold auxiliary losses indicate they are in \mmoe{}. Validation accuracy is the average contrastive accuracy in a minibatch of size 1024.}
\label{tab:aux_losses_sweep}
\centering
\begingroup
\setlength{\tabcolsep}{5pt}
\begin{tabular}{l@{\hskip 18pt}rr@{\hskip 18pt}rr@{\hskip 18pt}rr}
\toprule
               & \multicolumn{2}{r@{\hskip 18pt}}{Validation} 
               & \multicolumn{2}{r@{\hskip 18pt}}{0shot} 
               & \multicolumn{2}{r}{10shot}\\
Auxiliary loss &   \xmark &   \cmark            &\xmark&\cmark              &\xmark&\cmark \\
\midrule
                Importance &   70.5 &  70.6 &  55.4 &  56.2 &  51.1 &  51.3 \\
             \textbf{Load} &   70.3 &  70.6 &  56.2 &  55.7 &  51.3 &  51.1 \\
           \textbf{Z-Loss} &   70.3 &  70.6 &  55.8 &  56.2 &  50.5 &  51.3 \\
 \textbf{Global Ent Image} &   70.6 &  70.5 &  56.0 &  56.2 &  50.8 &  51.3 \\
  \textbf{Global Ent Text} &   69.1 &  70.6 &  54.3 &  56.2 &  51.1 &  51.3 \\
           Local Ent Image &   70.6 &  68.7 &  56.2 &  53.5 &  51.3 &  47.5 \\
   \textbf{Local Ent Text} &   67.2 &  70.6 &  53.3 &  56.2 &  47.5 &  51.3 \\
\bottomrule
\end{tabular}
\endgroup
\end{table}

\textbf{Threshold for global entropy losses.}
In Section~\ref{sec:entropy_aux_losses}, we introduced a threshold $\tau$ to encourage balanced expert distributions without forcing all modalities to use all experts.
To understand the importance of this threshold, we sweep over it for both the image and text global entropy losses.
Appendix~\ref{app:global_ent_threshold} contains a full analysis; the most important conclusions are:
\begin{itemize}[leftmargin=1em]
    \item $\tau_\text{image}$ did not affect the number of experts used for images, as global entropy was always high. Aside from these threshold experiments with very high $\tau_\text{image}$, this loss is usually inactive. It was used in our main experiments, but can likely be removed in future work.
    \item The threshold $\tau_\text{text}$ behaved exactly as a soft minimum for text experts: Sweeping $\tau_\text{text}$, we typically observed approximately $S = e^{\tau_\text{text}}$ text experts.
    \item Performance is robust to different values of $\tau_\text{text}$, provided it is not too low. 
    A low $\tau_\text{text}$ can be useful to limit the number of text experts, for later pruning, see Appendix~\ref{app:xp_pruning}.
     %\carlos{Interesting. May require further thoughts.}
\end{itemize}

\textbf{Mutual-information auxiliary loss.}
In Section~\ref{sec:entropy_aux_losses}, we discussed an alternative loss, namely $-\text{MI}(\texttt{experts}; m)$, based on the mutual information between experts and modalities. While it has the advantage of merging the local and global entropy losses for both the text and image modalities into a single term, without threshold parameters, it leads to slightly worse results: in a comparable setup, it had 1.5\% and 0.1\% worse zero-shot and 10-shot performance compared to Table~\ref{tab:ablations_base}.

\begin{wrapfigure}[11]{r}{0.45\textwidth}
    \vspace{-0.85cm}
    \includegraphics[width=0.45\textwidth, bb=0 0 443 307]{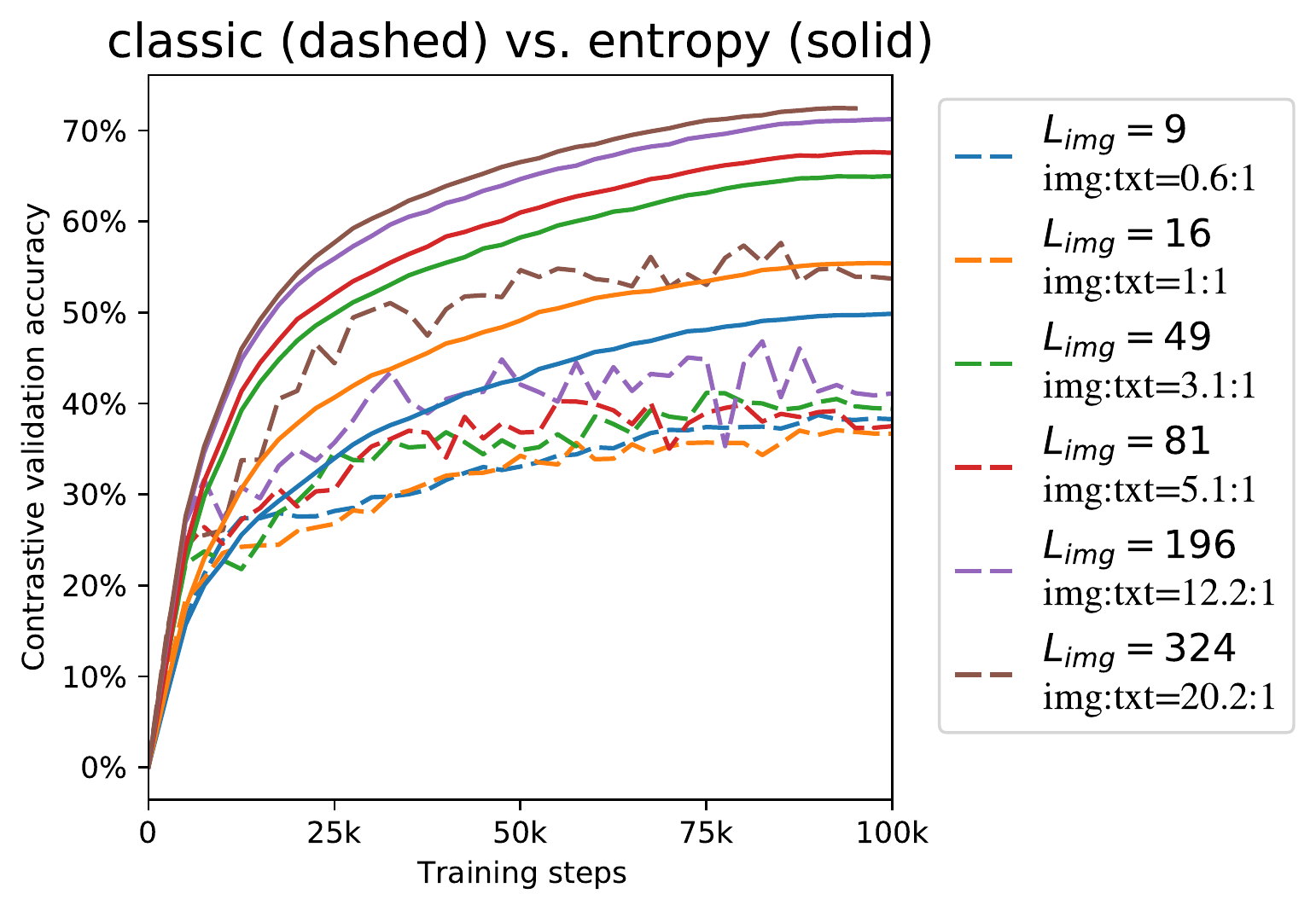}
    \caption{
    \textbf{Entropy losses are not just addressing a modality imbalance}. 
    With different image:text balancing, including completely balanced, the entropy losses substantially improves over the classic setting.}
    \label{fig:balancing}
\end{wrapfigure}

\paragraph{The effect of modality balancing.}
Our models use a text sequence length of 16, but image sequence lengths from 49 to 400 (for these ablations, 196).

Our ablations reveal that the entropy losses are most important when applied to the text tokens.
This leads to a hypothesis that these are only necessary or useful in the imbalanced case.
To test this, we vary the modality balance of \mmoe-B/16 by varying the patch size; this enables us to control the number of image tokens, and hence image:text balance, without changing the information content in the data.
Figure~\ref{fig:balancing} shows the results.
First, we observe that, with entropy routing, a longer image sequence length is always better.
This shows that entropy routing can effectively handle highly imbalanced setups, and mirrors the observation that for classical Vision Transformers: a longer sequence is better.
Importantly, entropy routing is always far superior to the classical setup with growing gaps, even when the modalities are balanced 1:1 ($L_{\text{img}}=16$).
This experiment also confirms the robustness of entropy routing to different setups.

\begin{figure}[h]
  \centering
  \includegraphics[width=0.75\textwidth, bb=0 0 627 235]{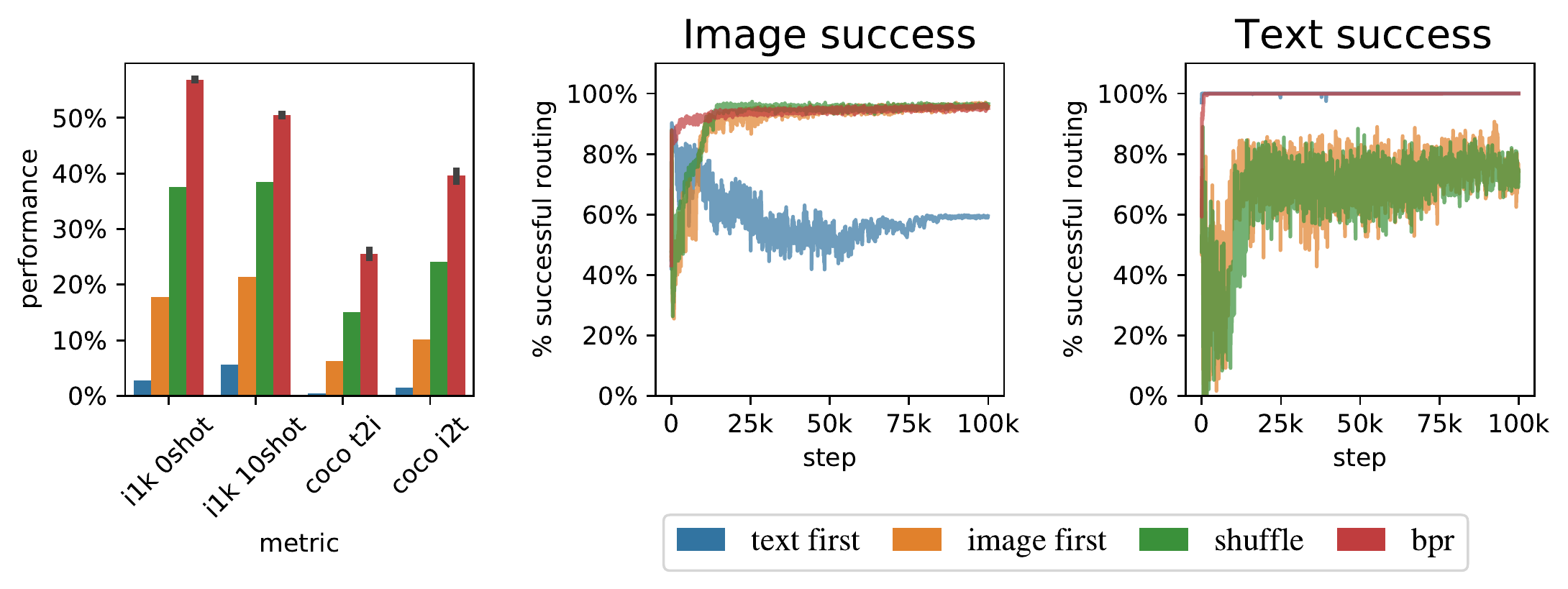}
  \caption{\textbf{BPR stabilizies training and enables performant models}; the first figure shows different performance metrics. The last two show \textit{success rates} for the MoE router in Layer 9.}
  \label{fig:bpr}
\end{figure}

\textbf{Batch priority routing as a training stabilizer.} Figure~\ref{fig:bpr} shows the effect of BPR during training. BPR not only ameliorates against token dropping, but also improves training stability. Models with no dispatch order intervention (first-in-first-out) perform extremely poorly, whether we route images first or text first.
These routers have low success rate. 
Randomly shuffling tokens (i.e. deciding which tokens to drop at random when an expert becomes full) partially ameliorates this, but its performance is still much worse than that of models trained with BPR. We further analyse BPR in Appendix~\ref{app:analysis_bpr_rankings} and show that it does not simply rank one modality above another. 

\subsection{Other ablations}
We summarize our other ablations here due to space constraints; details can be found in  Appendix~\ref{app:xp}.

\textbf{Router structure} (\textit{Appendix~\ref{app:xp_router_design}}). Our router is modality agnostic; we experiment with per-modality routers, and separate pools of per-modality experts. We find they all perform comparably to our generic, modality agnostic setup, but that separate pools of experts by design is more stable and does not require auxiliary losses for regularisation---while harder to scale to many modalities and tasks.

\textbf{Increasing selected experts per token $K$} (\textit{ Appendix~\ref{app:xp_increasing_k}}). We propose modifications to BPR and the local auxiliary loss to generalise to $K > 1$; by doing so we can steadily increase performance by increasing $K$, e.g. from 55.5\% zero-shot accuracy with $K=1$ to 61.0\% with $K = 5$.

\textbf{Total number experts} (\textit{Appendix~\ref{app:xp_increase_num_experts}}). We show that increasing the pool of available experts at fixed $K$ improves performance (unlike what was observed for vision-only tasks~\cite{riquelme2021vmoe}).

\textbf{Expert pruning} (\textit{Appendix~\ref{app:xp_pruning}}). We show using simple heuristics we can prune down to modality-specific experts for unimodal forward passes, thus avoiding expert collapse under unimodal batches.

\textbf{Training on public data} (\textit{Appendix~\ref{app:xp_laion}}) The majority of \mmoe{} models were trained on proprietary data~\cite{zhai2021lit}. We show that LIMoE works similarly well on publically available data, retaining performance improvements against a comparable dense model.

%% file: sections/analysis.tex
\section{Model Analysis}
\label{sec:analysis}
In this section, we explore some of the internal workings of \mmoe.
We use simple B/32 and B/16 models with 8 experts, and the large H/14 with 32. See Appendix~\ref{app:analysis} for further details and experiments.

\textbf{Multimodal experts arise} \textit{(Appendix~\ref{app:analysis_routing_distributions}).} Aside from encouraging diversity, we do not explicitly enforce experts to specialize. Nonetheless, we observe the emergence of both modality-specific experts, and multimodal experts which process both images and texts (per-expert distributions in~\ref{app:analysis_routing_distributions}).

\textbf{Qualitative analysis} \textit{(Appendix~\ref{app:analysis_routing_examples}).} We analyse some example data and show a clear emergence of semantically meaningful experts. With images for instance, some experts specialize on lower level features (colours, lines) while others on more complex features (faces and text), see Figure~\ref{fig:limoe_h14_examples_l17_main}.

\textbf{BPR ranking} \textit{(Appendix~\ref{app:analysis_bpr_rankings}).} The local loss encourages high max-routing weights for text, and BPR ranks according to this. We show however that this does not mean text is always prioritised first: Especially in later layers, the model often prioritises important image patches over text.

%% file: sections/literature_review.tex
\section{Related work}
\label{sec:lit_review}
Unimodal, task-specific neural networks have long been researched, with increasing convergence towards Transformer-based architectures~\cite{vaswani2017attn,devlin2019bert} for both NLP~\cite{tay2022transformerssurvey} and Computer Vision~\cite{dosovitskiy2020vit,liu2021swin, jeeveswaran2022vitsurvey}. \textit{Multimodal models} aim to process multiple types of data using a single neural network. 

Many approaches ``fuse'' modalities~\cite{tan2019lxmert,su2020vlbert,lu2019vilbert,li2019visualbert} to tackle inherently multimodal tasks. $\mmoe$ is more similar to approaches which do not do that, and still operate as unimodal feature extractors. Some co-train on distinct tasks~\cite{polyvit,akbari2021vatt,li2021vit_bert_cotrain,reed2022gato} without aligning or fusing representations---effectively sharing weights across tasks---whereas others include both unimodal aspects and fused multimodal aspects for functionality in both contexts~\cite{weng2021ufo}.

We build on deep \textit{Sparse Mixture of Experts models}, which have been studied independently in Computer Vision~\cite{riquelme2021vmoe,lou2022mixermoe} and NLP~\cite{shazeer2017outrageously,lepikhin2021gshard,fedus2022switch}, typically in the context of transfer learning. These models use a learned gating mechanism whereby only a subset of $K$ experts out of $E \gg K$ are activated for a given input. Many works aim to improve the gating mechanism itself, by making it differentiable~\cite{hazimeh2021dselectk}, reformulating as a linear assignment task~\cite{lewis2021base} or even swapping it out for a simple hashing algorithm~\cite{roller2021hash}. MoE models have also been studied for multitask learning~\cite{hazimeh2021dselectk}, with per-task routers~\cite{ma2018mmoe} but a shared pool of experts. To our knowledge, sparse models have not been explored for multimodal learning.

A large body of research exists on contrastive learning, usually in self-supervised~\cite{chen2020simclr} but also in supervised regimes~\cite{khosla2020supcon}. \textit{Multimodal contrastive learning} trains on aligned data from multiple modalities. Originally studied for medical images and reports~\cite{zhang2020convirt}, it was recently scaled to noisy web data~\cite{radford2021clip,jia2021align}, where strong image-text alignments enabled performant image classification and cross-modal image-text retrieval without finetuning on downstream data. Follow up works improved upon this significantly by scaling up and using pretrained models~\cite{pham2022basic, zhai2021lit} and multitask training with generative modelling~\cite{yu2022coca} or other vision tasks~\cite{lu2021florence}. These works use unimodal models which \emph{separately} process image and text data; we are not aware of previous research using a single model to process both images and texts for contrastive learning, neither with dense nor with sparse models.

%% file: sections/conclusions.tex
\section{Conclusions and Future Work}
\label{sec:conclusions}
We have presented~\mmoe, the first multimodal sparse mixture of experts model. We uncovered new failure modes specific to this setup and proposed entropy based auxiliary losses which stabilises training and results in highly performant models. It works across many model scales, with average improvements over FLOP-matched dense baselines of +10.2\% zero-shot accuracy. When scaled to a large H/14 model, we achieve \rocketacc\% accuracy, competitive with current SOTA approaches. 

\textbf{Societal impact and limitations}: The potential harms of large scale models~\cite{hai2020foundationmodels}, contrastive models~\cite{radford2021clip} and web-scale multimodal data~\cite{birhane2021multimodal} also carry over here, as \mmoe{} does not explicitly address them. On the other hand, it has been shown that \textit{pruning} models tends to cause low-resource groups to be forgotten~\cite{hooker2020pruningbias}, causing performance to disproportionally drop for some subgroups. This would be worth considering for our expert-pruning experiments, but by analogue, the ability to scale models with experts that can specialize deeply may result in better performance on underrepresented groups.

Environmentally speaking, training large models is costly, though efforts are made to use efficient datacenters and offset emitted CO\textsubscript{2}. Prior works however show that most environmental impact occurs during model inference, and that MoEs are significantly more efficient in that regard~\cite{patterson2021carbon}; \mmoe{} is naturally a good candidate for efficient, large-scale multimodal foundation models.

\textbf{Future work:} There are many interesting directions from here. The routing interference with multiple modalities still is not fully understood. In general, conclusions from applications of MoEs to NLP have not carried over perfectly to Vision, and vice-versa, and here we see again different behaviour between images and text. Naturally, extensions to more modalities should be explored; even with only two we see fascinating interactions between different data types and the routing algorithms, and that will only get more difficult, and interesting, with more modalities.

There are always more modalities to learn, and larger models to build: sparse models provide a very natural way to scale up while juggling very different tasks and data, and we look forward to seeing more research in this area.

%% file: appendix/1_training_setup.tex
\section{Training details}
\label{app:training_details}
All models were trained with adafactor, using the same modifications used for ViT-G~\cite{zhai2021scalingvit}. Unless otherwise specified, we use learning rate \num{1e-3} and decoupled weight decay of magnitude \num{1e-5}. We use a cosine learning rate decay schedule, with a linear warmup (40k steps for longer scaling study models, 10k steps for ablations). Models were trained on a mixture of Cloud TPU-v2, v3 and v4 pods.

Models were trained with 32 experts, with experts placed every 2 layers -- except where explicitly stated. Otherwise, architecture parameters (e.g. hidden size, number of layers) follow those of ViT~\cite{dosovitskiy2020vit}.
All models except for \mmoe-H/14 use dimensionality 512 for the final output representation; this final representation is cast to bfloat16 precision for reduced all-to-all costs and increased memory efficiency.
The learned contrastive temperature parameter is initialised at 10. Text sequences are tokenized to a sequence length of 16 using the T5 SentencePiece vocabulary~\cite{raffel2020t5}. Images were linearly renormalized to a value range of \texttt{[-1, 1]}.

\subsection{Scaling study}
\label{app:scaling_settings}
We train models at batch size 16,384 for 781,250 steps at resolution 224. This trains for the same number of examples as CLIP~\cite{radford2021clip}; they however use a larger batch size (32768), increase resolution in the final epoch, and use a larger dimensionality for the final contrastive feature representation, all of which improve performance.

\subsection{Ablations}
\label{app:ablation_settings}
These are B/16 models trained for 100,000 steps at batch size 8192. The threshold used for the text global entropy loss is $\tau_T = \log(9)$ -- that is, we incentivize the use of at least 9 experts (uniformly) or more (not necessarily in a uniform way). For images, $\tau_T = \log(20)$, but with this threshold, the loss is not applied at all and it can be ignored.

\subsection{\mmoe-H/14}
\label{app:rocket_settings}
The largest scale model is trained at batch size 21502, with resolution 288 and text sequence length 16. The global entropy loss thresholds are $\tau_\text{text} = \log(4)$ and $\tau_\text{text} = \log(25)$ for text and image respectively. There are MoE layers in 12 encoder blocks, namely, in 3, 7, 11, 15, 18, 21, 24, 26, 28, 30, 31, 32.
The default training data is mixed with data from JFT-4B with a ratio of 3:1. Text strings are generated from JFT-4B by simply concatenating the class names. JFT-4B was also deduplicated using the same method as previous works~\cite{zhai2021lit}.

\textbf{Checkpoint souping}. We adapt the methodology developed for finetuning~\cite{wortsman2022soup}, but instead combine checkpoints from the same run. We used a reverse-sqrt schedule~\cite{raffel2020t5}, which has a linear cooldown at the end. To generate diversity for the model soup, we launched multiple cooldowns, and greedily selected checkpoints to maximize zero-shot accuracy on the ImageNet validation set, using the smaller subset of prompts from CLIP~\cite{radford2021clip}. Checkpoints could be reused multiple times.

The model was trained for 700k steps pre-cooldown. There was one cooldown of length 125k steps from the final step, and 3 of length 40k steps starting from step 650k. Two of the cooldowns had no changes to the original setup described above. To generate diversity for the soup, we also trained one 40k cooldown with only JFT data, and one with no JFT data at all.

\begin{figure}[h]
  \centering
  \includegraphics[width=1.0\textwidth, bb=0 0 1059 266]{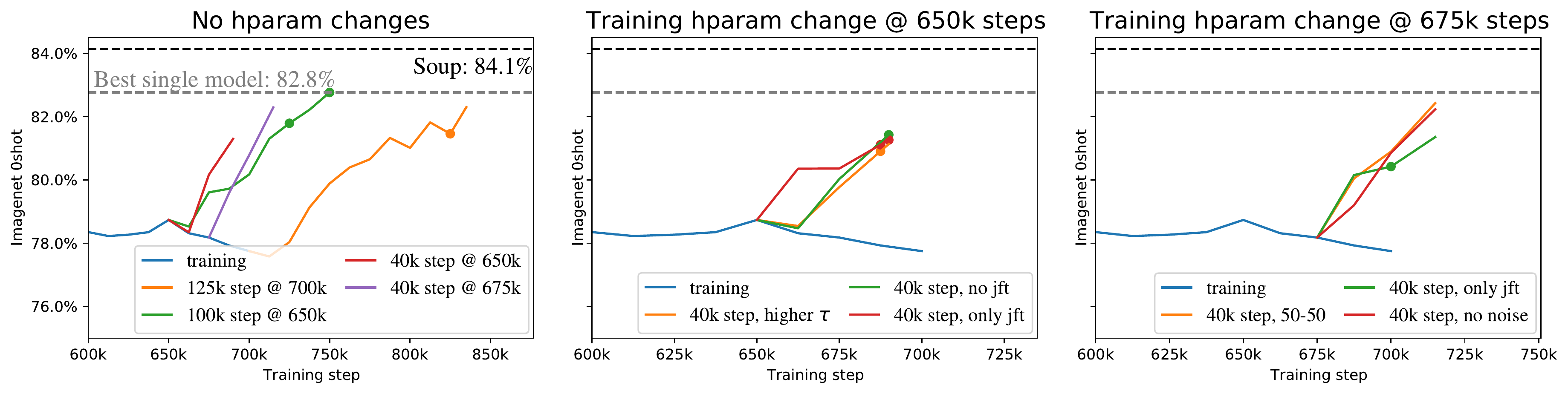}
  \caption{\textbf{Souping procedure for \mmoe{}-H/14}. Dots show checkpoints in the final model soup.}
  \label{fig:soups}
\end{figure}

Figure~\ref{fig:soups} shows the zero-shot accuracy evaluated at 12.5k step intervals during training, for all the different cooldowns, and the end of training. The final model soup consisted of 8 checkpoints in total.

% \subsection{Dataset details}
% We use the noisy web image-text data from LiT~\cite{zhai2021lit}, which contains 3.6B noisily aligned (image, text) examples. 

%% file: appendix/2_auxiliary_losses.tex
\section{Auxiliary losses}
\label{app:aux_losses}

\subsection{Definitions of all the auxiliary losses}
% \bm{TODO: Add aux losses}

In Section~\ref{sec:design_choice_aux_losses}, we study multiple combinations of auxiliary losses. For completeness, we recall below all their definitions.
Given a token $\vx \in \R^D$, we denote by $g(\vx) = \texttt{softmax}(\mW \vx) \in \R^E$ the gating weights across the $E$ experts, with $\mW \in \R^{E \times D}$ being the routing parameters. When we deal with a batch of multiple tokens $\{\vx_i\}_{i=1}^n$, we use the notation $\mX\in\R^{n \times D}$.

\textbf{Importance loss.} We consider the definition from~\cite{riquelme2021vmoe}, inspired by the original proposal of~\cite{shazeer2017outrageously}. The importance loss $\Omega_\text{imp}$ enforces a balanced profile of the gating weights across the experts. More formally, for any expert $e \in \{1,\dots, E\}$, we consider
$$
\text{imp}_e(\mX) = \sum_{\vx \in \mX} g(\vx)_e
$$
and define the loss $\Omega_\text{imp}$ via the squared coefficient of variation for $\text{imp}(\mX) = \{\text{imp}_e(\mX)\}_{e=1}^E$, namely
$$
\Omega_\text{imp}(\mX) = \left(
\frac{\texttt{std}(\text{imp}(\mX))}{\texttt{mean}(\text{imp}(\mX))}
\right)^2.
$$

\textbf{Load loss.}
Like previously, we follow~\cite{riquelme2021vmoe} whose definition is inspired by the original proposal of~\cite{shazeer2017outrageously}. We assume throughout that paragraph that the gating weights $g_\text{noisy}(\vx)$ are obtained by a noisy version of the routing, i.e., $g_\text{noisy}(\vx) = \texttt{softmax}(\mW \vx + \varepsilon)$ with $\varepsilon \sim \calN(\vzero, \sigma^2 \mI)$ and $\sigma=1/E$ (see details in~\cite{riquelme2021vmoe}). We introduce $\eta_K$ the $K$-th largest entry of $\mW \vx + \varepsilon$.

The load loss $\Omega_\text{load}$ complements the importance loss $\Omega_\text{imp}$ by trying to balance the \textit{number of assignments} across the experts. To circumvent the fact that the assignments are discrete, $\Omega_\text{imp}$ focuses instead on the probability of selecting the expert. For any $e \in \{1,\dots, E\}$, the probability is understood as the probability of having the expert $e$ still being among the Top-$K$ while resampling only the noise of that expert.
More formally, this corresponds to
$$
p_e(\vx) = 1 - \Phi\Big( \frac{\eta_K - (\mW \vx)_e}{\sigma}  \Big) 
$$
with $\Phi$ the cumulative distribution function of a Gaussian distribution.

The load loss $\Omega_\text{load}$ is eventually defined by
$$
\Omega_\text{load}(\mX) = \left(
\frac{\texttt{std}(\text{load}(\mX))}{\texttt{mean}(\text{load}(\mX))}
\right)^2
\ 
\text{with}
\ 
\text{load}(\mX) = \{\text{load}_e(\mX)\}_{e=1}^E
\ 
\text{and}
\ 
\text{load}_e(\mX) = \sum_{\vx \in \mX} p_e(\vx).
$$

\textbf{Z-loss.} The z-loss $\Omega_{\text{zloss}}$ introduced in~\cite{zoph2022stmoe} aims at controlling the maximum magnitude of the router activations $\mA = \{\mW \vx_i\}_{i=1}^n \in\R^{n \times E}$ with entries $a_{i,e} = (\mW \vx_i)_e$. The loss is defined by
$$
\Omega_{\text{zloss}}(\mX) = \frac{1}{n} \sum_{i=1}^n \left( \log\left(\sum_{e=1}^E \exp{(a_{i,e})} \right)  \right)^2.
$$

\textbf{The mutual-information loss $-\text{MI}(\texttt{experts}; m)$.} 
In Section~\ref{sec:entropy_aux_losses}, we allude to a variant of the local and global entropy losses in the form of the mutual information between the experts and the modalities (as a reminder, the sum of the local and global entropy losses corresponds instead to the (negative) mutual information between the experts and tokens, conditioned on the modality). Let us assume we have a total of $M$ modalities.
Formally, and reusing the notation from Section~\ref{sec:entropy_aux_losses}, we define $-\text{MI}(\texttt{experts}; m)$ as
$$
-\text{MI}(\texttt{experts}; m) = 
\frac{1}{M} \sum_{m'=1}^M \mathcal{H}(\tilde{p}_{m'}(\texttt{experts})) 
- 
\mathcal{H}
\left(
\frac{1}{M} \sum_{m'=1}^M \tilde{p}_{m'}(\texttt{experts})
\right)
$$
where, for each modality $m'$, we have computed the approximate marginal probability over the $n_{m'}$ tokens of that modality
$$
\tilde{p}_{m'}(\texttt{experts}) = \frac{1}{n_{m'}} \sum_{i=1}^{n_{m'}} p_{m'}(\texttt{experts} | \vx_i)
$$
and $\mathcal{H}$ denotes the entropy.

\textbf{Final aggregated auxiliary loss.}
When considering the combination of several auxiliary losses, the final auxiliary loss is computed as the average over all the losses. The average is weighted by a single regularization parameter that is a hyperparameter of our approach. After some preliminary tuning phase, we have set its value to $0.04$ in all our experiments and found this choice to be robust.

\subsection{In-depth analysis of global entropy threshold}
\label{app:global_ent_threshold}
Note again that we can view a threshold $\tau$ as a soft minimum, as the minimum number of experts which must be used by a modality to satisfy the loss is $S = e^{\tau}$. We find it more intuitive to think in terms of this soft minimum threshold $S_T$.

\begin{figure}[h]
  \centering
  \makebox[1\textwidth][c]{\includegraphics[width=1.2\textwidth, bb=0 0 1126 353]{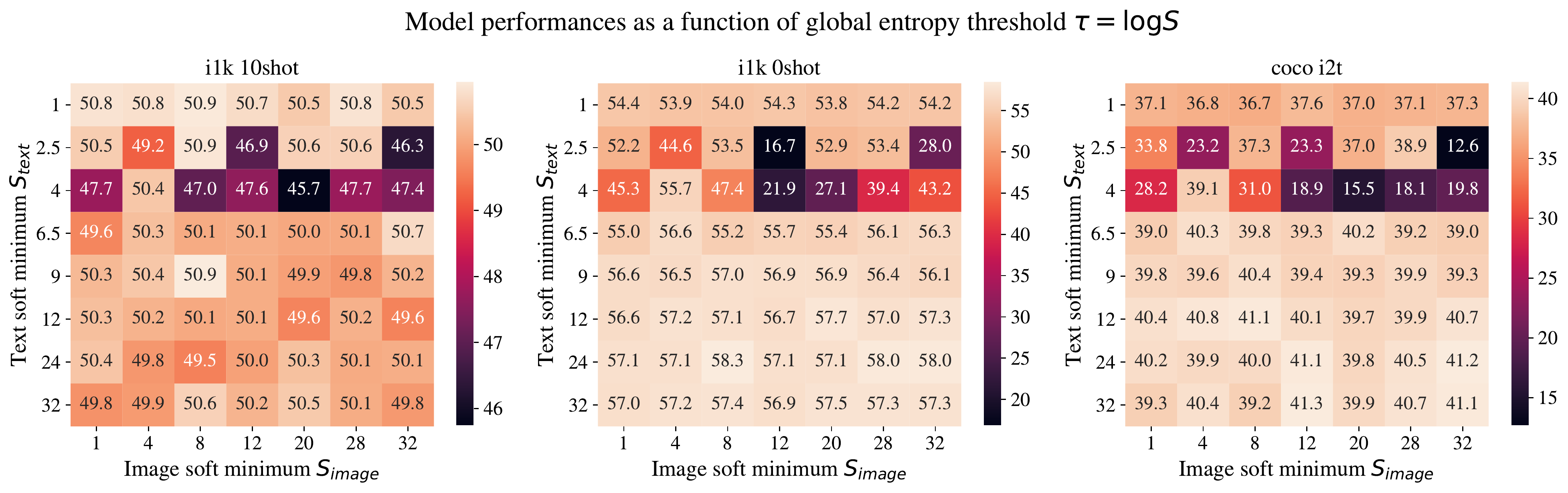}}
  \caption{\textbf{A high enough text threshold encourages stability}, but otherwise performance is somewhat invariant to the thresholds used. Note the plotted quantity is the \textit{soft minimum} $S_T = e^{\tau}$.}
  \label{fig:gt_performance}
\end{figure}

\textbf{Performance}.
Figure~\ref{fig:gt_performance} shows the effect of the threshold on performance.

There are three phenomenon of note:
\begin{enumerate}
    \item When the text threshold is too low, models are unstable and performance is poor.
    \item Past some limit however, performance of models w.r.t. text threshold is fairly consistent.
    \item Outside (and probably inside) the unstable region, the image threshold makes no systematic difference.
\end{enumerate}
%\carlos{The numbers in Figure~\ref{fig:gt_performance} are almost impossible to read. Maybe we can have one plot with two rows (2 and 1 plots per row) or something like that, and larger size.}

\begin{figure}[h]
  \centering
  \includegraphics[width=0.8\textwidth, bb=0 0 765 653]{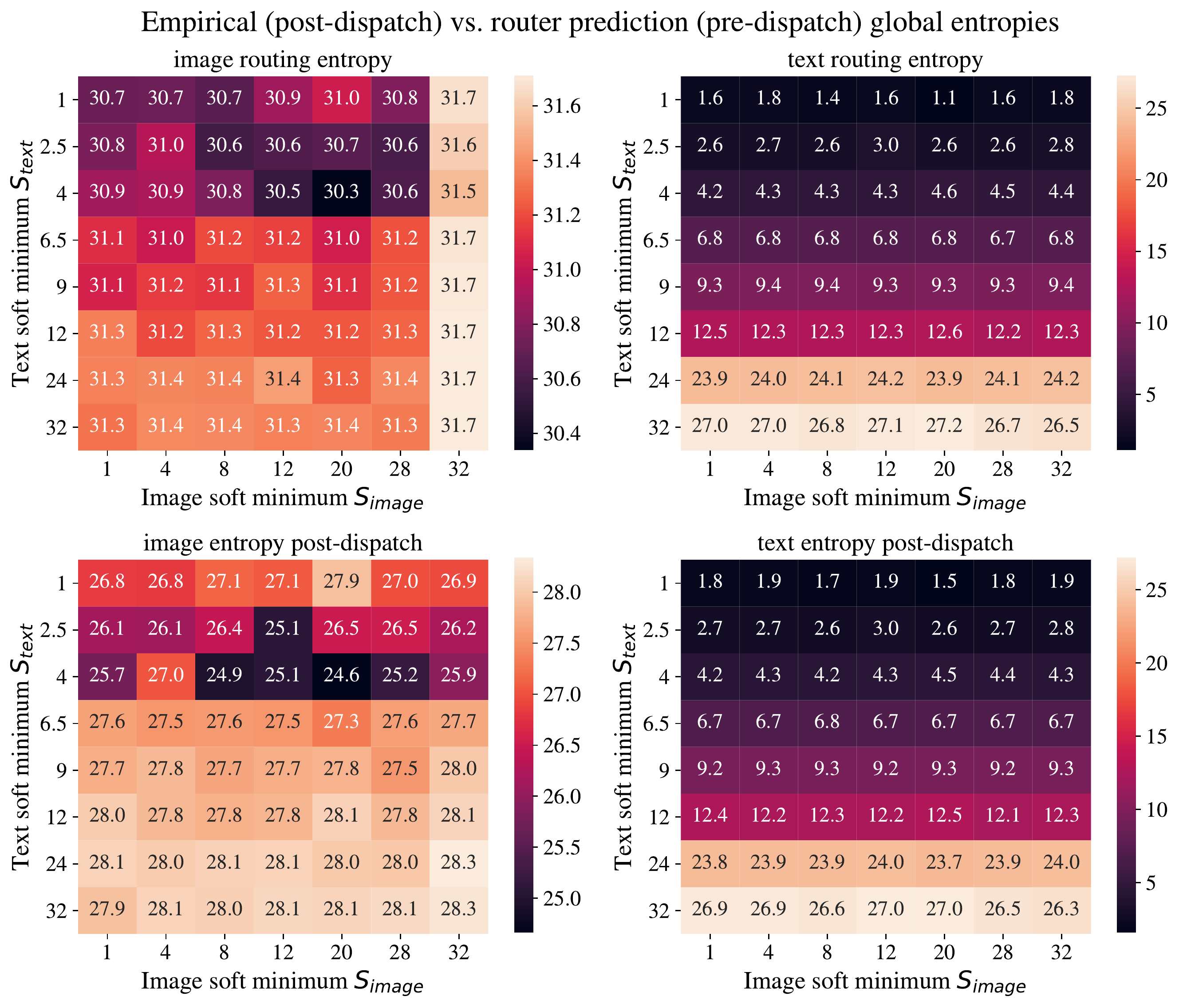}
  \caption{\textbf{Image entropy is always high, but text entropy closely tracks the target threshold}. \textit{Top}: The routing entropy is the global entropy of the predictions of the router, which is what is actually regularised. \textit{Bottom}: The post-dispatch entropy is the entropy of the distribution after top-K selection and capacity limits (token dropping) have interfered. For text tokens pre- and post-dispatch entropies pretty much coincide as their routing probabilities are high and BPR favors them --so little dropping happens. The story is a bit different for image tokens; some are dropped, and the pre- and post-dispatch entropies are not completely equivalent.}
  \label{fig:gt_empirical_ent}
\end{figure}

\textbf{Actual global entropies}.
Looking at the actual entropies of model routing helps at least explain why the image threshold is unimportant. Figure~\ref{fig:gt_empirical_ent} shows the empirical entropy. The image entropy is always large; note that when it is higher than the threshold $\tau$, the loss is not applied; ergo, for most of the settings, the global entropy loss is \textit{not applied to images}.
This also applies to almost all models trained for this paper.
On the other hand, analysing text entropies, it is clear that the model closely tracks the threshold $\tau_\text{text}$. As a side effect, image entropy tends to reduce as $\tau_\text{text}$ increases.

\begin{figure}[h]
  \centering
  \includegraphics[width=0.8\textwidth, bb=0 0 698 336]{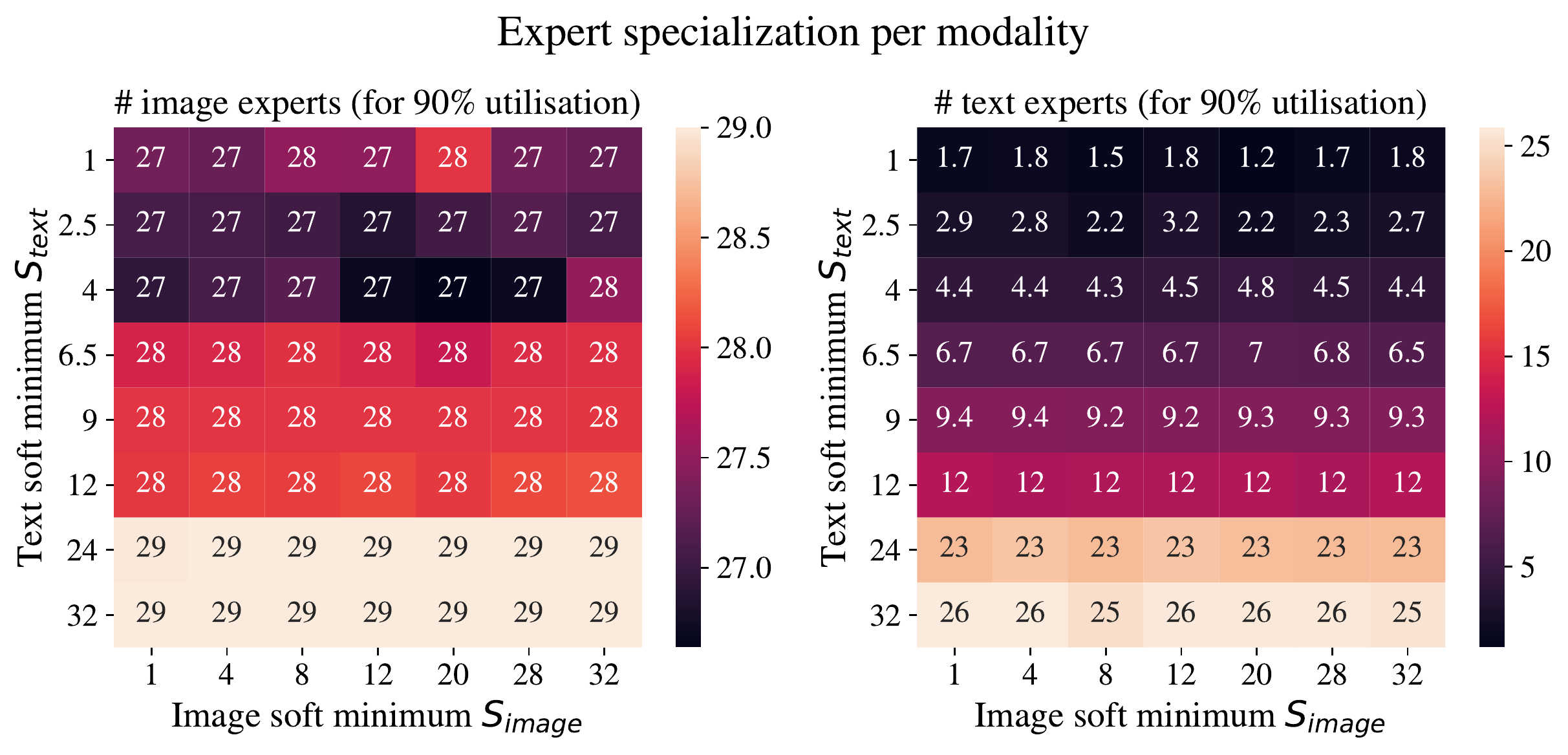}
  \caption{\textbf{Almost all experts process images, but the number of text experts closely follows the threshold}. Some experts process both modalities, so it is non trivial to classify an expert as being of a given modality. Our proxy for the number of modality specific experts is the number of experts needed to reach some routing success rate: if only 3 experts are needed for 90\% text success rate, then text routing is not highly distributed. }
  \label{fig:gt_specialization}
\end{figure}

\textbf{Expert specialization}. As discussed, the threshold can be viewed as setting an implicit soft minimum $S_T = e^{\tau}$. The number of experts actually used for each modality is shown in Figure~\ref{fig:gt_specialization}. The text threshold exactly behaves as a soft minimum; as it is increased, the model has more text experts and less image experts.

\textbf{Overall}. For text, the entropy loss behaves as expected; as it is increased, there are more text experts. A few questions remain: why does it not impact performance? Why does text behave differently than images - is it due to the imbalance between them during training, or is it simply a fundamental difference in routing behavior for the two modalities?

%% file: appendix/3_tabular_results.tex
\section{Tabular results}
\subsection{Scaling comparison, and architecture definitions.}
All results and parameters from Figure~\ref{fig:headline} are shown in Table~\ref{tab:headline_models}, alongside the results of \mmoe{}-H/14.
\begin{table}[H]
\caption{The results from Figure~\ref{fig:headline} and \mmoe{}-H/14.}
\label{tab:headline_models}
\centering
\begin{adjustbox}{center}
\begin{tabular}{rp{8.5mm}p{8.5mm}p{8.5mm}p{8.5mm}p{8.5mm}|cccc}
\toprule
Model & Patch size & Layers & Heads & Hidden size & MLP size & i1k 0shot & i1k 10shot & coco t2i & coco i2t \\
 \midrule
Dense S/32 & \multirow{2}{*}{32} & \multirow{2}{*}{8} & \multirow{2}{*}{8} & \multirow{2}{*}{512} & \multirow{2}{*}{2048} & 44.4 & 33.8 & 18.2 & 30.4 \\
\mmoe{} S/32 &  &  &  &  &  & 57.9 & 48.1 & 25.0 & 38.9 \\ \midrule
Dense S/16 & \multirow{2}{*}{16} & \multirow{2}{*}{8} & \multirow{2}{*}{8} & \multirow{2}{*}{512} & \multirow{2}{*}{2048} & 50.3 & 40.4 & 22.7 & 35.8 \\
\mmoe{} S/16 &  &  &  &  &  & 64.5 & 56.3 & 29.2 & 43.7 \\ \midrule
Dense B/32 & \multirow{2}{*}{32} & \multirow{2}{*}{12} & \multirow{2}{*}{12} & \multirow{2}{*}{768} & \multirow{2}{*}{3072} & 58.8 & 48.7 & 27.4 & 42.5 \\
\mmoe{} B/32 &  &  &  &  &  & 67.5 & 60.4 & 31.0 & 45.7 \\ \midrule
Dense B/16 & \multirow{2}{*}{16} & \multirow{2}{*}{12} & \multirow{2}{*}{12} & \multirow{2}{*}{768} & \multirow{2}{*}{3072} & 64.3 & 55.3 & 31.7 & 46.8 \\
\mmoe{} B/16 &  &  &  &  &  & 73.7 & 68.2 & 36.2 & 51.3 \\ \midrule
Dense L/32 & \multirow{2}{*}{32} & \multirow{2}{*}{24} & \multirow{2}{*}{16} & \multirow{2}{*}{1024} & \multirow{2}{*}{4096} & 68.1 & 60.7 & 34.6 & 51.2  \\
\mmoe{} L/32 &  &  &  &  &  & 74.6 & 69.7 & 37.2 & 54.5
  \\ \midrule
Dense L/16 & \multirow{2}{*}{16} & \multirow{2}{*}{24} & \multirow{2}{*}{16} & \multirow{2}{*}{1024} & \multirow{2}{*}{4096} & 73.5 & 67.6 & 38.3 & 54.3  \\
\mmoe{} L/16 &  &  &  &  &  & 78.6 & 74.7 & 39.6 & 55.7  \\ \midrule
\mmoe{}-H/14 & 14 & 32 & 16 & 1280 & 5120 & 
\rocketacc & 81.4 & 39.8 & 51.1 \\
\bottomrule
\end{tabular}
\end{adjustbox}
\end{table}

\newpage
\subsection{All tabular results}
\begin{widepage}
\tiny{%
\begin{longtable}{p{0.9cm}P{0.6cm}p{3cm}P{0.6cm}P{0.6cm}P{0.6cm}P{0.6cm}P{0.6cm}P{0.45cm}P{0.45cm}P{0.45cm}P{0.45cm}P{0.6cm}}
\caption{All models trained.}
\label{tab:all_results} \\
\toprule
model type & arch & notes & $\tau_\text{text}$ & $\tau_\text{image}$ & batch train & batch eval & data seen & 0shot & 10shot & coco i2t & coco t2i & val acc \\ \midrule
\endhead
\multicolumn{13}{c}{\textit{{Figure~\ref{fig:headline}}:} Sweep over scale with CLIP-esque training regime} \\
\midrule
dense & S/32 &  & - & - & 16384 & 1024 & 12.8B & 44.4 & 33.8 & 30.4 & 18.2 & 62.1 \\
\mmoe{} & S/32 &  & $\log(12)$ & $\log(17)$ & 16384 & 1024 & 12.8B & 57.9 & 48.1 & 38.9 & 25.0 & 73.1 \\
dense & S/16 &  & - & - & 16384 & 1024 & 12.8B & 50.3 & 40.4 & 35.8 & 22.7 & 67.6 \\
\mmoe{} & S/16 &  & $\log(9)$ & $\log(20)$ & 16384 & 1024 & 12.8B & 64.5 & 56.3 & 43.7 & 29.2 & 77.1 \\
dense & B/32 &  & - & - & 16384 & 1024 & 12.8B & 58.8 & 48.7 & 42.5 & 27.4 & 72.5 \\
\mmoe{} & B/32 &  & $\log(12)$ & $\log(17)$ & 16384 & 1024 & 12.8B & 67.5 & 60.4 & 45.7 & 31.0 & 79.2 \\
dense & B/16 &  & - & - & 16384 & 1024 & 12.8B & 64.3 & 55.3 & 46.8 & 31.7 & 76.4 \\
\mmoe{} & B/16 &  & $\log(9)$ & $\log(20)$ & 16384 & 1024 & 12.8B & 73.7 & 68.2 & 51.3 & 36.2 & 82.3 \\
dense & L/32 &  & - & - & 16384 & 1024 & 12.8B & 68.1 & 60.7 & 51.2 & 34.6 & 78.5 \\
\mmoe{} & L/32 &  & $\log(20)$ & $\log(1)$ & 16384 & 1024 & 12.8B & 74.6 & 69.7 & 54.5 & 37.2 & 83.3 \\
dense & L/16 &  & - & - & 16384 & 1024 & 12.8B & 73.5 & 67.6 & 54.3 & 38.3 & 82.2 \\
\mmoe{} & L/16 &  & $\log(28)$ & $\log(8)$ & 16384 & 1024 & 12.8B & 78.6 & 74.7 & 55.7 & 39.6 & 85.9 \\
\midrule
\multicolumn{13}{c}{\textit{{Table~\ref{tab:ablations_base}}:} The baselines for many of the ablation experiments below (1T = 1 Tower, 2T = 2 Towers)} \\
\midrule
dense (1T) & B/16 & Trial 0 & - & - & 8192 & 1024 & 819.2M & 49.9 & 43.7 & 37.7 & 23.7 & 66.0 \\
dense (1T) & B/16 & Trial 1 & - & - & 8192 & 1024 & 819.2M & 50.0 & 44.0 & 36.6 & 23.8 & 66.0 \\
dense (1T) & B/16 & Trial 2 & - & - & 8192 & 1024 & 819.2M & 49.5 & 43.6 & 36.0 & 23.6 & 66.0 \\
dense (2T) & B/16 & Trial 0 & - & - & 8192 & 1024 & 819.2M & 54.8 & 47.3 & 41.3 & 26.6 & 69.7 \\
dense (2T) & B/16 & Trial 1 & - & - & 8192 & 1024 & 819.2M & 54.4 & 47.0 & 41.0 & 26.5 & 69.4 \\
dense (2T) & B/16 & Trial 2 & - & - & 8192 & 1024 & 819.2M & 54.9 & 47.1 & 41.6 & 26.9 & 69.5 \\
\mmoe & B/16 & Trial 0 & $\log(9)$ & $\log(20)$ & 8192 & 1024 & 819.2M & 56.8 & 50.5 & 40.1 & 25.7 & 70.8 \\
\mmoe & B/16 & Trial 1 & $\log(9)$ & $\log(20)$ & 8192 & 1024 & 819.2M & 57.0 & 50.4 & 40.4 & 26.2 & 70.8 \\
\mmoe & B/16 & Trial 2 & $\log(9)$ & $\log(20)$ & 8192 & 1024 & 819.2M & 56.9 & 50.6 & 38.5 & 24.9 & 70.7 \\
\midrule
\multicolumn{13}{c}{\textit{{Table~\ref{tab:increase_k}}:} Increasing the number of selected experts, with adjustments to local entropy loss and BPR} \\
\midrule
\mmoe{} & B/16 & k=2, target entropy loss, max BPR & $\log(4)$ & $\log(25)$ & 8192 & 1024 & 819.2M & 55.9 & 48.1 & 36.6 & 25.6 & 69.1 \\
\mmoe{} & B/16 & k=3, target entropy loss, max BPR & $\log(4)$ & $\log(25)$ & 8192 & 1024 & 819.2M & 48.2 & 48.9 & 27.7 & 21.1 & 64.3 \\
\mmoe{} & B/16 & k=5, target entropy loss, max BPR & $\log(4)$ & $\log(25)$ & 8192 & 1024 & 819.2M & 11.7 & 36.4 & 7.1 & 5.6 & 23.2 \\
\mmoe{} & B/16 & k=2, merged loss, max BPR & $\log(4)$ & $\log(25)$ & 8192 & 1024 & 819.2M & 46.4 & 49.4 & 28.1 & 10.7 & 57.3 \\
\mmoe{} & B/16 & k=3, merged loss, max BPR & $\log(4)$ & $\log(25)$ & 8192 & 1024 & 819.2M & 52.6 & 47.9 & 33.0 & 23.2 & 65.5 \\
\mmoe{} & B/16 & k=5, merged loss, max BPR & $\log(4)$ & $\log(25)$ & 8192 & 1024 & 819.2M & 60.3 & 53.4 & 43.3 & 28.0 & 73.3 \\
\mmoe{} & B/16 & k=2, top1 loss, max BPR & $\log(4)$ & $\log(25)$ & 8192 & 1024 & 819.2M & 58.3 & 51.9 & 42.0 & 27.2 & 71.7 \\
\mmoe{} & B/16 & k=3, top1 loss, max BPR & $\log(4)$ & $\log(25)$ & 8192 & 1024 & 819.2M & 59.0 & 53.6 & 42.7 & 28.1 & 72.1 \\
\mmoe{} & B/16 & k=5, top1 loss, max BPR & $\log(4)$ & $\log(25)$ & 8192 & 1024 & 819.2M & 59.8 & 54.6 & 43.0 & 27.8 & 72.5 \\
\mmoe{} & B/16 & k=2, none loss, max BPR & $\log(4)$ & $\log(25)$ & 8192 & 1024 & 819.2M & 46.8 & 44.3 & 28.9 & 14.3 & 61.0 \\
\mmoe{} & B/16 & k=3, none loss, max BPR & $\log(4)$ & $\log(25)$ & 8192 & 1024 & 819.2M & 44.6 & 42.2 & 27.3 & 17.5 & 57.5 \\
\mmoe{} & B/16 & k=5, none loss, max BPR & $\log(4)$ & $\log(25)$ & 8192 & 1024 & 819.2M & 17.5 & 35.4 & 6.1 & 5.9 & 22.8 \\
\mmoe{} & B/16 & k=2, target entropy loss, sum BPR & $\log(4)$ & $\log(25)$ & 8192 & 1024 & 819.2M & 58.2 & 51.8 & 42.2 & 27.7 & 71.6 \\
\mmoe{} & B/16 & k=3, target entropy loss, sum BPR & $\log(4)$ & $\log(25)$ & 8192 & 1024 & 819.2M & 59.1 & 53.2 & 42.3 & 27.5 & 72.5 \\
\mmoe{} & B/16 & k=5, target entropy loss, sum BPR & $\log(4)$ & $\log(25)$ & 8192 & 1024 & 819.2M & 60.4 & 53.8 & 42.1 & 28.0 & 73.0 \\
\mmoe{} & B/16 & k=2, merged loss, sum BPR & $\log(4)$ & $\log(25)$ & 8192 & 1024 & 819.2M & 59.0 & 52.4 & 41.1 & 27.1 & 72.2 \\
\mmoe{} & B/16 & k=3, merged loss, sum BPR & $\log(4)$ & $\log(25)$ & 8192 & 1024 & 819.2M & 60.0 & 52.8 & 42.4 & 27.6 & 73.0 \\
\mmoe{} & B/16 & k=5, merged loss, sum BPR & $\log(4)$ & $\log(25)$ & 8192 & 1024 & 819.2M & 61.0 & 53.6 & 42.7 & 28.4 & 73.4 \\
\midrule
\multicolumn{13}{c}{\textit{{Figure~\ref{fig:total_experts}}:} Increasing the total number of available experts with fixed $k=1$} \\
\midrule
\mmoe{} & B/16 & Total \# experts = 4 & $\log(2.4)$ & $\log(0.8)$ & 8192 & 1024 & 819.2M & 52.3 & 46.9 & 37.8 & 24.4 & 67.9 \\
\mmoe{} & B/16 & Total \# experts = 8 & $\log(4.8)$ & $\log(1.6)$ & 8192 & 1024 & 819.2M & 54.4 & 48.2 & 39.5 & 25.5 & 69.4 \\
\mmoe{} & B/16 & Total \# experts = 16 & $\log(9.6)$ & $\log(3.2)$ & 8192 & 1024 & 819.2M & 55.7 & 49.5 & 38.9 & 25.5 & 70.2 \\
\mmoe{} & B/16 & Total \# experts = 32 & $\log(19.2)$ & $\log(6.4)$ & 8192 & 1024 & 819.2M & 57.3 & 50.4 & 40.1 & 26.0 & 70.9 \\
\mmoe{} & B/16 & Total \# experts = 64 & $\log(38.4)$ & $\log(12.8)$ & 8192 & 1024 & 819.2M & 58.0 & 50.7 & 41.2 & 26.5 & 71.3 \\
\midrule
\multicolumn{13}{c}{\textit{{Figure~\ref{fig:groups}}:} Varying the group size, trading off compute efficiency and stability} \\
\midrule
\mmoe{} & B/16 & Num groups = 1 & $\log(9)$ & $\log(20)$ & 8192 & 1024 & 819.2M & 56.7 & 49.4 & 40.3 & 25.8 & 70.7 \\
\mmoe{} & B/16 & Num groups = 2 & $\log(9)$ & $\log(20)$ & 8192 & 1024 & 819.2M & 56.6 & 50.4 & 40.4 & 25.4 & 70.8 \\
\mmoe{} & B/16 & Num groups = 4 & $\log(9)$ & $\log(20)$ & 8192 & 1024 & 819.2M & 56.1 & 49.5 & 39.1 & 24.8 & 69.8 \\
\mmoe{} & B/16 & Num groups = 8 & $\log(9)$ & $\log(20)$ & 8192 & 1024 & 819.2M & 47.4 & 44.0 & 29.5 & 19.6 & 62.4 \\
\mmoe{} & B/16 & Num groups = 16 & $\log(9)$ & $\log(20)$ & 8192 & 1024 & 819.2M & 50.1 & 45.3 & 33.4 & 21.1 & 65.2 \\
\mmoe{} & B/16 & Num groups = 32 & $\log(9)$ & $\log(20)$ & 8192 & 1024 & 819.2M & 23.8 & 31.7 & 11.6 & 8.3 & 38.9 \\
\mmoe{} & B/16 & Num groups = 64 & $\log(9)$ & $\log(20)$ & 8192 & 1024 & 819.2M & 1.6 & 17.8 & 1.7 & 0.9 & 6.2 \\
\mmoe{} & B/16 & Num groups = 128 & $\log(9)$ & $\log(20)$ & 8192 & 1024 & 819.2M & 0.1 & 38.3 & 0.0 & 0.0 & 0.1 \\
\midrule
\multicolumn{13}{c}{\textit{{Figure~\ref{fig:bpr}}:} Study different alternatives for routing dispatch ordering} \\
\midrule
\mmoe{} & B/16 & Dispatch = shuffle & $\log(9)$ & $\log(20)$ & 8192 & 1024 & 819.2M & 37.6 & 38.5 & 24.2 & 15.1 & 56.5 \\
\mmoe{} & B/16 & Dispatch = image first & $\log(9)$ & $\log(20)$ & 8192 & 1024 & 819.2M & 17.8 & 21.3 & 10.1 & 6.3 & 42.3 \\
\mmoe{} & B/16 & Dispatch = bpr & $\log(9)$ & $\log(20)$ & 8192 & 1024 & 819.2M & 56.8 & 50.5 & 40.1 & 25.7 & 70.8 \\
\mmoe{} & B/16 & Dispatch = bpr & $\log(9)$ & $\log(20)$ & 8192 & 1024 & 819.2M & 57.0 & 50.4 & 40.4 & 26.2 & 70.8 \\
\mmoe{} & B/16 & Dispatch = bpr & $\log(9)$ & $\log(20)$ & 8192 & 1024 & 819.2M & 56.9 & 50.6 & 38.5 & 24.9 & 70.7 \\
\mmoe{} & B/16 & Dispatch = text first & $\log(4)$ & $\log(25)$ & 8192 & 1024 & 819.2M & 0.1 & 1.6 & 0.0 & 0.0 & 3.3 \\
\midrule
\multicolumn{13}{c}{\textit{{Table~\ref{tab:per-modality-router}}:} Variations on the joint, modality agnostic router used for \mmoe{}} \\
\midrule
\mmoe{} & B/16 & Router = per modality & $\log(4)$ & $\log(25)$ & 8192 & 1024 & 819.2M & 56.8 & 50.5 & 40.1 & 25.6 & 70.4 \\
\mmoe{} & B/16 & Router = partitioned & - & - & 8192 & 1024 & 819.2M & 56.8 & 50.1 & 39.1 & 25.1 & 70.8 \\
\midrule
\multicolumn{13}{c}{\textit{{Figure~\ref{fig:balancing}}:} With fixed text seq len 16, vary image seq len to study effect of modality balancing.} \\
\midrule
\mmoe{} & B/12 & Image seq len 324. Losses: classic & - & - & 8192 & 1024 & 819.2M & 40.8 & 42.5 & 23.4 & 15.6 & 53.7 \\
\mmoe{} & B/16 & Image seq len 196. Losses: classic & - & - & 8192 & 1024 & 819.2M & 17.5 & 32.1 & 11.9 & 8.9 & 41.1 \\
\mmoe{} & B/24 & Image seq len 81. Losses: classic & - & - & 8192 & 1024 & 819.2M & 28.9 & 33.4 & 12.3 & 10.5 & 37.5 \\
\mmoe{} & B/32 & Image seq len 49. Losses: classic & - & - & 8192 & 1024 & 819.2M & 29.9 & 29.8 & 12.0 & 8.8 & 39.4 \\
\mmoe{} & B/48 & Image seq len 16. Losses: classic & - & - & 8192 & 1024 & 819.2M & 26.8 & 24.2 & 13.0 & 8.2 & 36.7 \\
\mmoe{} & B/64 & Image seq len 9. Losses: classic & - & - & 8192 & 1024 & 819.2M & 24.8 & 21.2 & 11.7 & 7.7 & 38.3 \\
\mmoe{} & B/12 & Image seq len 324. Losses: entropy & $\log(25)$ & $\log(1)$ & 8192 & 1024 & 819.2M & 58.1 & 50.4 & 40.4 & 27.0 & 72.4 \\
\mmoe{} & B/16 & Image seq len 196. Losses: entropy & $\log(25)$ & $\log(1)$ & 8192 & 1024 & 819.2M & 57.2 & 50.3 & 40.4 & 26.4 & 71.2 \\
\mmoe{} & B/24 & Image seq len 81. Losses: entropy & $\log(25)$ & $\log(1)$ & 8192 & 1024 & 819.2M & 54.1 & 45.4 & 37.3 & 24.0 & 67.5 \\
\mmoe{} & B/32 & Image seq len 49. Losses: entropy & $\log(25)$ & $\log(1)$ & 8192 & 1024 & 819.2M & 50.5 & 41.4 & 35.2 & 21.4 & 65.0 \\
\mmoe{} & B/48 & Image seq len 16. Losses: entropy & $\log(25)$ & $\log(1)$ & 8192 & 1024 & 819.2M & 40.7 & 31.0 & 26.4 & 15.2 & 55.4 \\
\mmoe{} & B/64 & Image seq len 9. Losses: entropy & $\log(25)$ & $\log(1)$ & 8192 & 1024 & 819.2M & 31.8 & 24.6 & 20.6 & 10.9 & 49.9 \\
\midrule
\multicolumn{13}{c}{\textit{{Appendix~\ref{app:global_ent_threshold}}:} Varying global entropy thresholds $\tau_\text{text}$ and $\tau_\text{image}$ independently} \\
\midrule
\mmoe{} & B/16 &  & $\log(1)$ & $\log(1)$ & 8192 & 1024 & 819.2M & 54.4 & 50.8 & 37.1 & 24.4 & 69.1 \\
\mmoe{} & B/16 &  & $\log(1)$ & $\log(4)$ & 8192 & 1024 & 819.2M & 53.9 & 50.8 & 36.8 & 24.1 & 68.9 \\
\mmoe{} & B/16 &  & $\log(1)$ & $\log(8)$ & 8192 & 1024 & 819.2M & 54.0 & 50.9 & 36.7 & 24.1 & 68.8 \\
\mmoe{} & B/16 &  & $\log(1)$ & $\log(12)$ & 8192 & 1024 & 819.2M & 54.3 & 50.7 & 37.6 & 23.8 & 68.6 \\
\mmoe{} & B/16 &  & $\log(1)$ & $\log(20)$ & 8192 & 1024 & 819.2M & 53.8 & 50.5 & 37.0 & 24.1 & 68.5 \\
\mmoe{} & B/16 &  & $\log(1)$ & $\log(28)$ & 8192 & 1024 & 819.2M & 54.2 & 50.8 & 37.1 & 24.1 & 69.1 \\
\mmoe{} & B/16 &  & $\log(1)$ & $\log(32)$ & 8192 & 1024 & 819.2M & 54.2 & 50.5 & 37.3 & 24.2 & 69.1 \\
\mmoe{} & B/16 &  & $\log(2.5)$ & $\log(1)$ & 8192 & 1024 & 819.2M & 52.2 & 50.5 & 33.8 & 19.6 & 67.0 \\
\mmoe{} & B/16 &  & $\log(2.5)$ & $\log(4)$ & 8192 & 1024 & 819.2M & 44.6 & 49.2 & 23.2 & 17.3 & 57.7 \\
\mmoe{} & B/16 &  & $\log(2.5)$ & $\log(8)$ & 8192 & 1024 & 819.2M & 53.5 & 50.9 & 37.3 & 23.8 & 69.6 \\
\mmoe{} & B/16 &  & $\log(2.5)$ & $\log(12)$ & 8192 & 1024 & 819.2M & 16.7 & 46.9 & 23.3 & 14.1 & 49.4 \\
\mmoe{} & B/16 &  & $\log(2.5)$ & $\log(20)$ & 8192 & 1024 & 819.2M & 52.9 & 50.6 & 37.0 & 23.9 & 69.6 \\
\mmoe{} & B/16 &  & $\log(2.5)$ & $\log(28)$ & 8192 & 1024 & 819.2M & 53.4 & 50.6 & 38.9 & 24.5 & 69.7 \\
\mmoe{} & B/16 &  & $\log(2.5)$ & $\log(32)$ & 8192 & 1024 & 819.2M & 28.0 & 46.3 & 12.6 & 8.2 & 48.5 \\
\mmoe{} & B/16 &  & $\log(4)$ & $\log(1)$ & 8192 & 1024 & 819.2M & 45.3 & 47.7 & 28.2 & 20.8 & 64.2 \\
\mmoe{} & B/16 &  & $\log(4)$ & $\log(4)$ & 8192 & 1024 & 819.2M & 55.7 & 50.4 & 39.1 & 25.0 & 70.3 \\
\mmoe{} & B/16 &  & $\log(4)$ & $\log(8)$ & 8192 & 1024 & 819.2M & 47.4 & 47.0 & 31.0 & 18.2 & 63.9 \\
\mmoe{} & B/16 &  & $\log(4)$ & $\log(12)$ & 8192 & 1024 & 819.2M & 21.9 & 47.6 & 18.9 & 18.8 & 55.1 \\
\mmoe{} & B/16 &  & $\log(4)$ & $\log(20)$ & 8192 & 1024 & 819.2M & 27.1 & 45.7 & 15.5 & 17.0 & 51.8 \\
\mmoe{} & B/16 &  & $\log(4)$ & $\log(28)$ & 8192 & 1024 & 819.2M & 39.4 & 47.7 & 18.1 & 14.4 & 49.4 \\
\mmoe{} & B/16 &  & $\log(4)$ & $\log(32)$ & 8192 & 1024 & 819.2M & 43.2 & 47.4 & 19.8 & 15.5 & 54.2 \\
\mmoe{} & B/16 &  & $\log(6.5)$ & $\log(1)$ & 8192 & 1024 & 819.2M & 55.0 & 49.6 & 39.0 & 25.4 & 70.6 \\
\mmoe{} & B/16 &  & $\log(6.5)$ & $\log(4)$ & 8192 & 1024 & 819.2M & 56.6 & 50.3 & 40.3 & 25.1 & 70.7 \\
\mmoe{} & B/16 &  & $\log(6.5)$ & $\log(8)$ & 8192 & 1024 & 819.2M & 55.2 & 50.1 & 39.8 & 25.6 & 70.7 \\
\mmoe{} & B/16 &  & $\log(6.5)$ & $\log(12)$ & 8192 & 1024 & 819.2M & 55.7 & 50.1 & 39.3 & 25.2 & 70.8 \\
\mmoe{} & B/16 &  & $\log(6.5)$ & $\log(20)$ & 8192 & 1024 & 819.2M & 55.4 & 50.0 & 40.2 & 25.2 & 70.7 \\
\mmoe{} & B/16 &  & $\log(6.5)$ & $\log(28)$ & 8192 & 1024 & 819.2M & 56.1 & 50.1 & 39.2 & 25.4 & 70.6 \\
\mmoe{} & B/16 &  & $\log(6.5)$ & $\log(32)$ & 8192 & 1024 & 819.2M & 56.3 & 50.7 & 39.0 & 25.3 & 70.7 \\
\mmoe{} & B/16 &  & $\log(9)$ & $\log(1)$ & 8192 & 1024 & 819.2M & 56.6 & 50.3 & 39.8 & 25.6 & 71.0 \\
\mmoe{} & B/16 &  & $\log(9)$ & $\log(4)$ & 8192 & 1024 & 819.2M & 56.5 & 50.4 & 39.6 & 25.7 & 70.7 \\
\mmoe{} & B/16 &  & $\log(9)$ & $\log(8)$ & 8192 & 1024 & 819.2M & 57.0 & 50.9 & 40.4 & 26.4 & 70.8 \\
\mmoe{} & B/16 &  & $\log(9)$ & $\log(12)$ & 8192 & 1024 & 819.2M & 56.9 & 50.1 & 39.4 & 26.0 & 70.9 \\
\mmoe{} & B/16 &  & $\log(9)$ & $\log(20)$ & 8192 & 1024 & 819.2M & 56.9 & 49.9 & 39.3 & 25.6 & 70.7 \\
\mmoe{} & B/16 &  & $\log(9)$ & $\log(28)$ & 8192 & 1024 & 819.2M & 56.4 & 49.8 & 39.9 & 25.9 & 70.7 \\
\mmoe{} & B/16 &  & $\log(9)$ & $\log(32)$ & 8192 & 1024 & 819.2M & 56.1 & 50.2 & 39.3 & 24.8 & 70.8 \\
\mmoe{} & B/16 &  & $\log(12)$ & $\log(1)$ & 8192 & 1024 & 819.2M & 56.6 & 50.3 & 40.4 & 26.2 & 71.0 \\
\mmoe{} & B/16 &  & $\log(12)$ & $\log(4)$ & 8192 & 1024 & 819.2M & 57.2 & 50.2 & 40.8 & 25.8 & 71.1 \\
\mmoe{} & B/16 &  & $\log(12)$ & $\log(8)$ & 8192 & 1024 & 819.2M & 57.1 & 50.1 & 41.1 & 25.5 & 71.2 \\
\mmoe{} & B/16 &  & $\log(12)$ & $\log(12)$ & 8192 & 1024 & 819.2M & 56.7 & 50.1 & 40.1 & 25.2 & 71.2 \\
\mmoe{} & B/16 &  & $\log(12)$ & $\log(20)$ & 8192 & 1024 & 819.2M & 57.7 & 49.6 & 39.7 & 25.3 & 70.8 \\
\mmoe{} & B/16 &  & $\log(12)$ & $\log(28)$ & 8192 & 1024 & 819.2M & 57.0 & 50.2 & 39.9 & 26.0 & 71.1 \\
\mmoe{} & B/16 &  & $\log(12)$ & $\log(32)$ & 8192 & 1024 & 819.2M & 57.3 & 49.6 & 40.7 & 25.9 & 70.8 \\
\mmoe{} & B/16 &  & $\log(24)$ & $\log(1)$ & 8192 & 1024 & 819.2M & 57.1 & 50.4 & 40.2 & 25.8 & 71.3 \\
\mmoe{} & B/16 &  & $\log(24)$ & $\log(4)$ & 8192 & 1024 & 819.2M & 57.1 & 49.8 & 39.9 & 25.8 & 70.9 \\
\mmoe{} & B/16 &  & $\log(24)$ & $\log(8)$ & 8192 & 1024 & 819.2M & 58.3 & 49.5 & 40.0 & 26.1 & 71.1 \\
\mmoe{} & B/16 &  & $\log(24)$ & $\log(12)$ & 8192 & 1024 & 819.2M & 57.1 & 50.0 & 41.1 & 25.9 & 70.9 \\
\mmoe{} & B/16 &  & $\log(24)$ & $\log(20)$ & 8192 & 1024 & 819.2M & 57.1 & 50.3 & 39.8 & 25.4 & 71.1 \\
\mmoe{} & B/16 &  & $\log(24)$ & $\log(28)$ & 8192 & 1024 & 819.2M & 58.0 & 50.1 & 40.5 & 26.0 & 71.2 \\
\mmoe{} & B/16 &  & $\log(24)$ & $\log(32)$ & 8192 & 1024 & 819.2M & 58.0 & 50.1 & 41.2 & 25.7 & 71.0 \\
\mmoe{} & B/16 &  & $\log(32)$ & $\log(1)$ & 8192 & 1024 & 819.2M & 57.0 & 49.8 & 39.3 & 25.3 & 71.2 \\
\mmoe{} & B/16 &  & $\log(32)$ & $\log(4)$ & 8192 & 1024 & 819.2M & 57.2 & 49.9 & 40.4 & 25.9 & 71.1 \\
\mmoe{} & B/16 &  & $\log(32)$ & $\log(8)$ & 8192 & 1024 & 819.2M & 57.4 & 50.6 & 39.2 & 25.8 & 71.1 \\
\mmoe{} & B/16 &  & $\log(32)$ & $\log(12)$ & 8192 & 1024 & 819.2M & 56.9 & 50.2 & 41.3 & 26.4 & 71.0 \\
\mmoe{} & B/16 &  & $\log(32)$ & $\log(20)$ & 8192 & 1024 & 819.2M & 57.5 & 50.5 & 39.9 & 26.0 & 71.1 \\
\mmoe{} & B/16 &  & $\log(32)$ & $\log(28)$ & 8192 & 1024 & 819.2M & 57.3 & 50.1 & 40.7 & 26.0 & 71.0 \\
\mmoe{} & B/16 &  & $\log(32)$ & $\log(32)$ & 8192 & 1024 & 819.2M & 57.3 & 49.8 & 41.1 & 26.3 & 71.0 \\
\midrule
\multicolumn{13}{c}{\textit{{Table~\ref{tab:laion}}:} Training on publically available LAION400M data.} \\
\midrule
dense & B/16 & Trial 0 & - & - & 16384 & 1024 & 1.4B & 56.1 & 47.9 & 43.0 & 27.9 & 96.6 \\
dense & B/16 & Trial 1 & - & - & 16384 & 1024 & 1.4B & 56.0 & 47.7 & 42.8 & 27.6 & 96.5 \\
dense & B/16 & Trial 2 & - & - & 16384 & 1024 & 1.4B & 55.8 & 47.5 & 42.5 & 27.9 & 96.6 \\
\mmoe{} & B/16 & Trial 0 & $\log(9)$ & $\log(20)$ & 16384 & 1024 & 1.4B & 61.1 & 54.4 & 44.1 & 28.9 & 97.9 \\
\mmoe{} & B/16 & Trial 1 & $\log(9)$ & $\log(20)$ & 16384 & 1024 & 1.4B & 60.9 & 54.4 & 43.5 & 28.7 & 97.9 \\
\mmoe{} & B/16 & Trial 2 & $\log(9)$ & $\log(20)$ & 16384 & 1024 & 1.4B & 61.1 & 54.1 & 43.6 & 29.0 & 97.9 \\
\end{longtable}
}
\end{widepage}

%% file: appendix/4_practical_aspects.tex
\section{Computational costs of \mmoe{}}

\subsection{Unimodal evaluation with multimodal experts}
Recall that each expert has a capacity $C$ - it can process at most $C$ tokens, and if it is assigned more, those above $C$ will not be processed.
This capacity is usually set relative to some `ideal'. If there are $N$ tokens and $E$ experts, we usually assume each expert can handle at most $C_R\times\frac{N}{E}$ tokens, where $C_R \ge 1$ is a slack factor.
This way we try to reach a balanced setup where most expert process a similar number of tokens.

Multimodal routing presents a unique issue here. During training, the model learns to balance tokens when it has both images \textit{and} text available to it. When there is only one modality, it will not use all the experts due to natural emergence of modality-specific experts - but the expert capacity size will be set \textit{assuming all experts are used}. This results in high rates of token dropping, depending on the ratio of modality-specific experts.

In this effort, we encounter this during zero-shot classification and retrieval; models first compute representations for all text tokens, and then separately for all image tokens. In order to get around this token dropping, we simply evaluate with a high slack factor $C_R = 16$.

There are however other natural solutions; for many circumstances, one could trivially restructure evaluation such that image and text inputs are processed simultaneously. A more interesting, MoE specific solution is \textit{pruning} modality specific experts, which is explored and shown to work in~\ref{app:xp_pruning}. \mmoe{} models could have been evaluated at a `normal' capacity, with pruned experts.

\subsection{Understanding the compute costs of \mmoe}
\label{app:profiling_discussion}
Zero-shot evaluation on ImageNet with 6 prompts requires 6000 text forward passes and 50000 image forward passes. With 80 prompts, a la CLIP~\cite{radford2021clip}, it is 80000 for text instead. How does one compare compute cost vs. performance?
The costs of \mmoe{}, its dense baselines, and other two-tower models, were computed assuming a full batch of images and texts, as this is the approach which makes the least assumptions about the downstream setup. This does not generalise perfectly: if, for example, a particular use case processed very large numbers of texts but only few images, models with smaller/cheaper text towers would be clearly advantaged.

\begin{figure}
     \centering
     \begin{subfigure}[b]{\textwidth}
         \centering
         \includegraphics[width=\textwidth, bb=0 0 915 224]{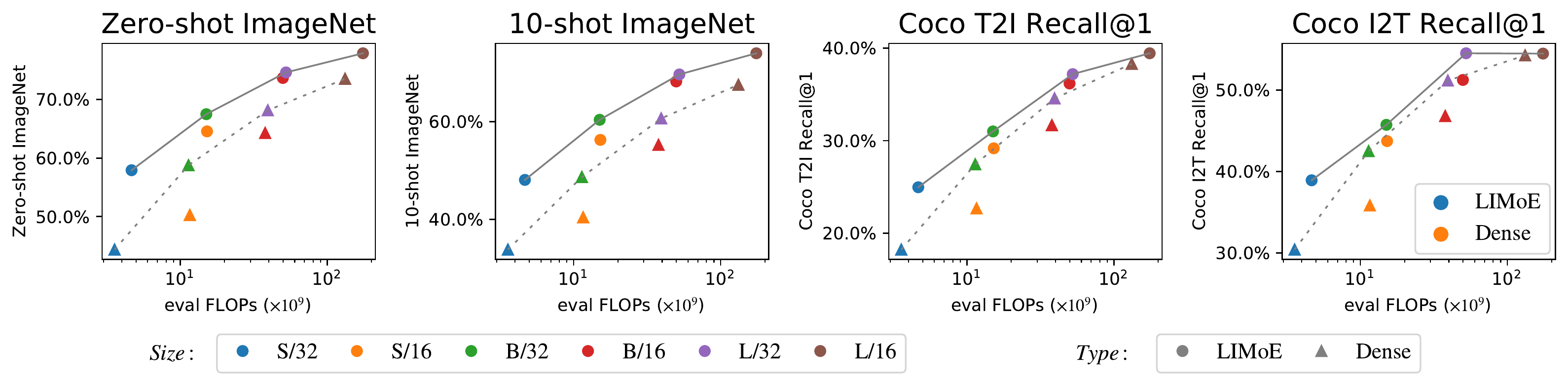}
         \caption{Inference FLOPs}
         \label{fig:pareto_eval_flops}
     \end{subfigure}
     \hfill
     \begin{subfigure}[b]{\textwidth}
         \centering
         \includegraphics[width=\textwidth, bb=0 0 915 227]{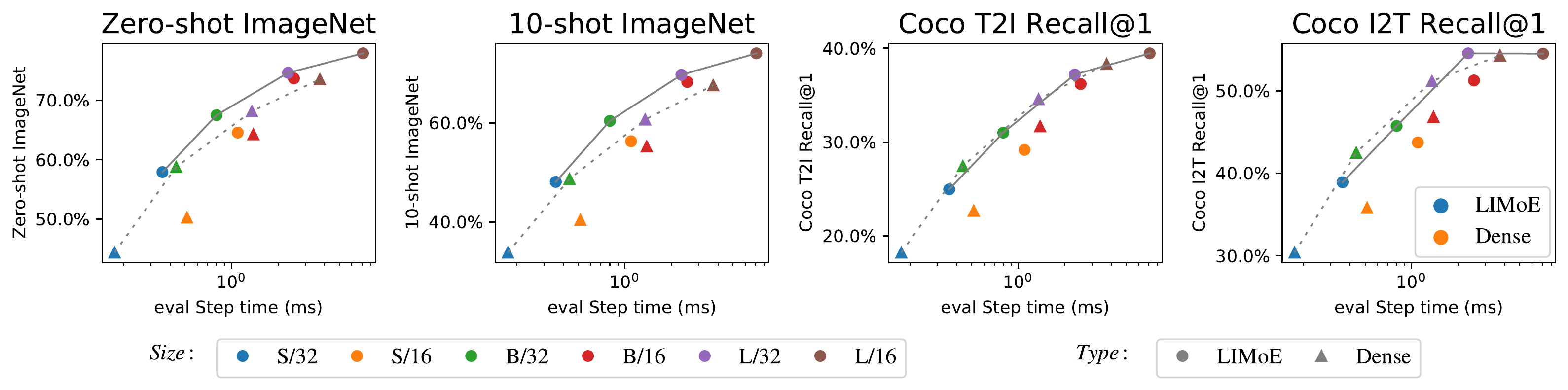}
         \caption{Inference step time}
         \label{fig:pareto_eval_time}
     \end{subfigure}
     \hfill
     \begin{subfigure}[b]{\textwidth}
         \centering
         \includegraphics[width=\textwidth, bb=0 0 915 224]{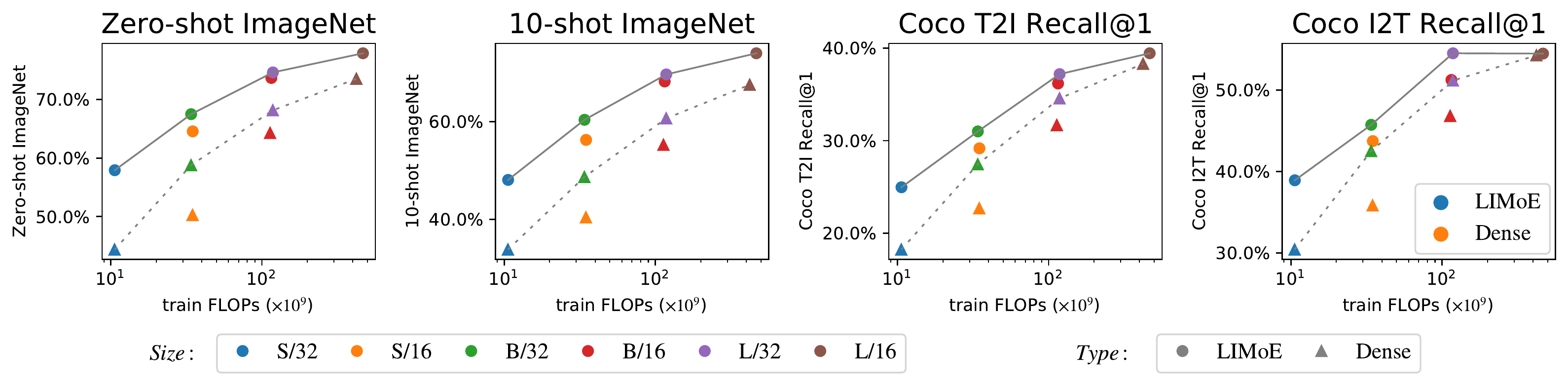}
         \caption{Train FLOPs}
         \label{fig:pareto_train_flops}
     \end{subfigure}
     \hfill
     \begin{subfigure}[b]{\textwidth}
         \centering
         \includegraphics[width=\textwidth, bb=0 0 915 227]{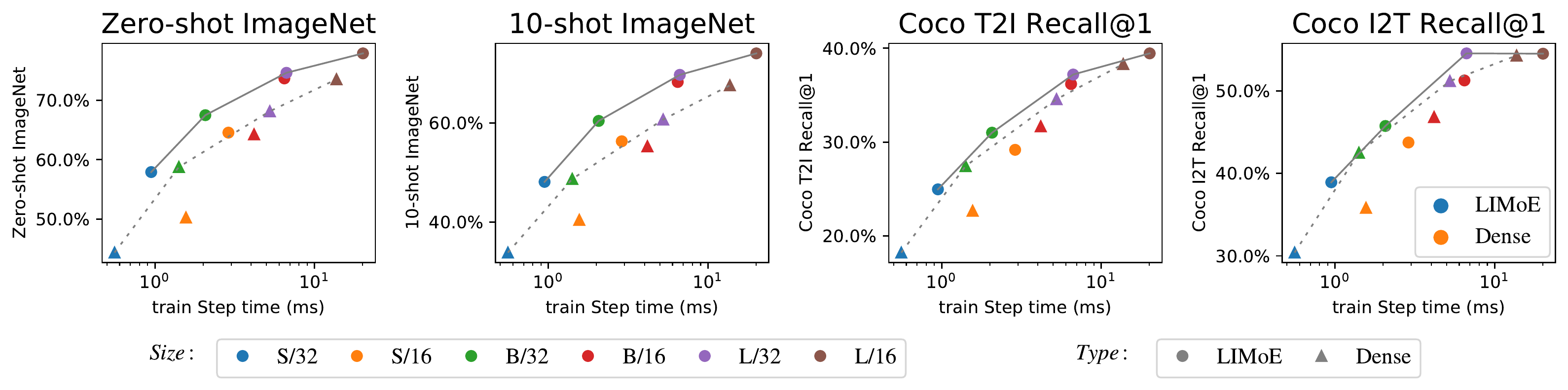}
         \caption{Train step time}
         \label{fig:pareto_train_time}
     \end{subfigure}
    \caption{Pareto frontiers with respect to different measures of computational cost}
    \label{fig:paretos}
\end{figure}

\textbf{Full profiling data for Section~\ref{sec:experiments}}. For training and evaluation, we used a variety of TPU versions. For consistency, we profiled computation times on a TPUv3 (\texttt{v3-32} to be more precise\footnote{\url{https://cloud.google.com/tpu/docs/types-topologies}}). Figure~\ref{fig:paretos} shows performance with respect to different proxies for compute cost.
As discussed in Section~\ref{sec:experiments}, \mmoe{} is clearly pareto optimal with respect to total FLOPs. However, this does not fully account for certain costs related to MoE models, such as cross-device communication. {Figures~\ref{fig:pareto_eval_time} and ~\ref{fig:pareto_train_time}} show the performance with respect to step time. With respect to zeroshot and 10-shot classification accuracy, the performance improvements of \mmoe{} are significant enough that it is still clearly pareto optimal; for retrieval metrics on COCO, \mmoe{}'s gains exactly justify the costs, and it is not significantly more efficient than dense baselines. The story is similar whether looking at train or evaluation cost.

%% file: appendix/5_experiment_overflow.tex
\section{Further experiments}
\label{app:xp}

In this section, we present further ablations  not included in the main text due to space constraints.

\subsection{Increasing the number of selected experts}
\label{app:xp_increasing_k}
All models in this paper select $K=1$ expert per token to match the cost of a dense backbone. 

There are two main challenges with increasing $K$:

\textit{Modifications to auxiliary losses.} The local entropy loss effectively encourages that router choices are one-hot. When increasing $K$, the model is still incentivized to only use 1 expert, assigning other experts weights near 0, thereby effectively behaving as $K=1$.

We try two modifications to the local loss to ameliorate this:
\begin{itemize}[leftmargin=1em]
    \item \textit{Target entropy}: Encourage the local entropy to be $\log{K}$ -- at least a uniform distribution over $K$ experts -- instead of 0: we minimize  $\Omega_\text{target}^\text{K} = (\log(K) - \Omega_\text{local}(\mG_m))^2$.
    \item \textit{Merged entropy}: We sum the top $K$ and the bottom $N - K$ routing probabilities to give a binomial distribution, and optimize this to have entropy 0. This encourages the routing weight to all be in the top $K$ experts, but does not care exactly how it is distributed among them.
\end{itemize}

% Awkwardly placed wrapfig
\begin{wrapfigure}[16]{r}{0.37\textwidth}
  \begin{center}
  \vspace{-0.75cm}
  \includegraphics[width=0.37\textwidth, bb=0 0 290 262]{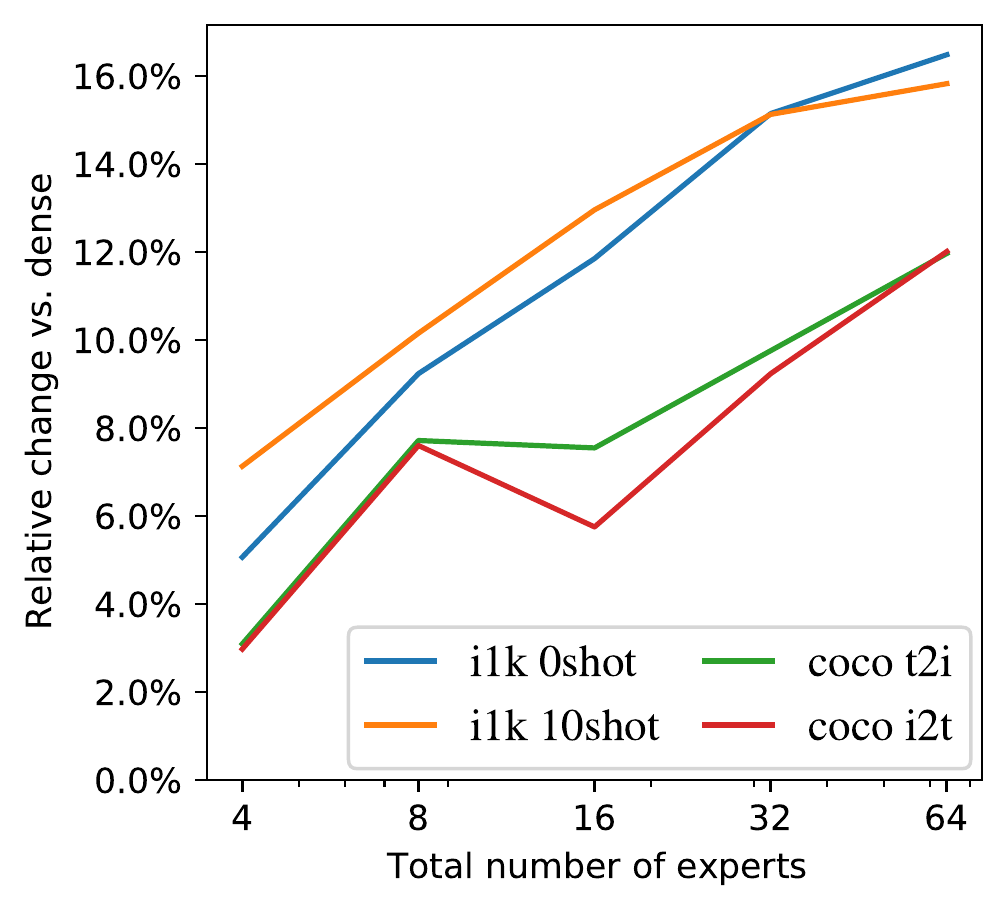}
  \caption{\textbf{Increasing the total number of experts} consistently improves model performance; all are better than the dense baseline.}
  \label{fig:total_experts}
  \end{center}
\end{wrapfigure}

\textit{BPR modifications} With these losses, the router uses  $K > 1$ experts per token. However, BPR prioritises tokens according to their max routing probability, which decreases when probabilities are distributed over $K$ choices. The stabilisation effect BPR provides training is consequently lost. We alter it to prioritise tokens by the sum of top $K$ probabilities. In vision tasks, the two approaches perform identically~\cite{riquelme2021vmoe}, but here the latter stabilises training and unlocks $K > 1$.

Table~\ref{tab:increase_k} shows the final results. Without changes to the local entropy loss (BPR score = max, local entropy method = default), there are some improvements which stem from using $K > 1$ for image tokens - the local loss on text means it is effectively using $K = 1$ for text anyway. Without the modifications to the BPR score, the modifications to the local loss can result in fairly unstable models.
Once the BPR score is modified, we see consistent improvements with increasing $K$, particularly with the 'Merged' variant.

\begin{table}
\caption{Increasing number of selected experts improves performance with appropriate changes to auxiliary losses and BPR. Table entries are ImageNet zero-shot accuracy in \%.}
\label{tab:increase_k}
\centering
\begin{tabular}{r@{\hskip 2em}c@{ }c@{ }c@{ }c@{\hskip 2em}c@{ }c}
\toprule
       & \multicolumn{4}{c}{BPR score = max} & \multicolumn{2}{c}{BPR score = sum} \\
\textit{Local entropy method} &            None &                    Default & Target Ent &  Merged &      Target Ent &  Merged \\
\midrule
 $K = 1$ &               &  55.5\ci{53.3}{57.7} &          &       &               &         \\
 $K = 2$ &          46.8 &                 58.3 &     55.9 &  46.4 &          58.2 &  59.0 \\
 $K = 3$ &          44.6 &                 59.0 &     48.2 &  52.6 &          59.1 &  60.0 \\
 $K = 5$ &          17.5 &                 59.8 &     11.7 &  60.3 &          60.4 &  61.0 \\
\bottomrule
\end{tabular}
\end{table}

\subsection{Increasing the total number of experts}
\label{app:xp_increase_num_experts}
There is thus far no consensus on the optimal number of experts in MoEs; early NLP research scaled to 1000s of experts~\cite{shazeer2017outrageously,fedus2022switch}, before reducing to 32 or 64~\cite{zoph2022stmoe}, which is the standard setup for vision~\cite{riquelme2021vmoe}. In Figure~\ref{fig:total_experts}, we vary this for \mmoe{}, and show that larger expert pools yield consistent performance improvements.

\subsection{Router design choice}
\label{app:xp_router_design}
Recall the router is simply a dense layer; by default we have a \textit{joint} router for all tokens, independent of modality, with no constraints on gating. We consider two other options:
\begin{itemize}
    \item \textit{Per-modality router.} We consider modality-dependent routers which can leverage knowledge of token modality to improve performance (that is, one router for image tokens, and a different one for text tokens). They both output routing distributions over a shared pool of experts, similar to prior works have per-task routers for multitask learning~\cite{ma2018mmoe}.
    \item \textit{Disjoint experts and routers.} We define separate pools of image and text experts. This way, image tokens can only go to a set of experts $\mathcal{E}_{\text{img}}$, and text tokens can only be assigned to another set of experts $\mathcal{E}_{\text{txt}}$. In principle, these sets may or may not intersect. In Table 7, we report results when the sets are indeed disjoint.
\end{itemize}
The results in Table~\ref{tab:per-modality-router} show the three approaches lead to comparable performance. 
In general, the disjoint setup was more stable, and did not need entropy regularisation as per-modality balance/independence is enforced by design.
While convenient and well-behaved here, this approach may not be as general for the case with dozens of tasks and modalities.

\begin{table}
\caption{A simple routing setup without modality-specific adjustments is competitive with specialized approaches. 0shot and 10shot columns show accuracy (\%), t2i and i2t show recall@1 (\%).}
\label{tab:per-modality-router}
\centering
\begin{tabular}{rcccc}
\toprule
\textit{routers} &                  i1k 0shot &                 i1k 10shot &                   coco t2i &                   coco i2t \\
\midrule
joint        &  56.9\ci{56.7}{57.2} &  50.5\ci{49.0}{52.0} &  25.5\ci{16.8}{34.3} &  39.5\ci{27.1}{51.8} \\
per modality &                 56.8 &                 50.5 &                 25.6 &                     40.1 \\
disjoint experts (5 for text, 27 for image)  &                 56.8 &                 50.1 &                 25.1 &                     39.1 \\
\bottomrule
\end{tabular}
\end{table}

\subsection{Pruning Multimodal Experts}
\label{app:xp_pruning}
During training we track what fraction of each modality's tokens went to each expert. It is therefore trivial to identify which experts are processing predominately text and which are processing predominately images. We show here that this information can be trivially used to prune experts for single-modality forward passes, demonstrating on two 32-expert \mmoe-S/16 models: one trained with global text entropy threshold $\tau_\text{text} = \log(4)$, and one with $\tau_\text{text} = \log(9)$.

\begin{figure}[h]
  \centering
  \includegraphics[width=0.4\textwidth, bb=0 0 288 254]{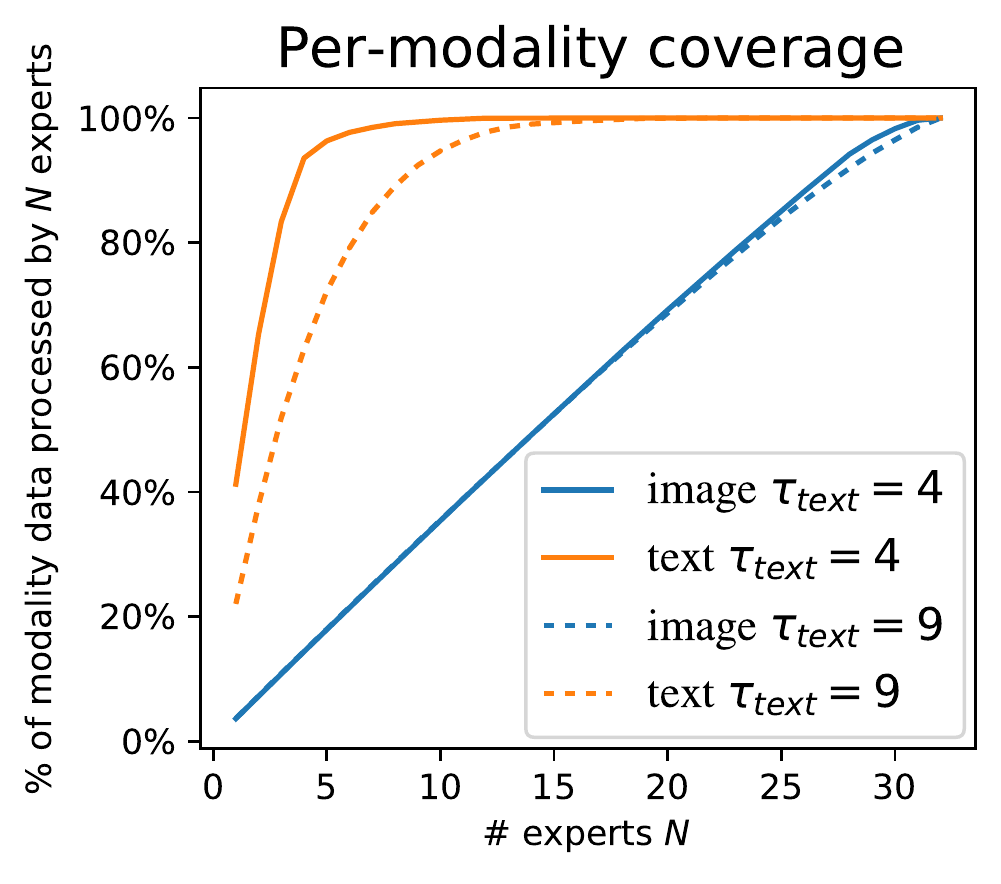}
  \caption{\textbf{Per-modality coverage}; for each modality, we progressively prune the least important experts. The coverage shows the percentage of router top-1 predictions which are still serviceable with the remaining experts. For text (orange), many experts can be pruned, but that is not the case for images. The global entropy threshold $\tau$ controls the prunability, as it encourages use of at least $\log(\tau)$ experts.}
  \label{fig:prune_coverage}
\end{figure}

\textbf{Choosing what to prune}.
Note that we separately choose what experts to prune per-modality; we use text as an illustrative example. Pruning is simple: For each MoE layer, we rank experts according to the fraction of text tokens they processed during training (we average over the last 2500 steps with measurements sampled every 50 steps). We then start pruning according to the one that processed the least tokens, and so on.
Figure~\ref{fig:prune_coverage} shows how the coverage of different modalities changes as experts are pruned. Following the relationship between the global text entropy threshold and the idea of the `soft minimum', we see that around $e^{\tau_\text{text}}$ text experts are needed to process the majority of text tokens; e.g. with $\tau_\text{text} = 4$ for a single-modality forward pass, 28 experts could be comfortably pruned.
Image experts are more distributed; almost all the experts are needed to process all image tokens, as expected.

\textbf{How to run \mmoe \ inference with fewer experts}.
While some experts are pruned, the model is not further trained to adapt to this new situation. One must therefore think carefully on the best way to apply models with a subset of experts.
The router predicts $p(\texttt{expert}|\vx)$. The top-$K$ experts are activated, and the output of the expert layer is the weighted average of the expert outputs. The weighting used for expert $i$ is the \textit{unnormalized} $p(\texttt{expert}_i | \vx)$. This is important, as removing some of the experts and their logits modifies the concentration of $p(\texttt{expert}|\vx)$, and could result in expert weights higher than those used at training time.

\begin{figure}[h]
  \centering
  \includegraphics[width=0.4\textwidth, bb=0 0 280 236]{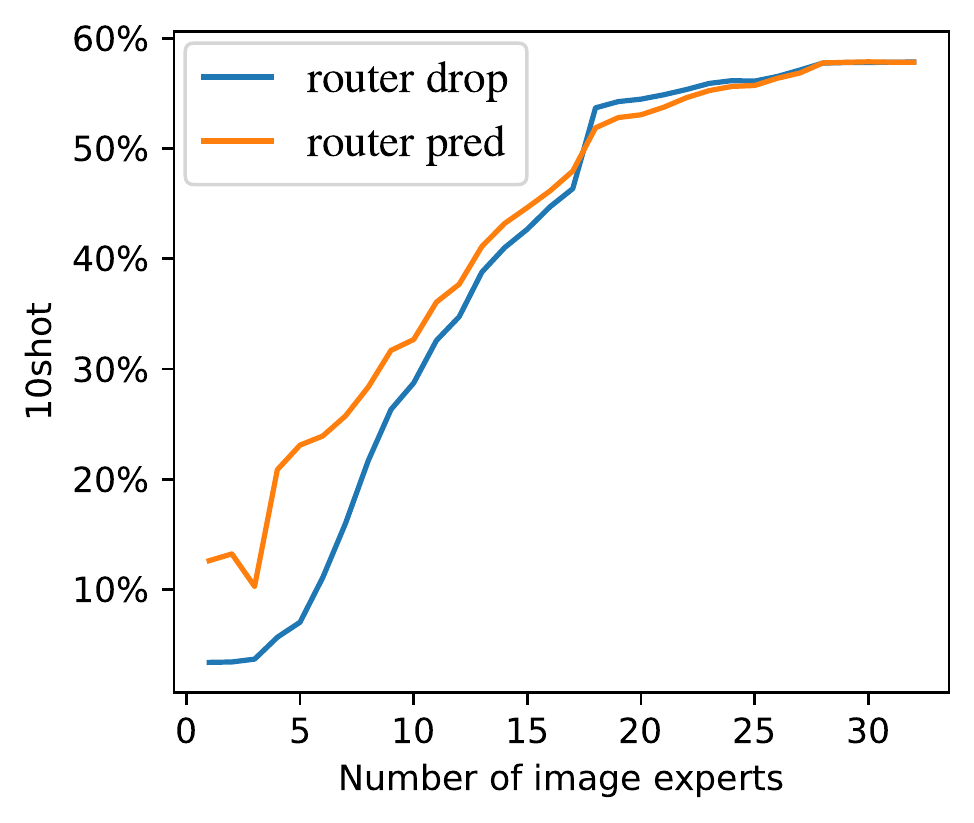}
  \caption{It is better to predict weights for pruned experts and mask them out \textit{after} the softmax.}
  \label{fig:prune_method}
\end{figure}

When removing some of the experts, there are therefore two natural options:
\begin{enumerate}
    \item \texttt{router-drop}: Completely remove the experts from the router. The softmax for $p(\texttt{expert}|\vx)$ will be computed over a subset of experts, thereby adjusting the weights as discussed above.
    \item \texttt{router-pred}: The router still predicts probabilities for pruned experts. However, it is unable to actually use the pruned experts; the top-$K$ operation will ignore those that are unavailable. This preserves the original scaling the model was trained with.
\end{enumerate}
The two approaches are naturally very similar if very few experts are removed. Illustrating with ImageNet-10shot (linear few-shot evaluation), Figure~\ref{fig:prune_method} compares the two options. When a large number of experts are pruned, the \texttt{router-pred} is significantly better, but if only a few experts are pruned, they both perform similarly.

\begin{figure}[h]
  \centering
  \includegraphics[width=0.8\textwidth, bb=0 0 583 278]{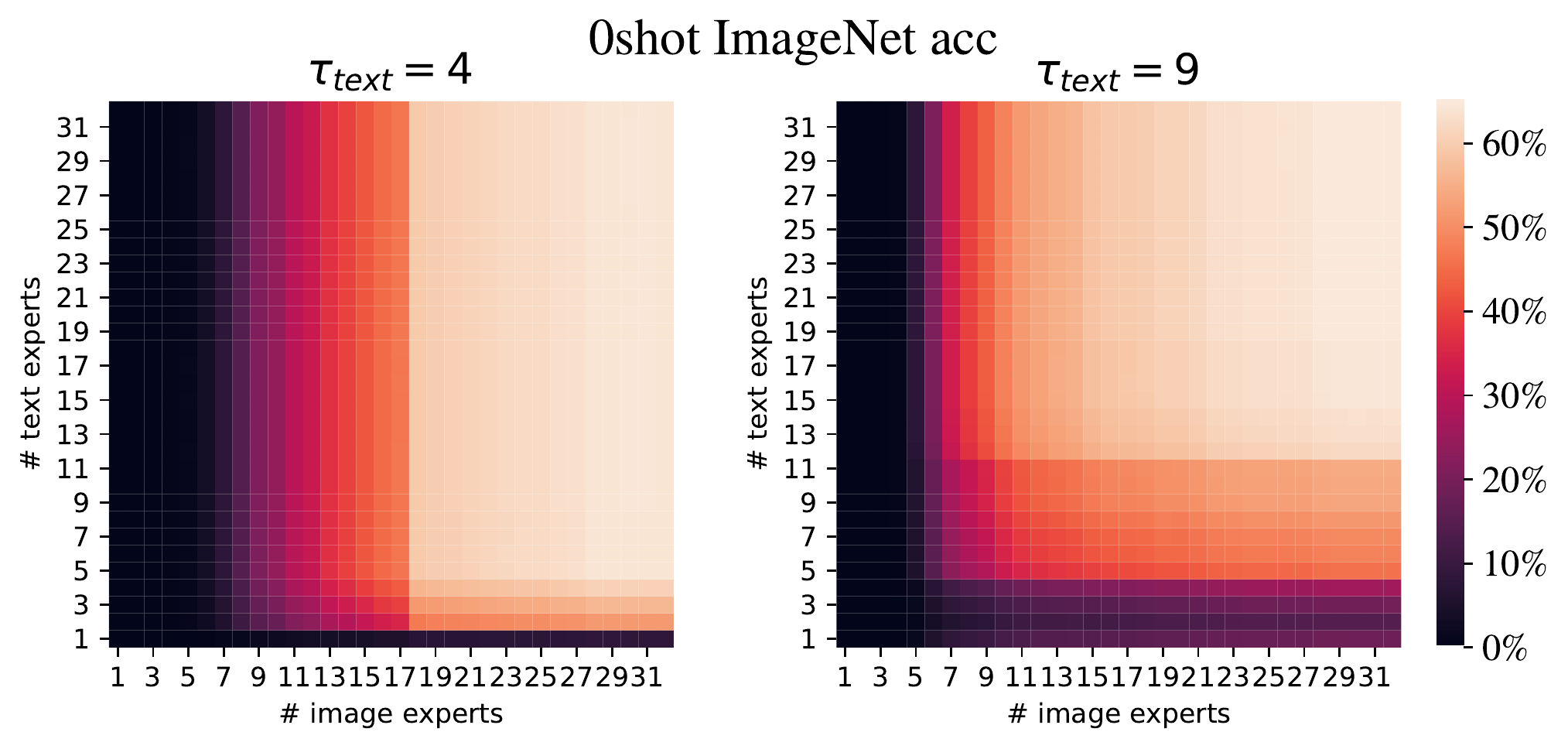}
  \caption{The impact of pruning on ImageNet zero-shot accuracy, comparing two \mmoe-S/16 models trained with different global text entropy thresholds.}
  \label{fig:prune_performance}
\end{figure}

\paragraph{The effect of pruning on performance}
Figure~\ref{fig:prune_performance} shows the impact of pruning image and text experts on zero-shot ImageNet accuracy. Recall that image and text inputs are processed independently for this evaluation, and so the experts used for each modality can be independently pruned.

As expected, we can prune down to only 4 experts during text evaluation without significantly harming performance. On the other hand, the less pruning of image experts, the better.

\subsection{Grouped routing}
\label{app:xp_grouped_routing}
\begin{figure}[h]
  \centering
  \includegraphics[width=\textwidth, bb=0 0 976 194]{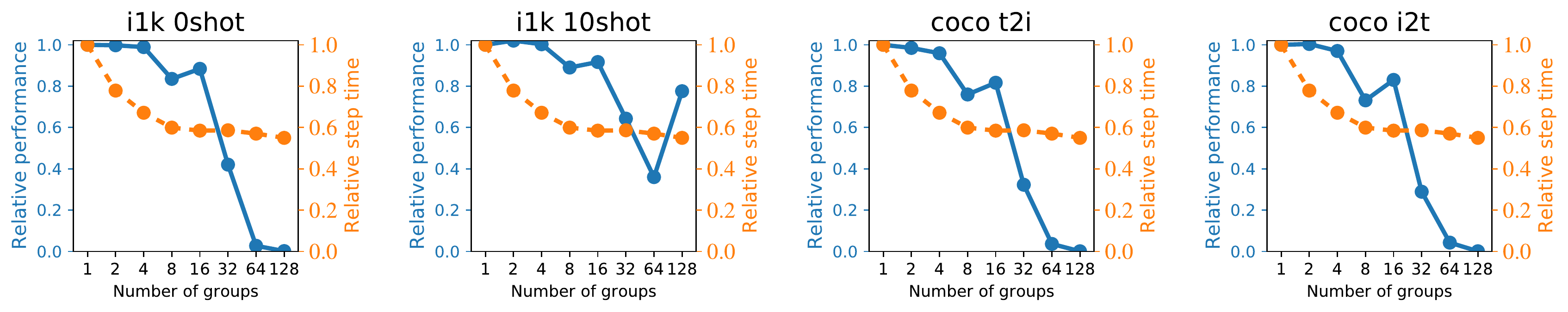}
  \caption{\textbf{Grouped routing can reduce step time (orange)}, but too much becomes unstable.
  }
  \label{fig:groups}
\end{figure}

Splitting batches into groups before dispatching can reduce routing cost significantly, which depending on implementation can scale $\sim\mathcal{O}(\texttt{num tokens}^2)$. There are two sources of potential issues though: in our implementation, auxiliary losses are computed in each group then averaged. The necessary batch-wise statistics become less reliable with more numerous, smaller groups.  Secondly, with smaller groups, it is more likely to get an almost homogenous batch, which makes distributing across experts harder.
To study this, we sweep the group size in a parallel setup with 128 examples per device. Group size 1 means processing and dispatching $128 \times (196 + 16) = 27136$ tokens at once, whereas e.g. group size 8 involves splitting into 8 groups of 3392 tokens. Figure~\ref{fig:groups} shows the effect of this; up to 4 groups, performance is good, but any more than that and training becomes unstable, harming performance. This is more fragile than image-only routing, where group sizes as small as 400 are stable (equivalent to $\sim 68$ groups here). Nonetheless, with 4 groups, step time is reduced by 30\%, capturing 75\% of potential efficiency gains from grouped routing.

\subsection{Experiments on public data}
\label{app:xp_laion}
In order to ascertain \mmoe{}'s efficacy on public data, and reproducibility, we train B/16 models on LAION-400M~\cite{laion400m}. We train for 5 epochs at batch size 16,384. Table~\ref{tab:laion} shows the outcome of three trials, compared against a dense baseline. Once again, we see significant improvements performance, especially in ImageNet zero-shot (+5.0\% absolute, +8.9\% relative) and 10-shot (+6.6\% absolute, +13.8\% relative) performance.

\begin{table}
\caption{\mmoe{} performance on LAION400M, against a dense baseline, three trials.}
\label{tab:laion}
\centering
\begin{tabular}{lllll}
\toprule
       &               10shot &                0shot &             COCO t2i &             COCO i2t \\
\midrule
Dense &  47.7\ci{47.3}{48.1} &  56.0\ci{55.6}{56.3} &  27.8\ci{27.4}{28.3} &  42.8\ci{42.1}{43.4} \\
 \mmoe{} &  54.3\ci{53.9}{54.7} &  61.0\ci{60.7}{61.4} &  28.9\ci{28.5}{29.2} &  43.7\ci{42.9}{44.6} \\
\bottomrule
\end{tabular}

\end{table}

%% file: appendix/6_analysis.tex
\section{Model Analysis}
\label{app:analysis}

\subsection{Routing Distributions}
\label{app:analysis_routing_distributions}

In this section, we explore how routing is distributed across different layers, experts, and modalities.
In particular, we focus on which tokens are dropped.
We analyze two models, B/32 and B/16, each with 8 experts.
This way we can appreciate the impact of having a significantly different ratio of text:image tokens.
Moreover, the global entropy targets $S = e^\tau$ for (text, image) tokens are (3, 25) and (6, 6) for the B/32 and B/16 models, respectively.

\begin{figure}[h]
  \centering
  \includegraphics[width=1.0\textwidth, bb=0 0 855 563]{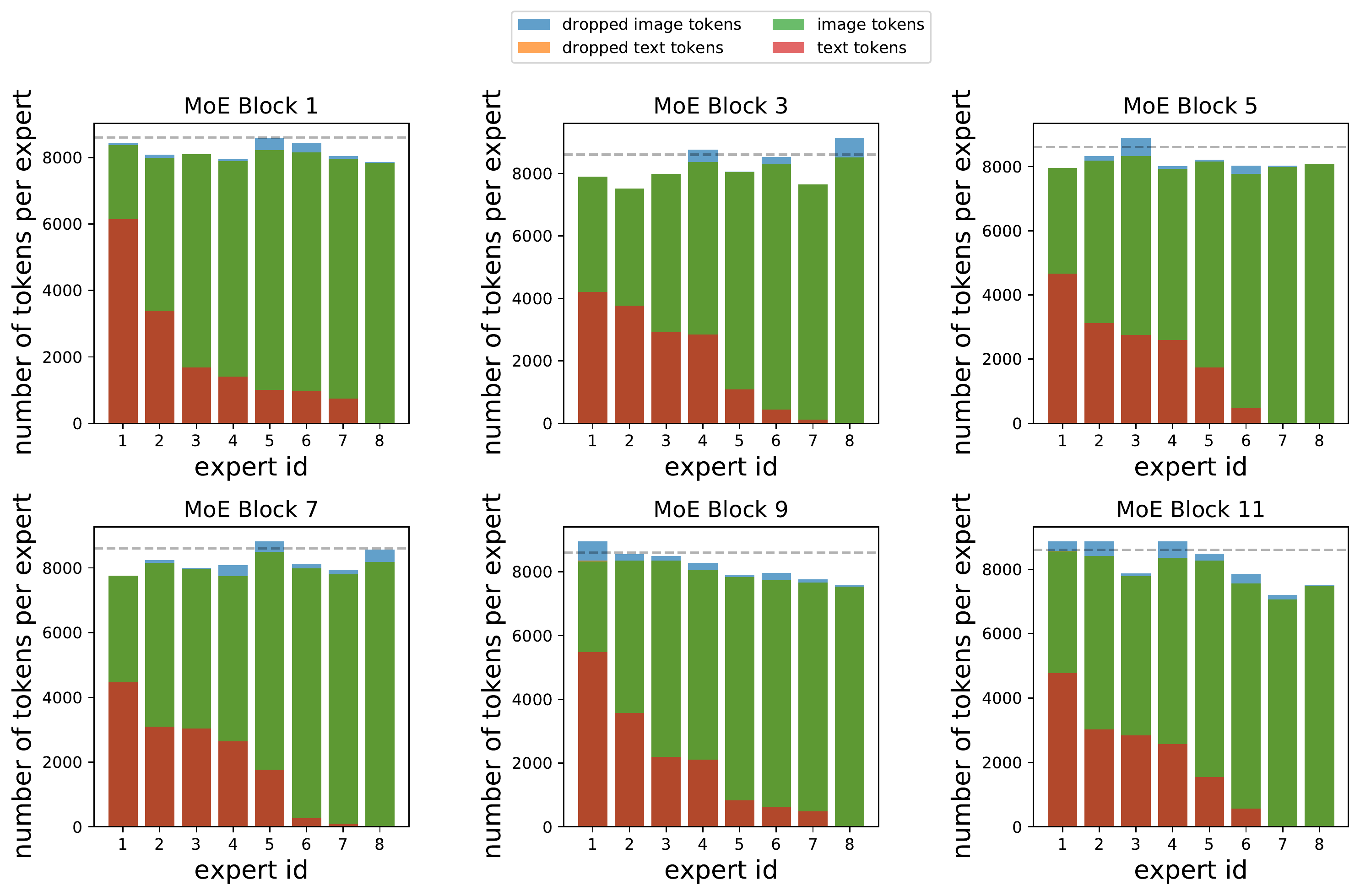}
  \caption{\textbf{Token Distribution for training data.} B/32 model with 8 experts. We display utilization and dropping for a forward pass with batch size 1024. The discontinuous line represents the maximum capacity per expert. Note that we enforce capacity locally per device, so some tokens may not be able to be dispatched even within global capacity constraints. We observe very little token dropping as this is the training data for which auxiliary losses lead to balance.}
  \label{fig:token_distribution_argus_b32}
\end{figure}

\begin{figure}[h]
  \centering
  \includegraphics[width=1.0\textwidth, bb=0 0 855 563]{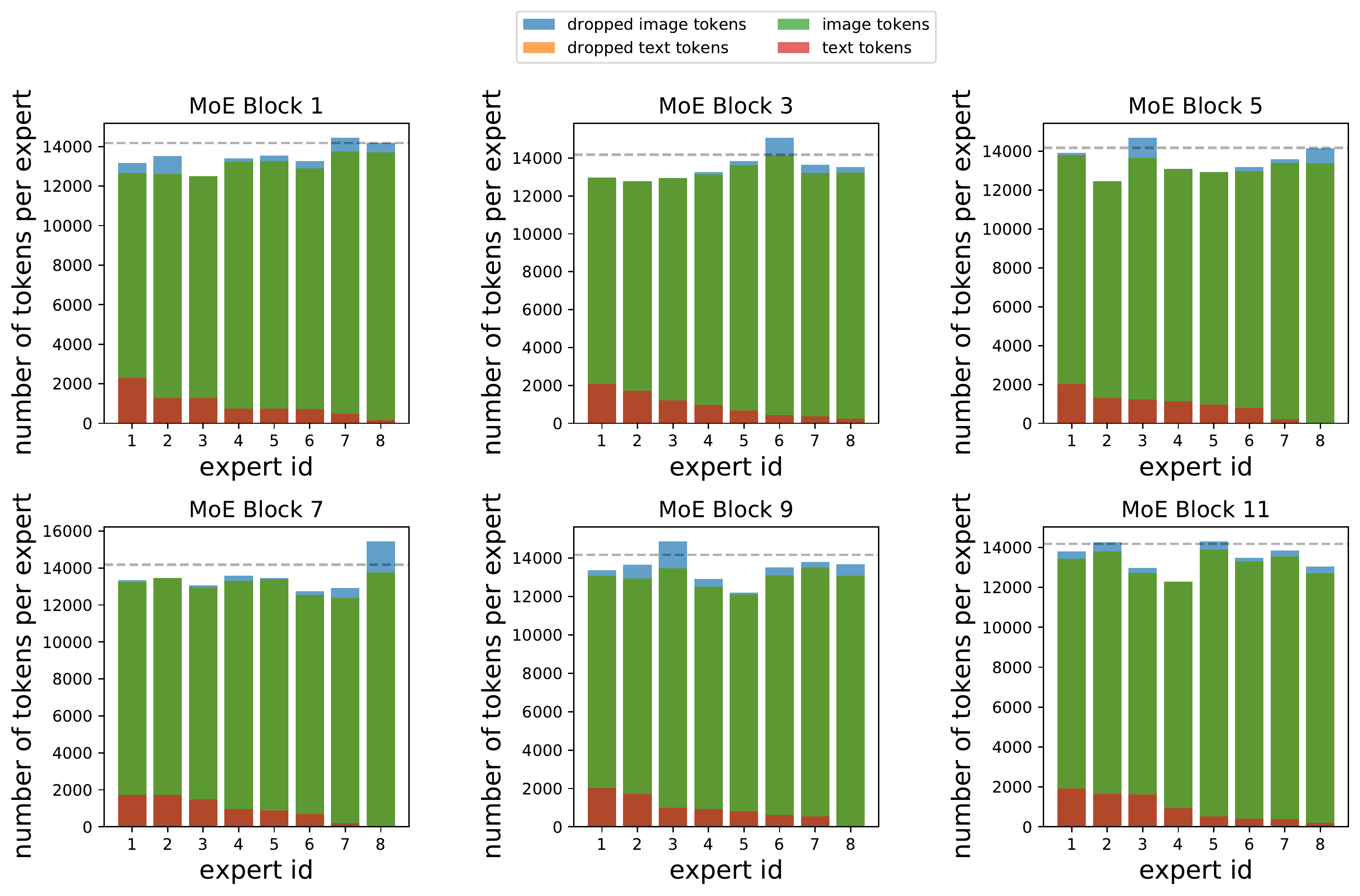}
  \caption{\textbf{Token Distribution for training data.} B/16 model with 8 experts. We display utilization and dropping for a forward pass with batch size 512. The discontinuous line represents the maximum capacity per expert. Note that we enforce capacity locally per device, so some tokens may not be able to be dispatched even within global capacity constraints. We observe very little token dropping as this is the training data for which auxiliary losses lead to balance.
  Compared to Figure~\ref{fig:token_distribution_argus_b32}, we can see how text tokens generally represent a quite small fraction of the in-flow for every expert.}
  \label{fig:token_distribution_argus_b16}
\end{figure}

We first show the routing distributions under the training distribution in Figures~\ref{fig:token_distribution_argus_b32} and~\ref{fig:token_distribution_argus_b16}.
In both cases --as expected-- routing works very well.
Moreover, most experts handle both image and text tokens.

\begin{figure}[h]
  \centering
  \includegraphics[width=1.0\textwidth, bb=0 0 855 563]{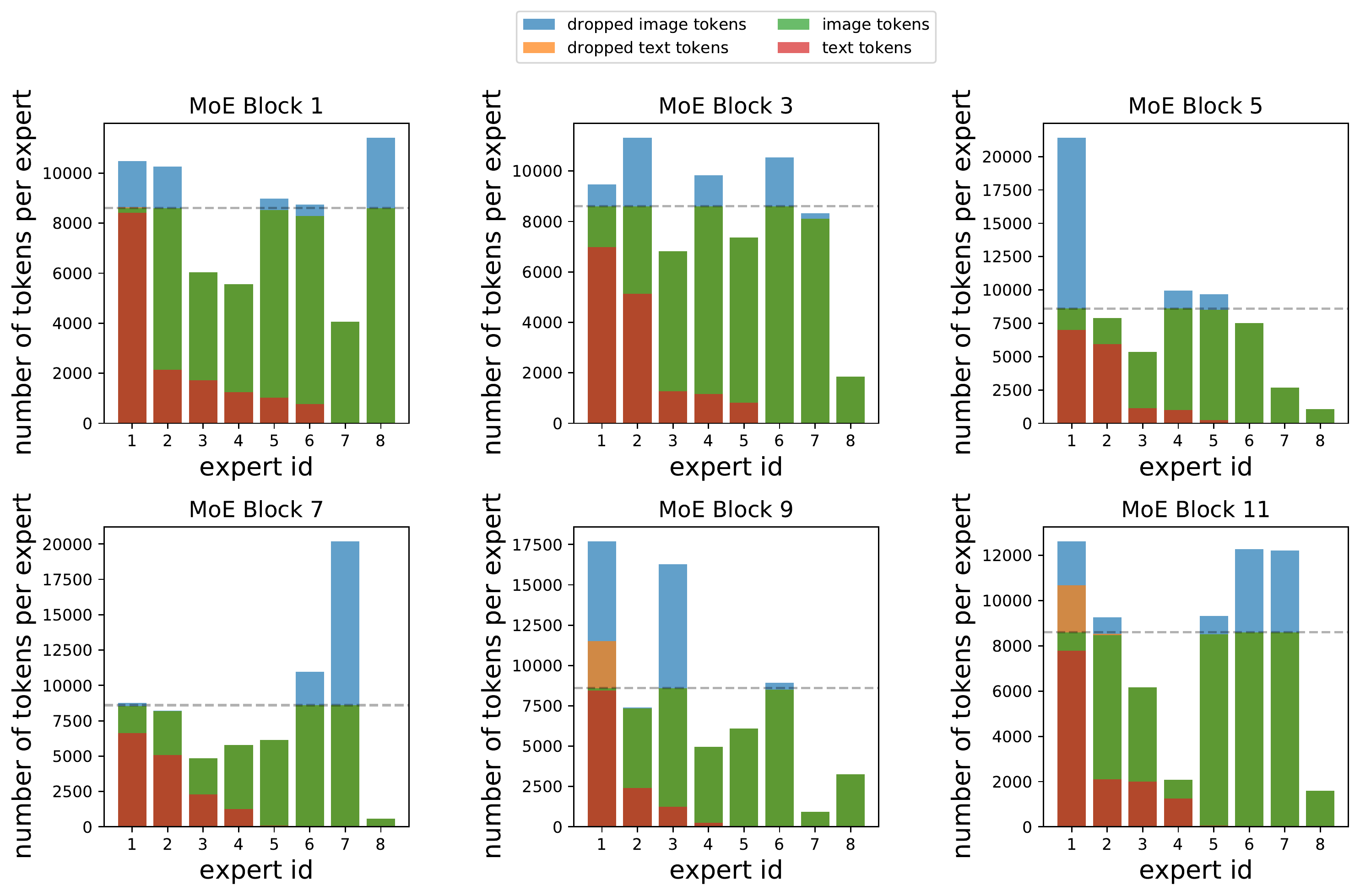}
  \caption{\textbf{Token Distribution for COCO data.} B/32 model with 8 experts. We display utilization and dropping for a forward pass with batch size 1024. The discontinuous line represents the maximum capacity per expert. Note that we enforce capacity locally per device, so some tokens may not be able to be dispatched even within global capacity constraints. Compared to Figure~\ref{fig:token_distribution_argus_b32}, in this case, as there is a distribution shift --while no further training or finetuning--, we see distributions of tokens per expert becoming fairly unbalanced. Moreover, a non-trivial amount of tokens are dropped (above discontinuous horizontal line). Even text tokens are dropped sometimes, and some experts --like Expert 1 in the MoE Block 1 or 9-- end up only processing text tokens.}
  \label{fig:token_distribution_coco_b32}
\end{figure}

\begin{figure}[h]
  \centering
  \includegraphics[width=1.0\textwidth, bb=0 0 855 563]{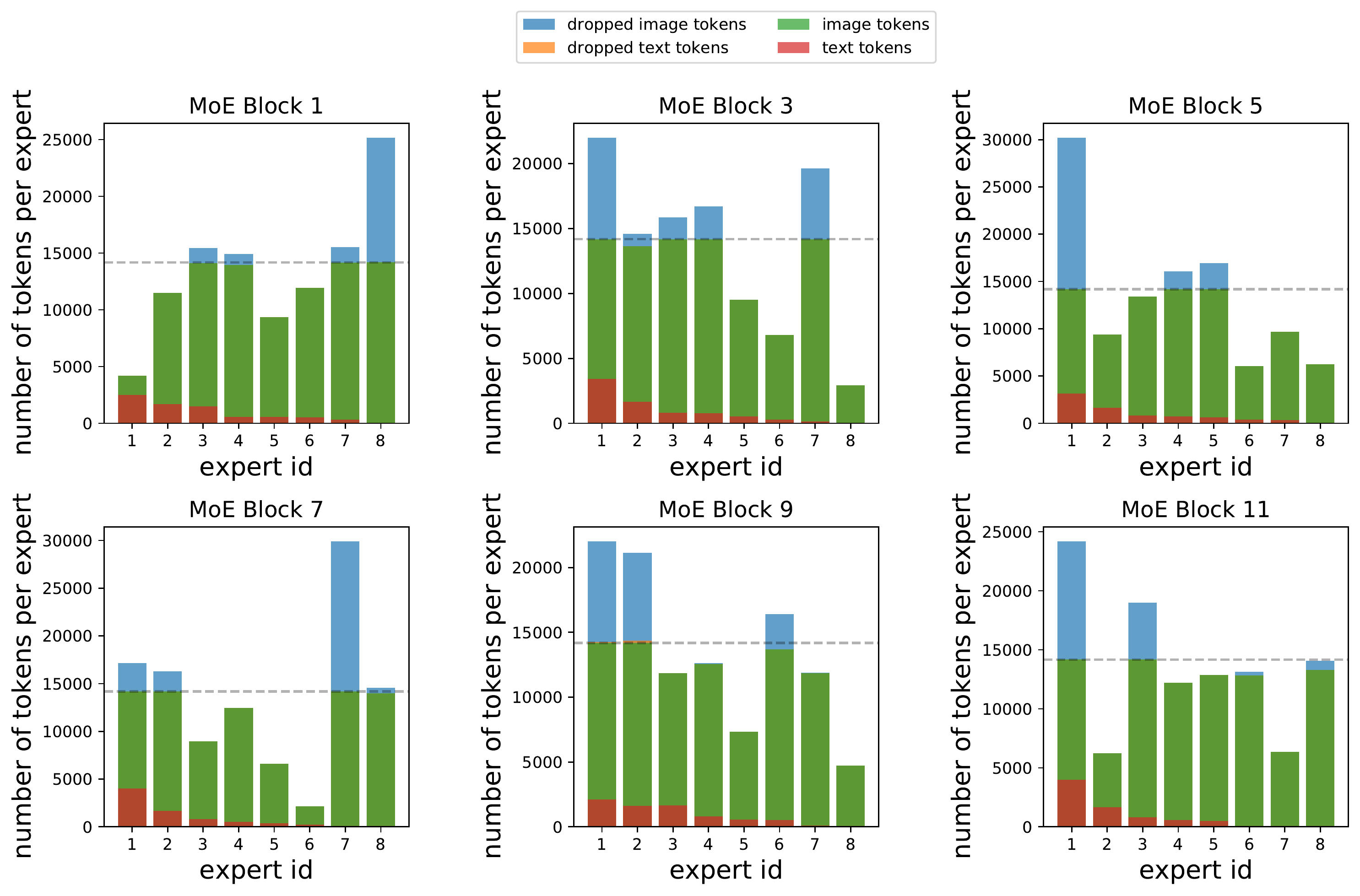}
  \caption{\textbf{Token Distribution for COCO data.} B/16 model with 8 experts. We display utilization and dropping for a forward pass with batch size 512. The discontinuous line represents the maximum capacity per expert. Note that we enforce capacity locally per device, so some tokens may not be able to be dispatched even within global capacity constraints. Compared to Figure~\ref{fig:token_distribution_argus_b16}, in this case, as there is a distribution shift --while no further training or finetuning--, we see distributions of tokens per expert becoming fairly unbalanced. Moreover, a non-trivial amount of tokens are dropped (above discontinuous horizontal line). Text tokens are still mostly processed as BPR shields them via their high priorities and they still represent a small percentage of the tokens.}
  \label{fig:token_distribution_coco_b16}
\end{figure}

\begin{figure}[h]
  \centering
  \includegraphics[width=1.0\textwidth, bb=0 0 855 563]{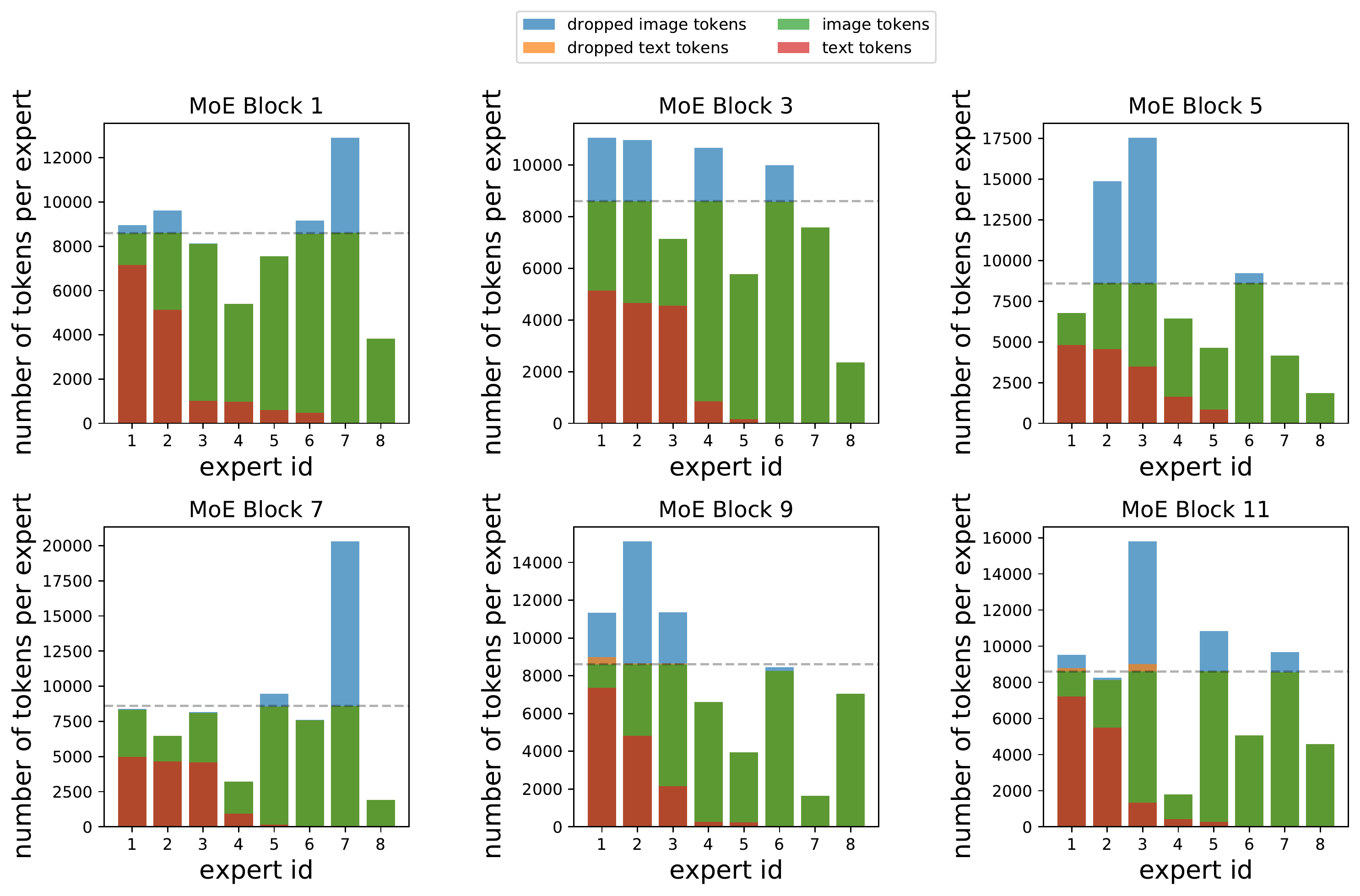}
  \caption{\textbf{Token Distribution for ImageNet data.} B/32 model with 8 experts. We display utilization and dropping for a forward pass with batch size 1024. The discontinuous line represents the maximum capacity per expert. Note that we enforce capacity locally per device, so some tokens may not be able to be dispatched even within global capacity constraints. Compared to Figure~\ref{fig:token_distribution_argus_b32}, in this case, as there is a distribution shift --while no further training or finetuning--, we see distributions of tokens per expert becoming fairly unbalanced. Moreover, a non-trivial amount of tokens are dropped (above discontinuous horizontal line). Very few text tokens are dropped (there is a significant amount of padding, and prompt tokens that are probably processed with very high confidence scores by BPR).}
  \label{fig:token_distribution_imagenet_b32}
\end{figure}

\begin{figure}[h]
  \centering
  \includegraphics[width=1.0\textwidth, bb=0 0 855 563]{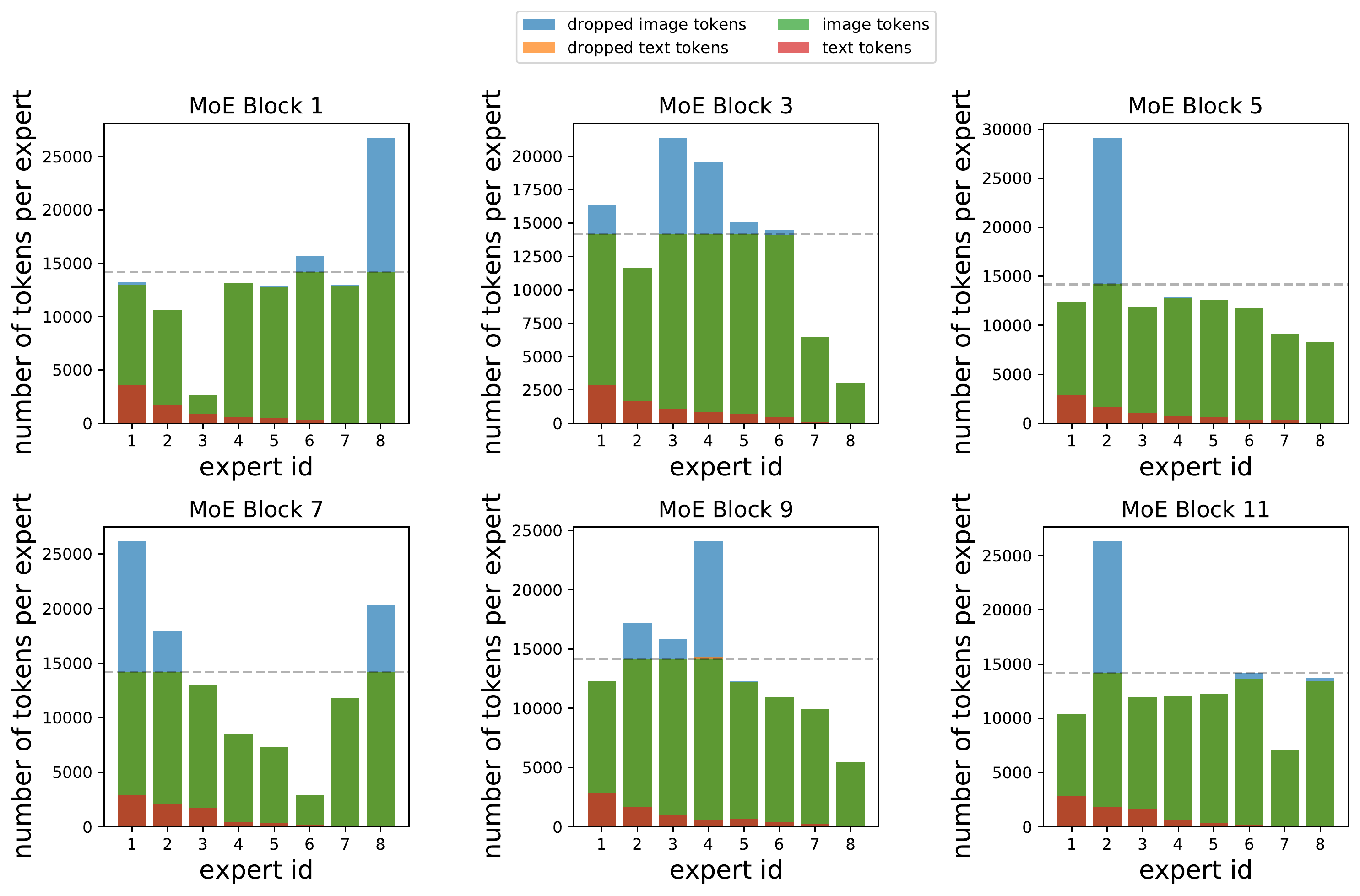}
  \caption{\textbf{Token Distribution for ImageNet data.} B/16 model with 8 experts. We display utilization and dropping for a forward pass with batch size 512. The discontinuous line represents the maximum capacity per expert. Note that we enforce capacity locally per device, so some tokens may not be able to be dispatched even within global capacity constraints. Compared to Figure~\ref{fig:token_distribution_argus_b16}, in this case, as there is a distribution shift --while no further training or finetuning--, we see distributions of tokens per expert becoming fairly unbalanced. Moreover, a non-trivial amount of tokens are dropped (above discontinuous horizontal line). Almost no text tokens are dropped (there is a significant amount of padding, and prompt tokens that are probably processed with very high confidence scores by BPR).}
  \label{fig:token_distribution_imagenet_b16}
\end{figure}

\clearpage
\subsection{Routing Examples}
\label{app:analysis_routing_examples}

In this section, we share practical examples of image and text token routing on the B/32 and B/16 models introduced at the beginning of the section.
All evaluations are on ImageNet (that is, not on the training data).
While the number of experts is clearly smaller than the number of different semantic concepts in images and text, we still highlight some cool patterns in most experts -- especially in the context of images, as text tokens tend to use a reduced number of experts.
We show some of the patches with the highest routing confidence, as analyzing all the thousands of patches that are assigned to each expert is difficult.
However, we expect many other semantic concepts present in the training data to be almost exclusively served by individual experts.

\begin{figure}[h]
  \centering
  \includegraphics[width=0.95\textwidth, bb=0 0 1361 716]{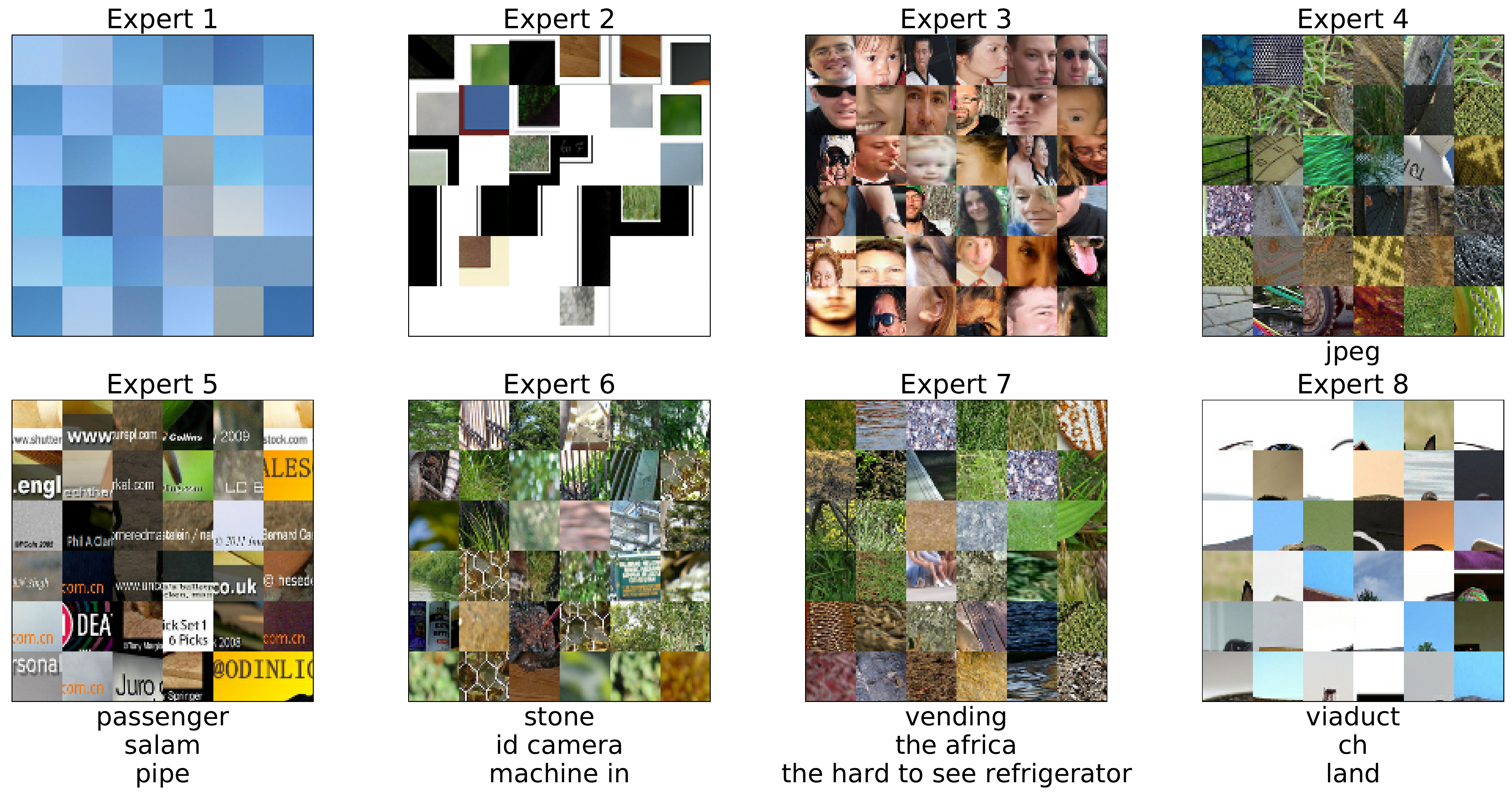}
  \caption{\textbf{Token routing for Imagenet.} B/32 model with 8 experts, we show some of the original tokens (both image and text) as routed at the second MoE layer (corresponds to the fourth encoder).}
  \label{fig:token_routing_imagenet_full}
\end{figure}

\begin{figure}[h]
  \centering
  \includegraphics[width=1.0\textwidth, bb=0 0 1335 716]{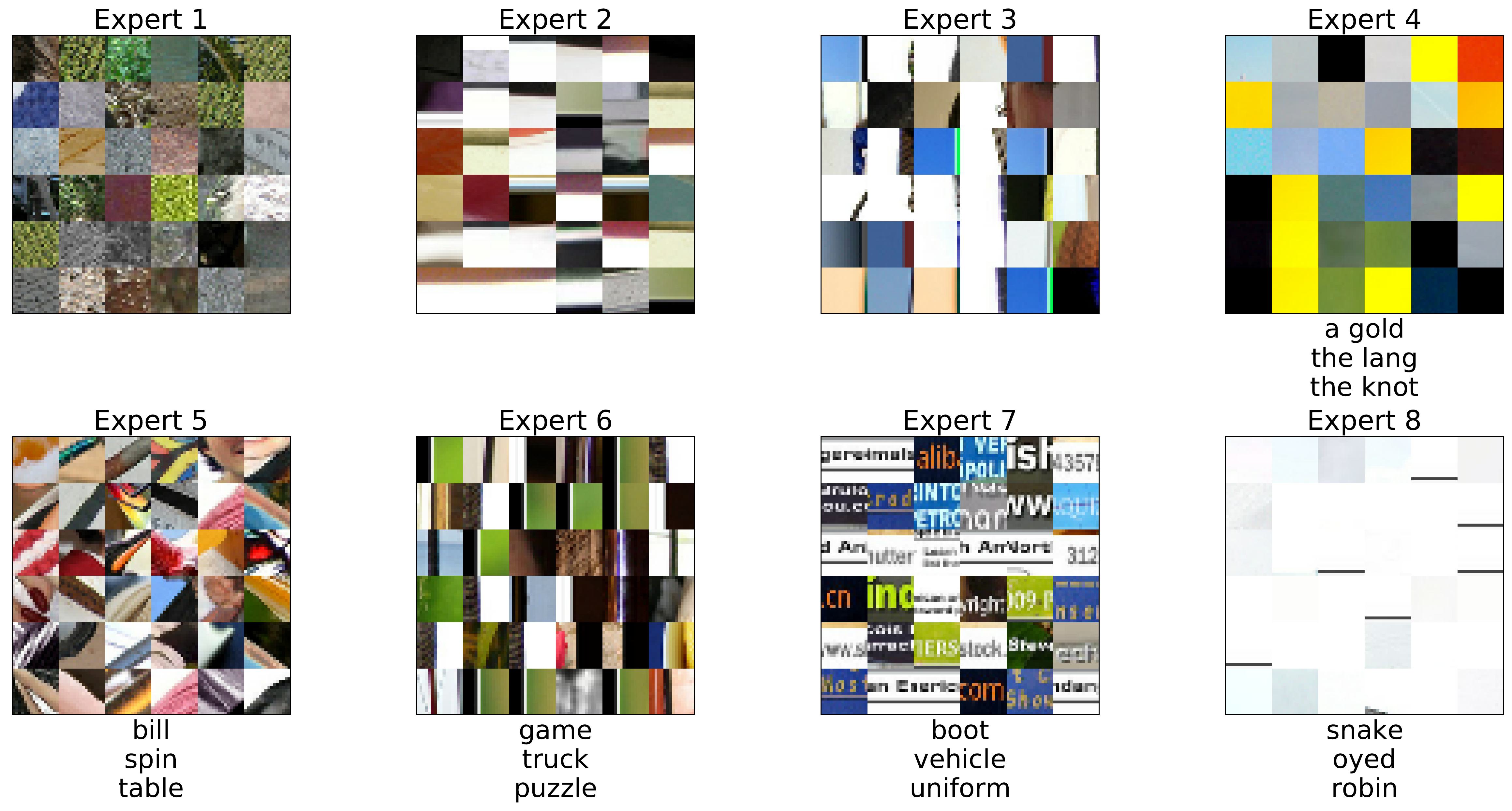}
  \caption{\textbf{Token routing for Imagenet.} B/16 model with 8 experts, we show original tokens (both image and text) as routed at the first MoE layer (corresponds to the second encoder block).}
  \label{fig:token_routing_imagenet_b16_lyr_1}
\end{figure}

\begin{figure}[h]
  \centering
  \includegraphics[width=1.0\textwidth, bb=0 0 1335 716]{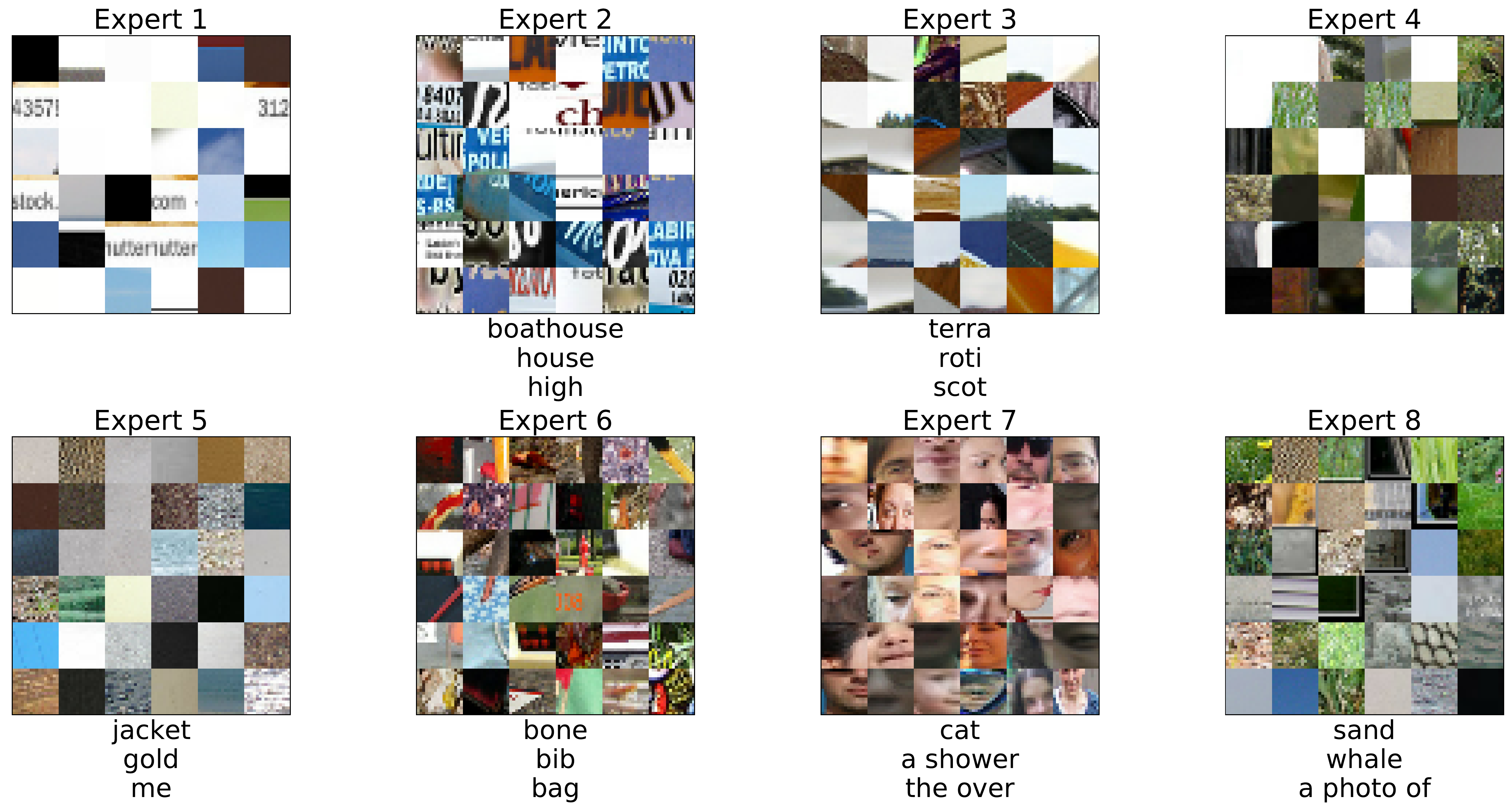}
  \caption{\textbf{Token routing for Imagenet.} B/16 model with 8 experts, we show original tokens (both image and text) as routed at the second MoE layer (corresponding to the fourth encoder block).}
  \label{fig:token_routing_imagenet_b16_lyr_3}
\end{figure}

\clearpage
\subsection{Routing for Individual Inputs}

In this subsection, we show the expert split for a specific given input -- image and text.
Recall tokens from different modalities do not interact in the forward pass (other than via sharing expert capacity).

\begin{figure}[h]
  \centering
  \includegraphics[width=1.0\textwidth, bb=0 0 1072 590]{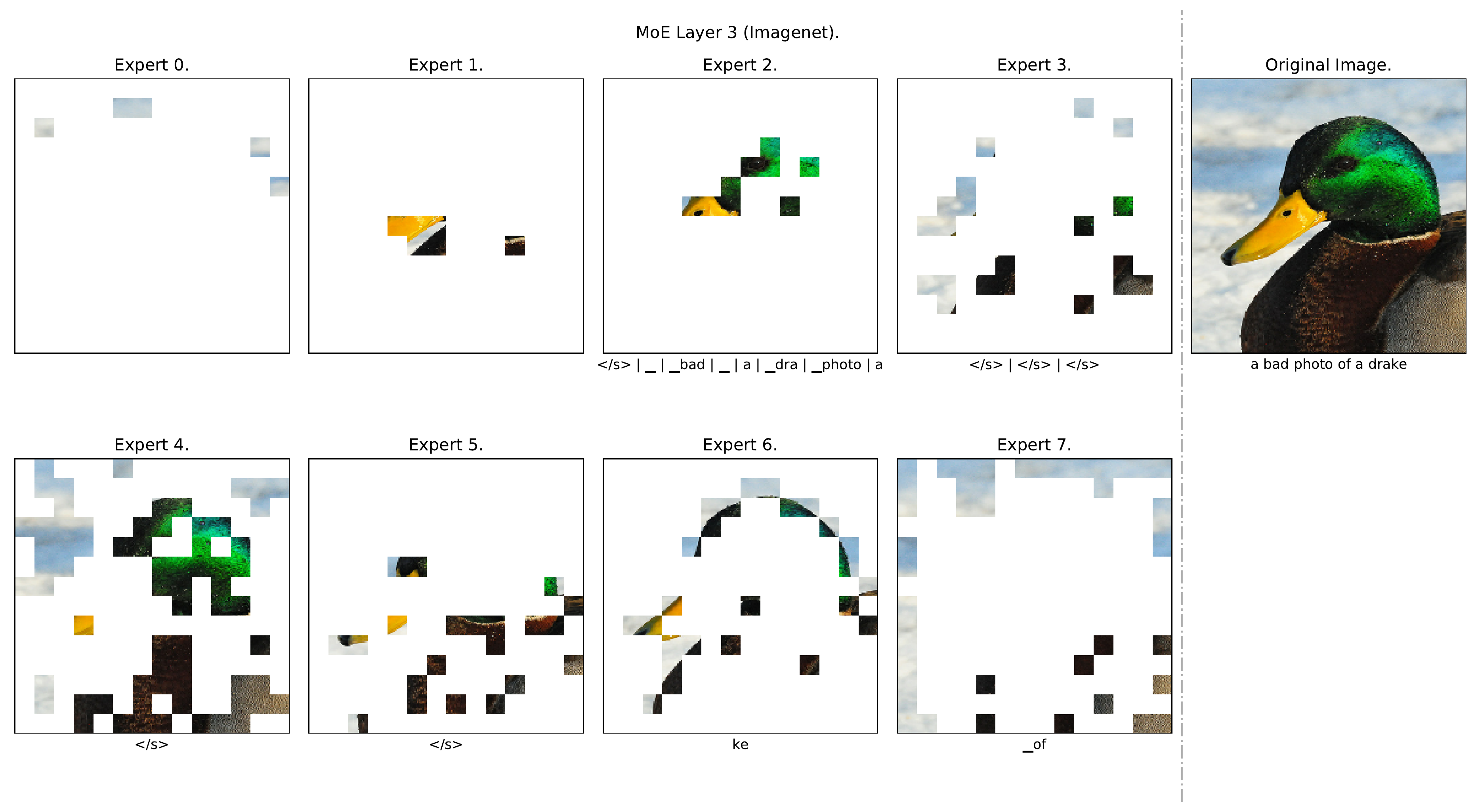}
  \caption{\textbf{Token routing for an Imagenet input.} B/16 model with 8 experts, we show original tokens (both image and text) as routed at the second MoE layer (corresponding to the fourth encoder block, while we use zero-indexing). The original image and text are displayed on the right-hand side.}
  \label{fig:token_routing_imagenet_b16_l3_example_116}
\end{figure}

\begin{figure}[h]
  \centering
  \includegraphics[width=1.0\textwidth, bb=0 0 1072 590]{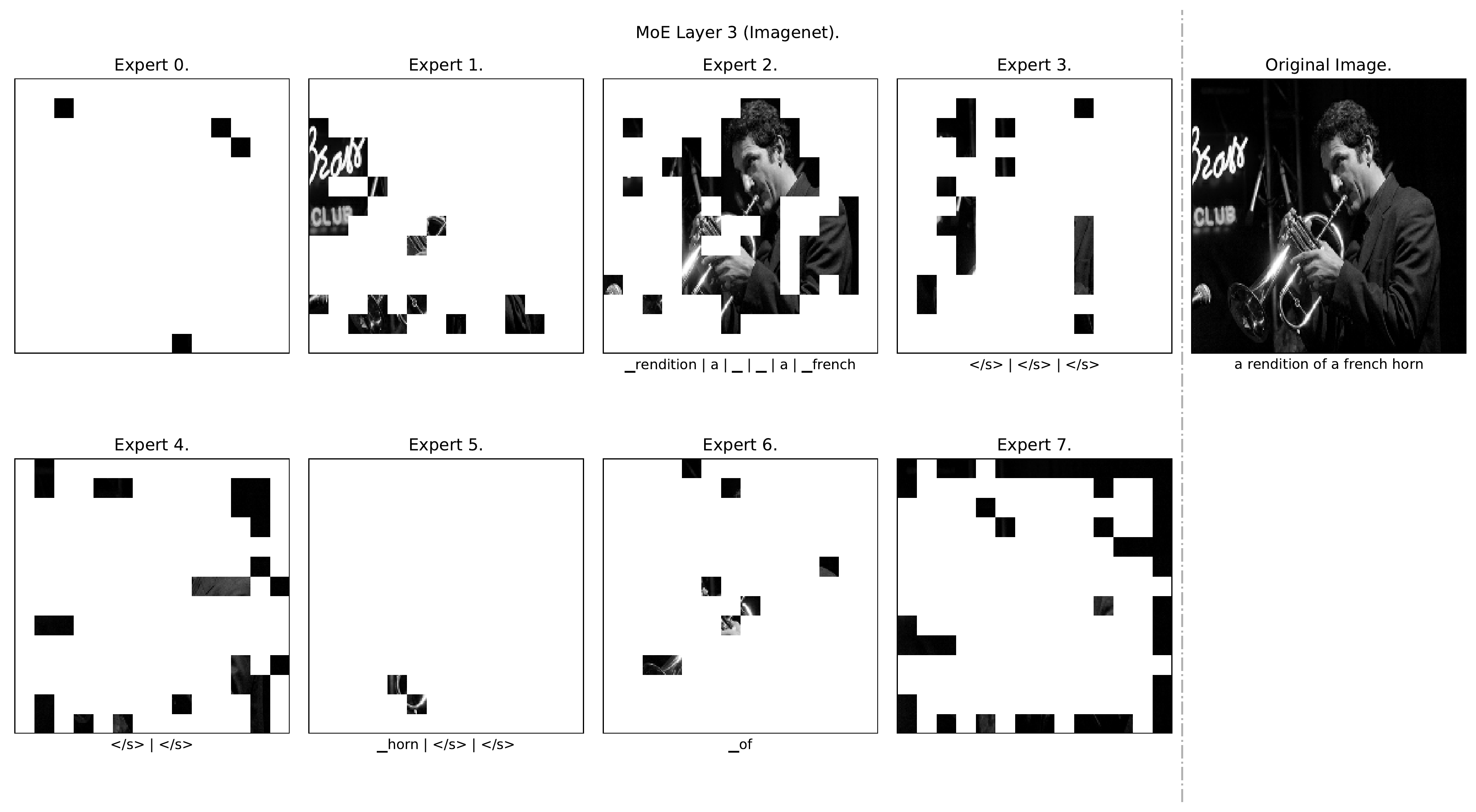}
  \caption{\textbf{Token routing for an Imagenet input.} B/16 model with 8 experts, we show original tokens (both image and text) as routed at the second MoE layer (corresponding to the fourth encoder block, while we use zero-indexing). The original image and text are displayed on the right-hand side.}
  \label{fig:token_routing_imagenet_b16_l3_example_110}
\end{figure}

\begin{figure}[h]
  \centering
  \includegraphics[width=1.0\textwidth, bb=0 0 1072 590]{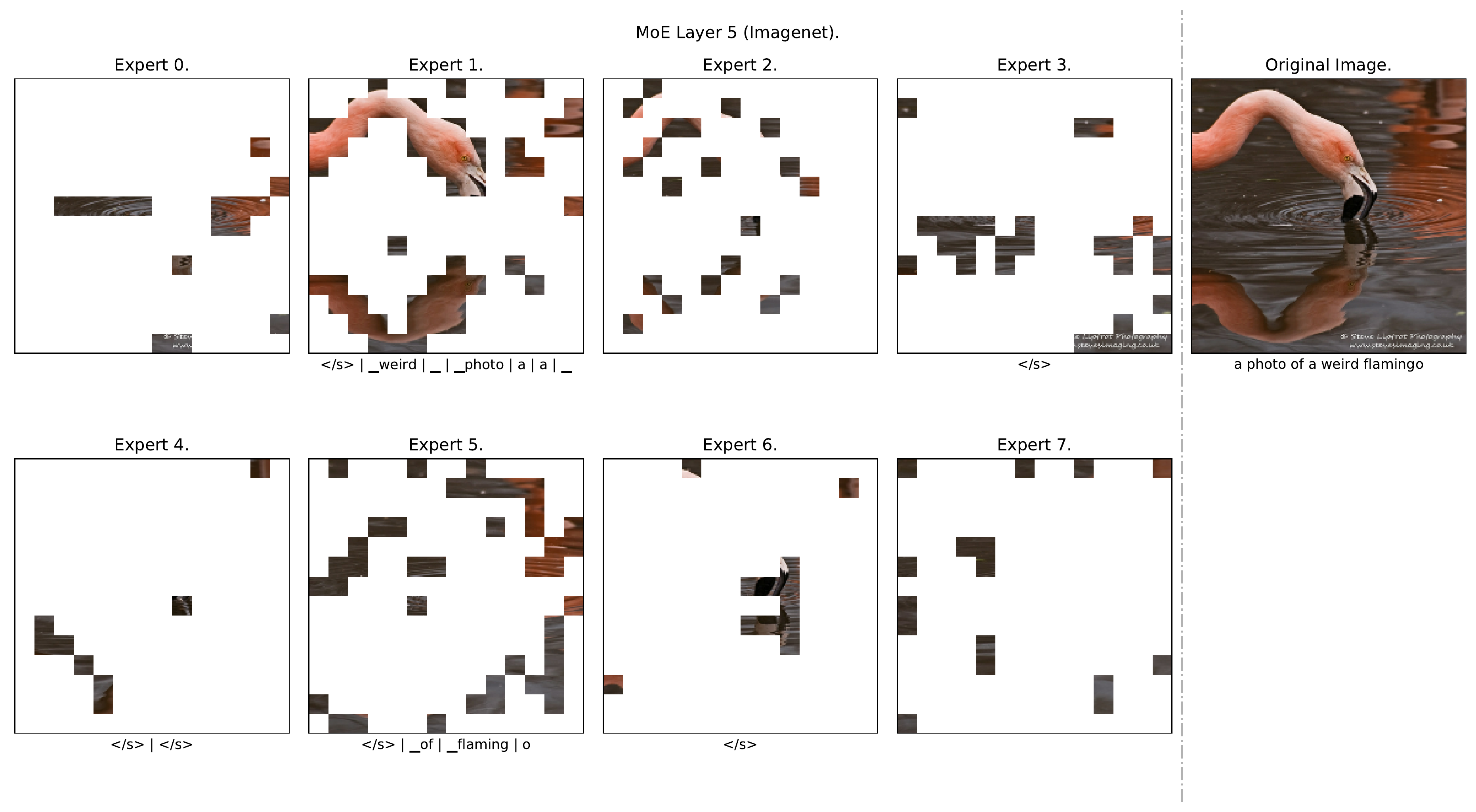}
  \caption{\textbf{Token routing for an Imagenet input.} B/16 model with 8 experts, we show original tokens (both image and text) as routed at the third MoE layer (corresponding to the sixth encoder block). The original image and text are displayed on the right-hand side.}
  \label{fig:token_routing_imagenet_b16_l5_example_446}
\end{figure}

\begin{figure}[h]
  \centering
  \includegraphics[width=1.0\textwidth, bb=0 0 1072 590]{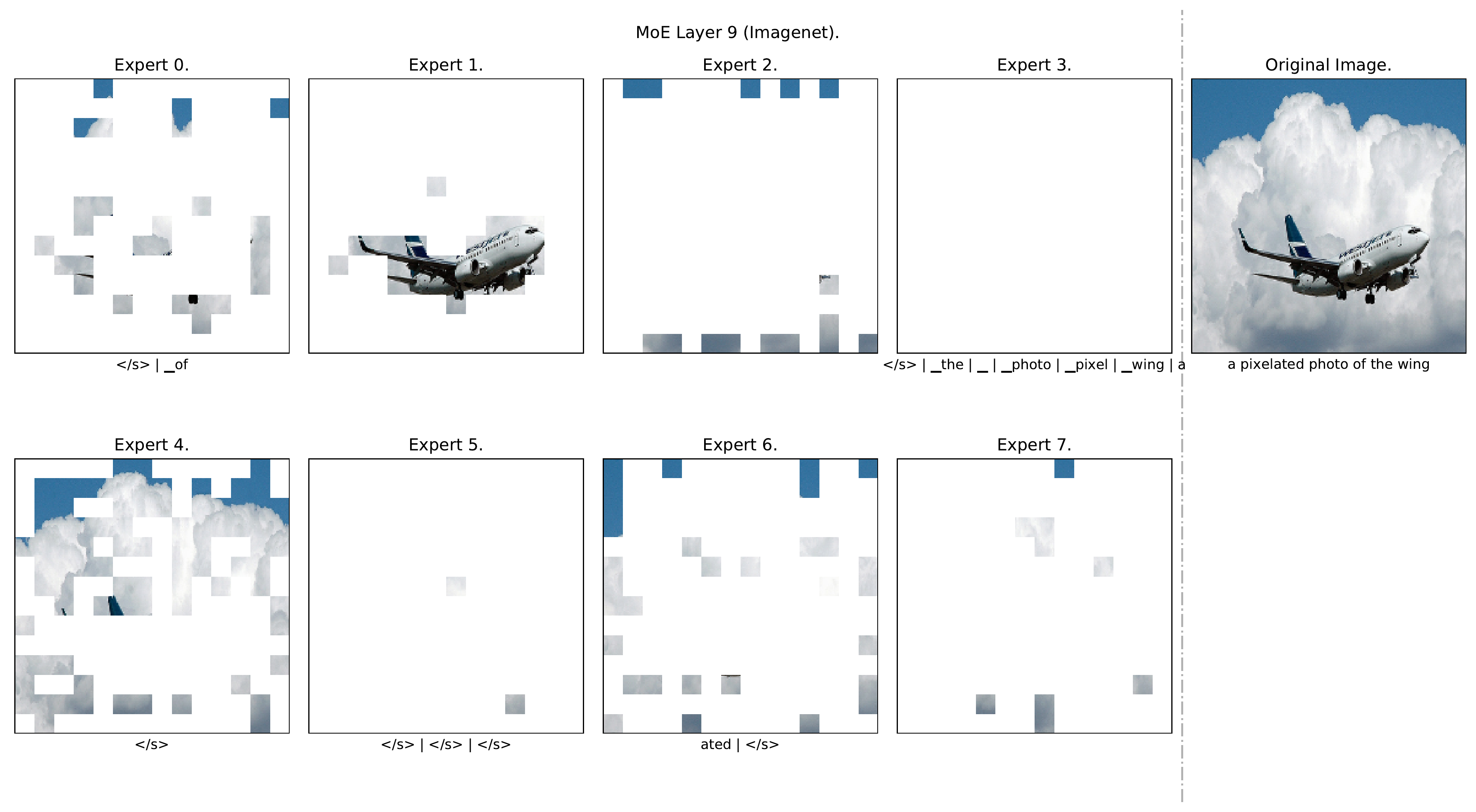}
  \caption{\textbf{Token routing for an Imagenet input.} B/16 model with 8 experts, we show original tokens (both image and text) as routed at the previous-to-last MoE layer (corresponding to the tenth encoder block, while we use zero-indexing). The original image and text are displayed on the right-hand side.}
  \label{fig:token_routing_imagenet_b16_l9_example_29}
\end{figure}

\clearpage
\subsection{Routing Trajectories}
\label{app:analysis_routing_trajectories}
\begin{figure}[h]
  \centering
  \includegraphics[width=0.8\textwidth, bb=0 0 611 512]{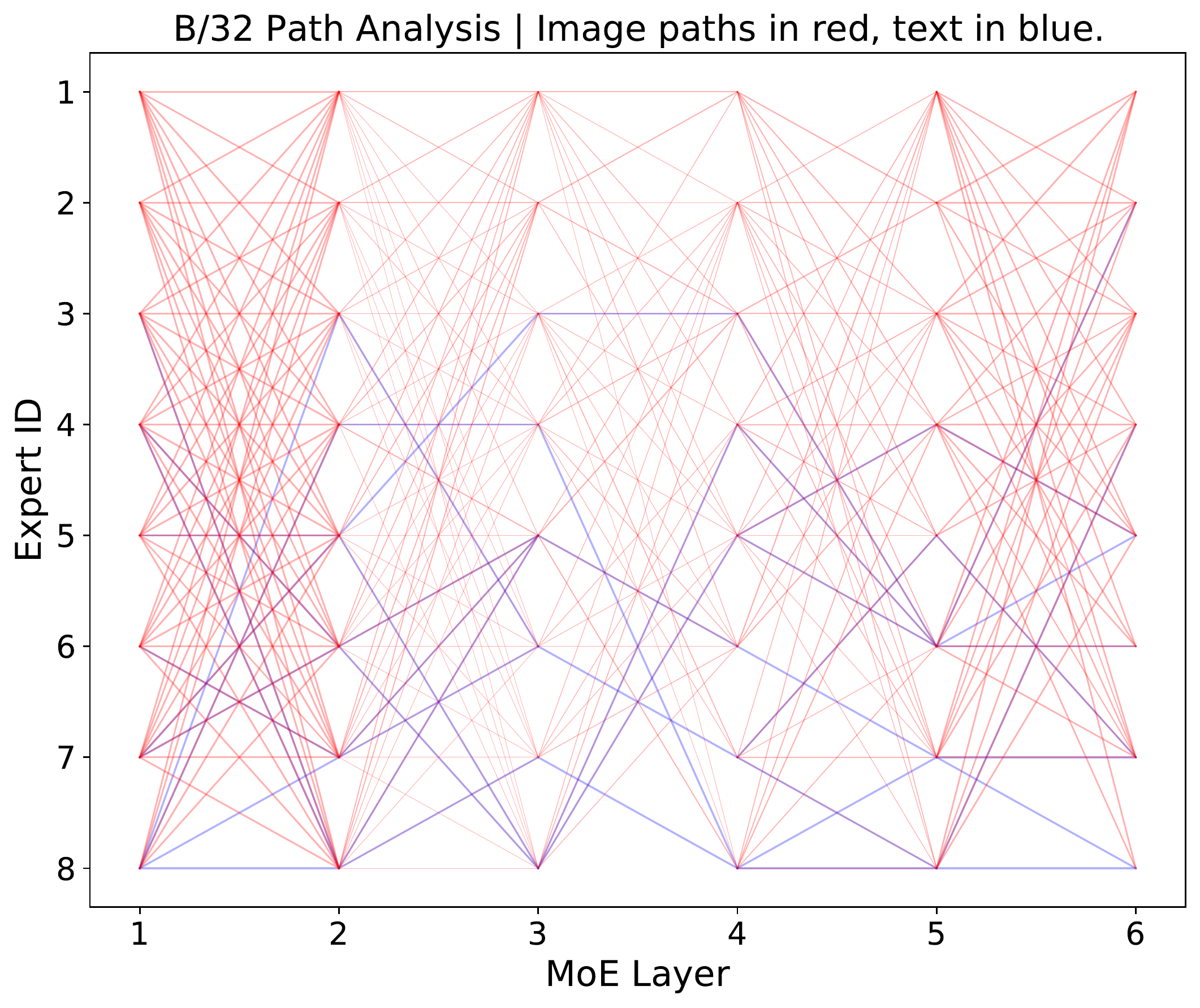}
  \caption{\textbf{Token trajectories.} B/32 model with 8 experts, we show the main expert-routes followed by text tokens (in blue) and image tokens (in red).}
  \label{fig:token_trajectories_b32}
\end{figure}

\begin{figure}[h]
  \centering
  \includegraphics[width=0.8\textwidth, bb=0 0 611 512]{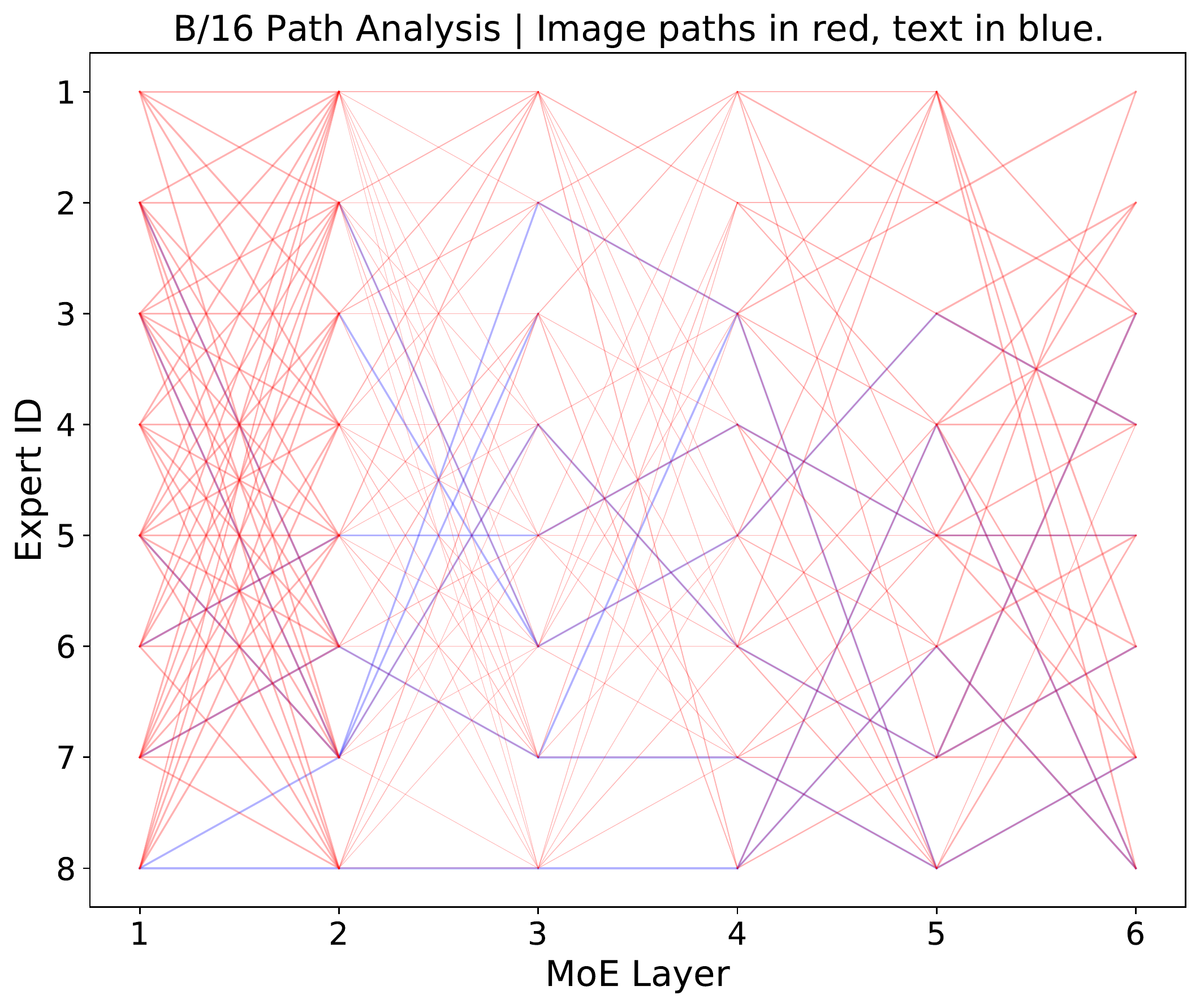}
  \caption{\textbf{Token trajectories.} B/16 model with 8 experts, we show the main expert-routes followed by text tokens (in blue) and image tokens (in red).}
  \label{fig:token_trajectories_b16}
\end{figure}

In this section, we try to have a look at the overall trajectories followed by both image and text tokens across the network.
While definitely a complex endeavor, we show in Figure~\ref{fig:token_trajectories_b32} for B/32 and Figure~\ref{fig:token_trajectories_b16} for B/16 the main trajectories followed by such tokens.
Interestingly enough, it seems that for both models and image tokens, the first two/three MoE layers are fairly interconnected -- in other words, given the expert selected for some token in one layer, it may be hard to predict the next steps.
Text tokens (probably given that very few experts are indeed often used for text) have more predictable trajectories.

\clearpage

\subsection{BPR rankings}
\label{app:analysis_bpr_rankings}
\begin{figure}[h]
  \centering
  \includegraphics[width=0.95\textwidth, bb=0 0 855 577]{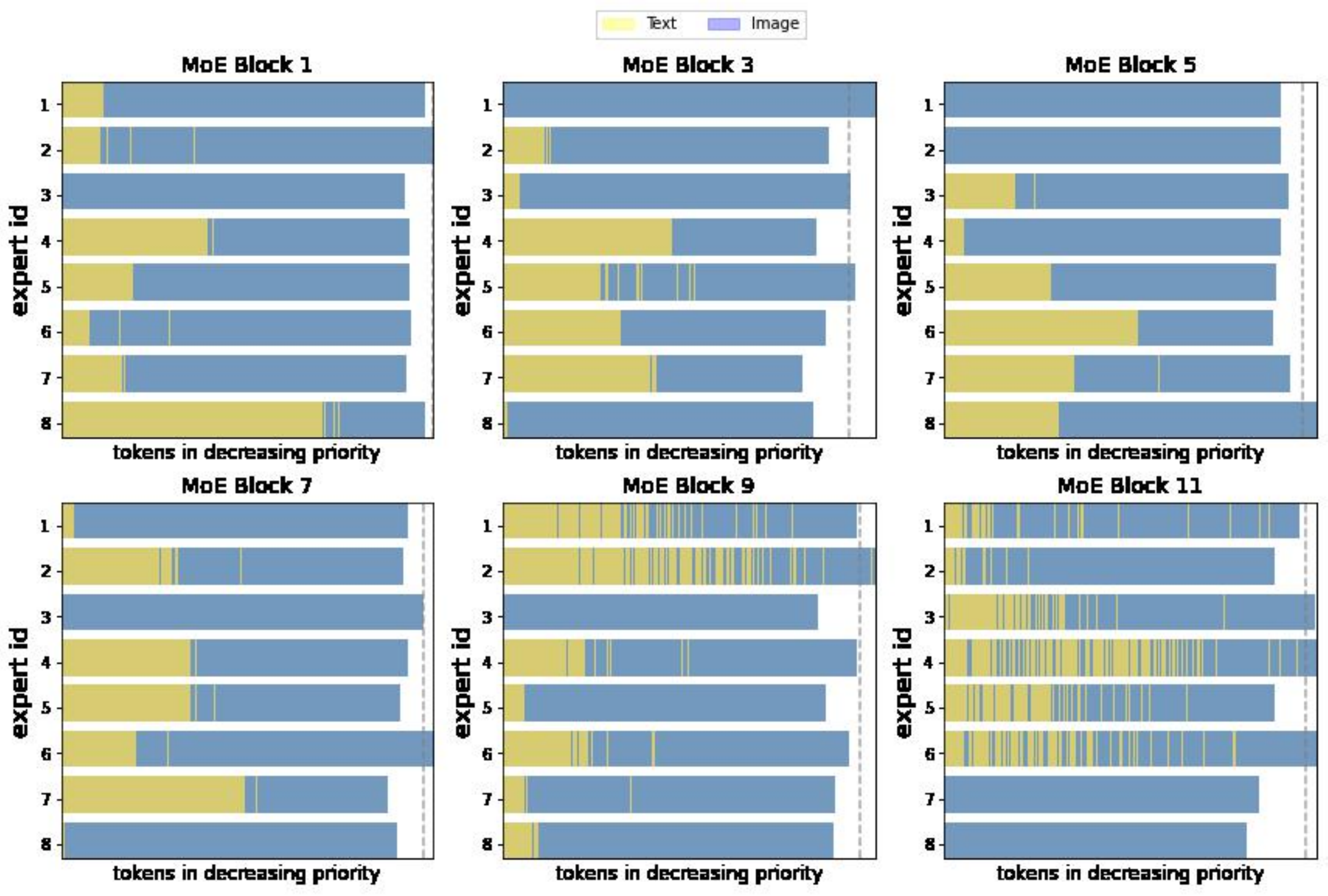}
  \caption{\textbf{Token priorities for training data.} B/32 model with 8 experts. We see that --especially in later layers-- token priorities are mingled across modalities, whereas text tokens tend to have higher scores (and, thus, BPR priorities). Tokens to the left of the $x$-axis are given more priority. The vertical discontinuous line corresponds to the per-expert global capacity limit. Tokens beyond that point are not processed by the expert.}
  \label{fig:token_priority_argus_b32}
\end{figure}

\begin{figure}[h]
  \centering
  \includegraphics[width=0.95\textwidth, bb=0 0 855 577]{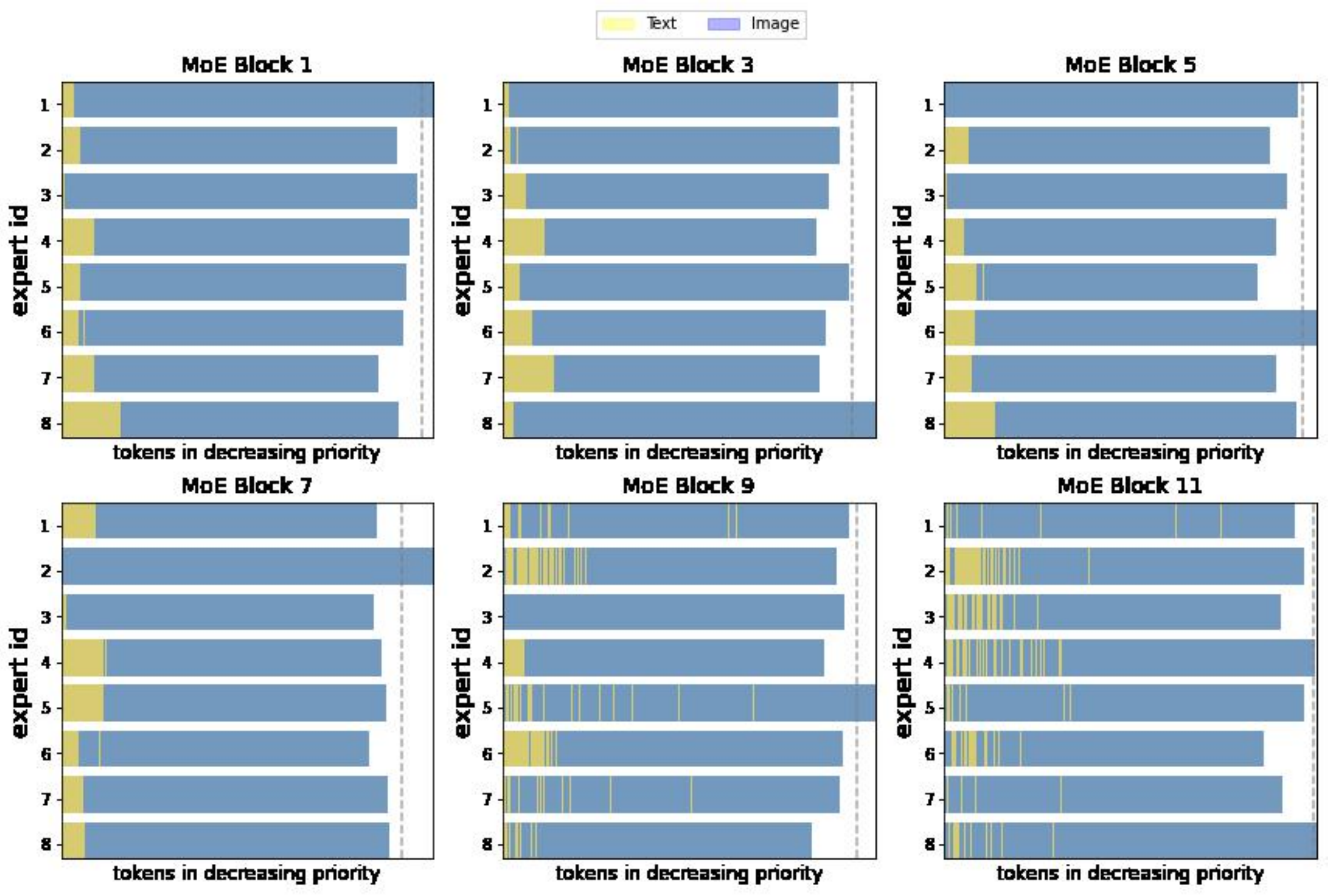}
  \caption{\textbf{Token priorities for training data.} B/16 model with 8 experts. We see that --especially in later layers-- token priorities are mingled across modalities, whereas text tokens tend to have higher scores (and, thus, BPR priorities). Compared to the B/32 model, here we see a longer tail of low-priority image tokens. Tokens to the left of the $x$-axis are given more priority. The vertical discontinuous line corresponds to the per-expert global capacity limit. Tokens beyond that point are not processed by the expert.}
  \label{fig:token_priority_argus_b16}
\end{figure}

\begin{figure}[h]
  \centering
  \includegraphics[width=0.95\textwidth, bb=0 0 855 577]{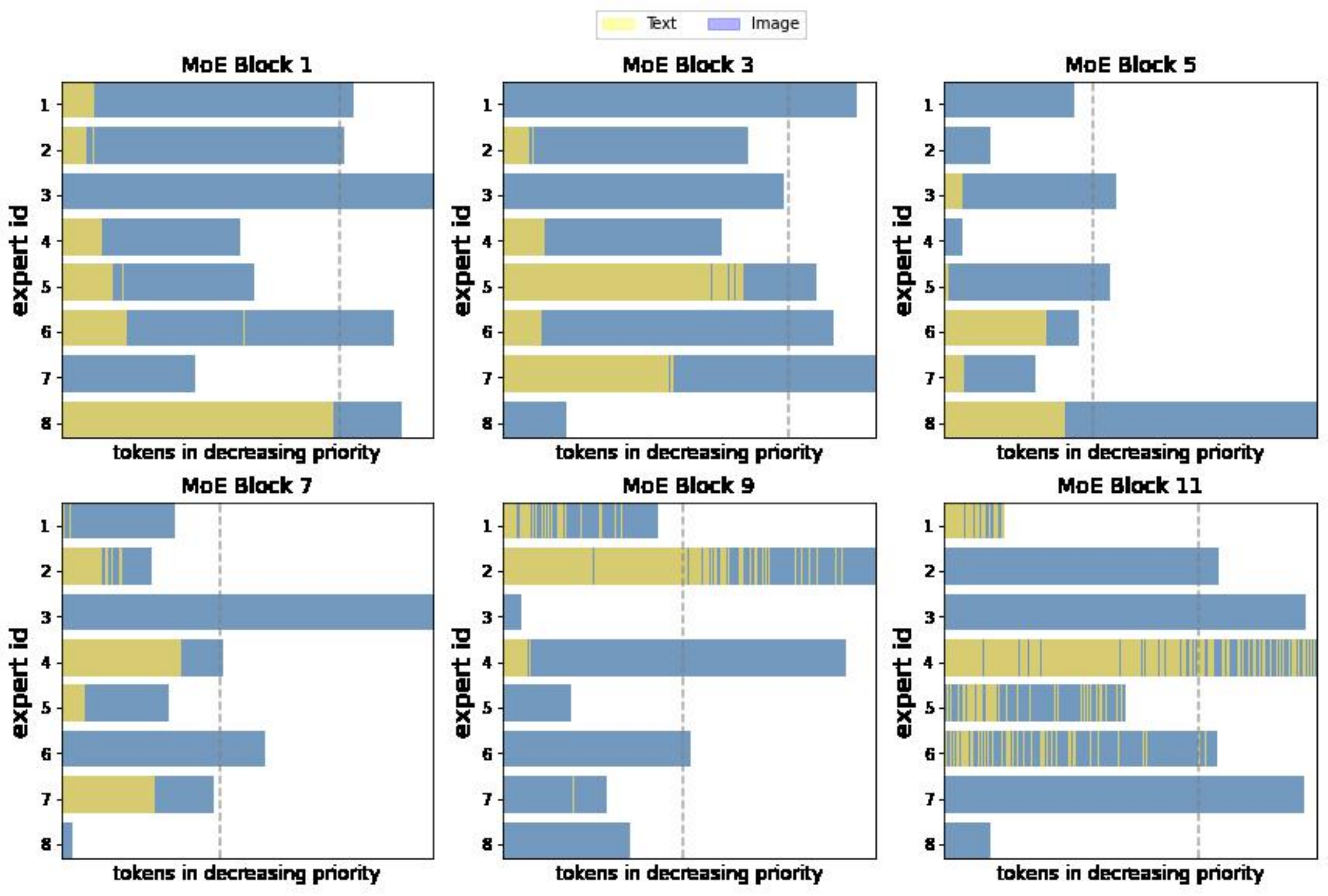}
  \caption{\textbf{Token priorities for COCO data.} B/32 model with 8 experts. We see that --especially in later layers-- token priorities are mingled across modalities, whereas text tokens tend to have higher scores (and, thus, BPR priorities). Tokens to the left of the $x$-axis are given more priority. The vertical discontinuous line corresponds to the per-expert global capacity limit. Tokens beyond that point are not processed by the expert.
  Due to the distribution shift (this is evaluated on COCO, which was not the training data), we see lots of dropping is actually happening (mostly images, but also some text tokens).
  }
  \label{fig:token_priority_coco_b32}
\end{figure}

\begin{figure}[h]
  \centering
  \includegraphics[width=0.95\textwidth, bb=0 0 855 577]{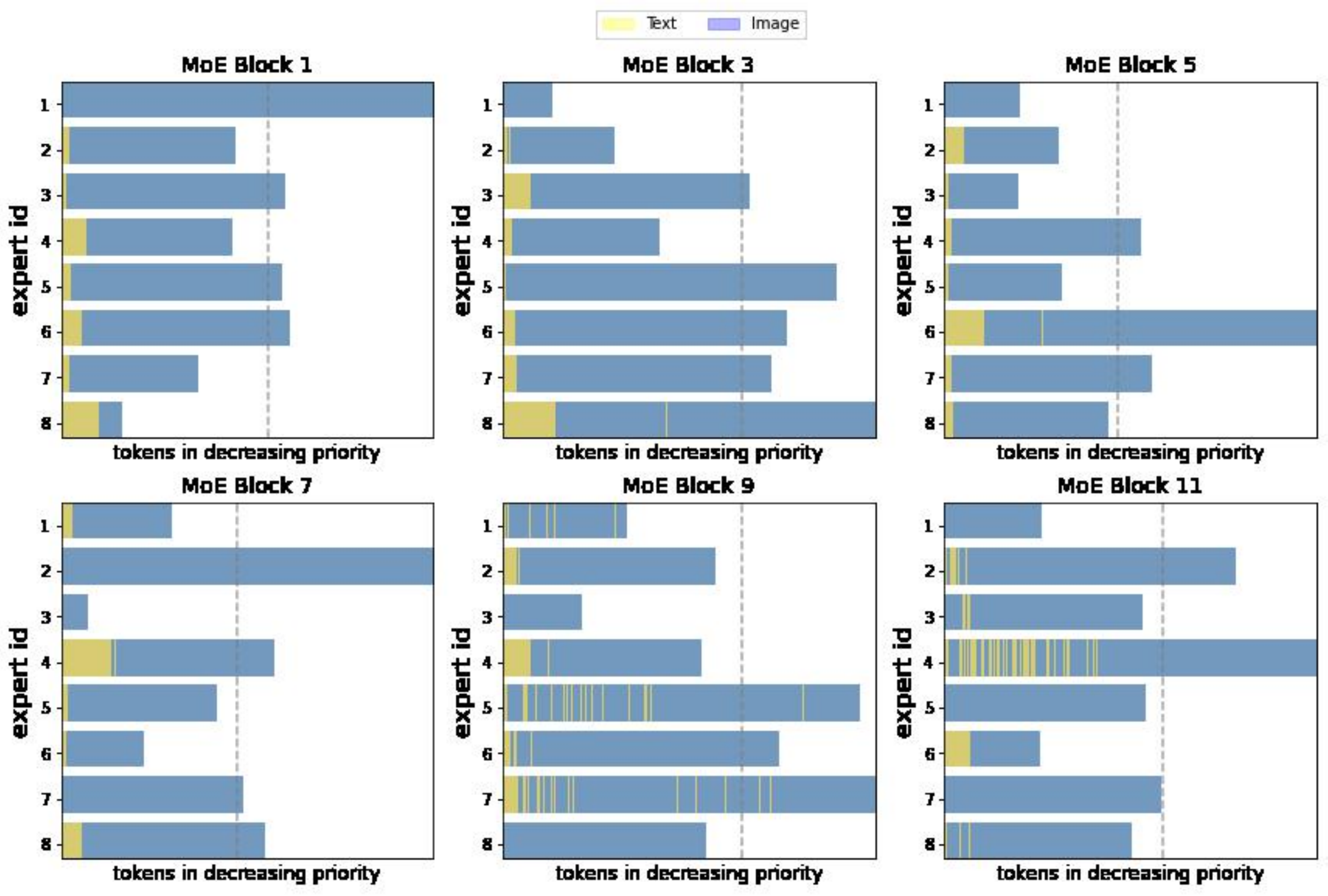}
  \caption{\textbf{Token priorities for COCO data.} B/16 model with 8 experts. We see that --especially in later layers-- token priorities are mingled across modalities, whereas text tokens tend to have higher scores (and, thus, BPR priorities). Tokens to the left of the $x$-axis are given more priority. The vertical discontinuous line corresponds to the per-expert global capacity limit. Tokens beyond that point are not processed by the expert.
  Due to the distribution shift (this is evaluated on COCO, which was not the training data), we see lots of dropping is actually happening (while pretty much only image tokens).
  }
  \label{fig:token_priority_coco_b16}
\end{figure}

\begin{figure}[h]
  \centering
  \includegraphics[width=0.95\textwidth, bb=0 0 855 577]{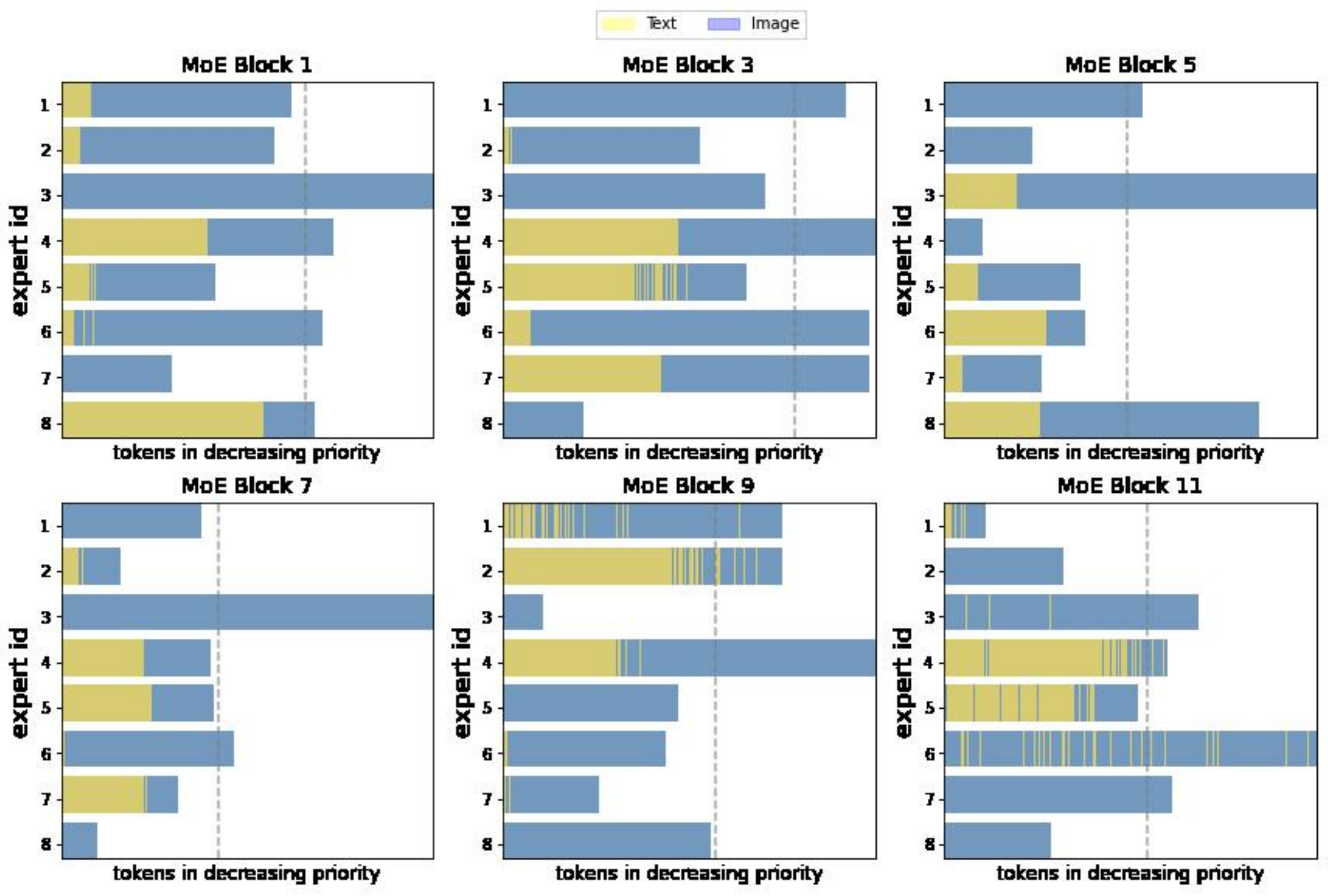}
  \caption{\textbf{Token priorities for ImageNet data.} B/32 model with 8 experts. We see that --especially in later layers-- token priorities are mingled across modalities, whereas text tokens tend to have higher scores (and, thus, BPR priorities). Tokens to the left of the $x$-axis are given more priority. The vertical discontinuous line corresponds to the per-expert global capacity limit. Tokens beyond that point are not processed by the expert.
  Due to the distribution shift (this is evaluated on ImageNet, which was not the training data), we see lots of dropping is actually happening (mostly images, but also some text tokens).
  }
  \label{fig:token_priority_imagenet_b32}
\end{figure}

\begin{figure}[h]
  \centering
  \includegraphics[width=0.95\textwidth, bb=0 0 855 577]{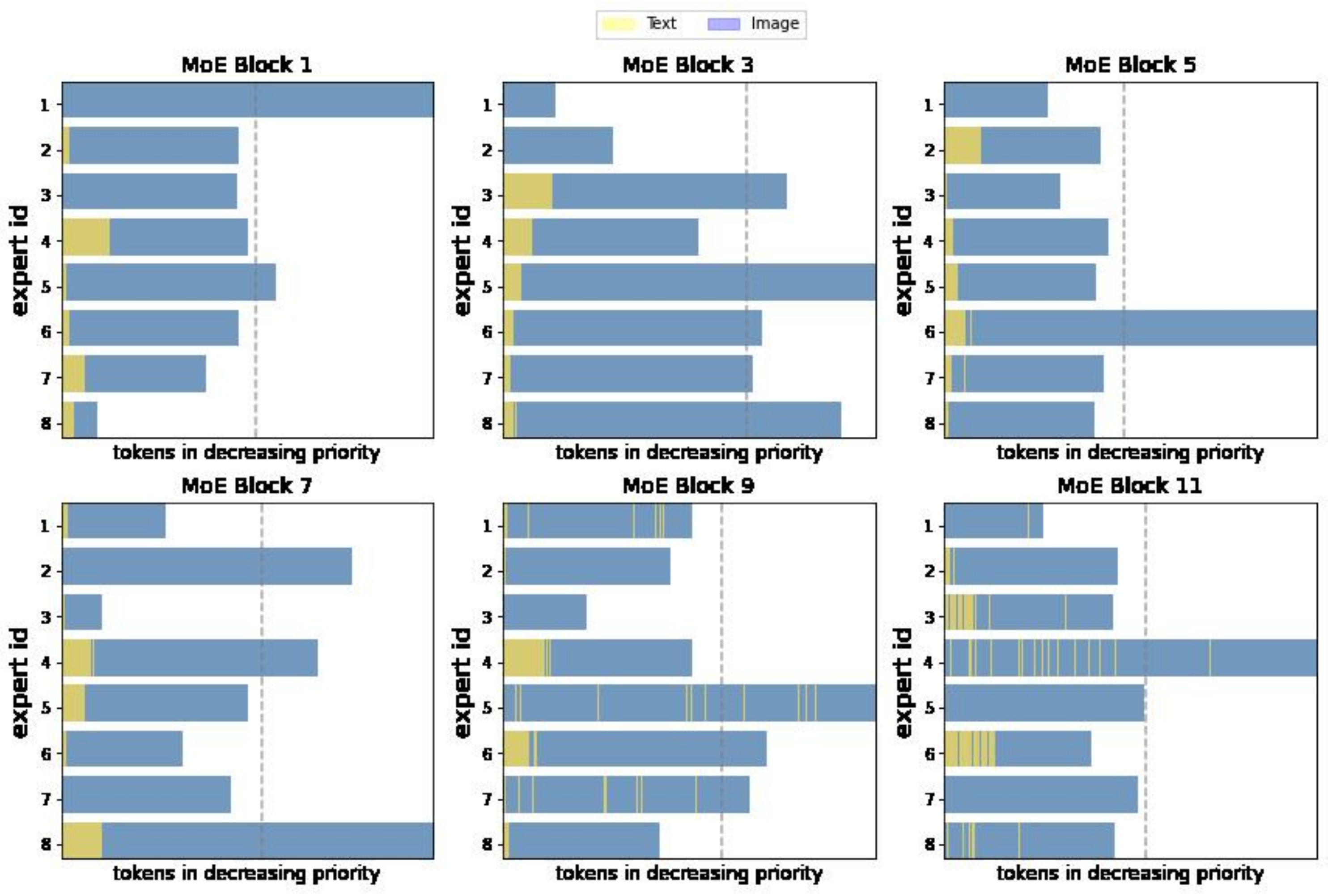}
  \caption{\textbf{Token priorities for ImageNet data.} B/16 model with 8 experts. We see that --especially in later layers-- token priorities are mingled across modalities, whereas text tokens tend to have higher scores (and, thus, BPR priorities). Tokens to the left of the $x$-axis are given more priority. The vertical discontinuous line corresponds to the per-expert global capacity limit. Tokens beyond that point are not processed by the expert.
  Due to the distribution shift (this is evaluated on ImageNet, which was not the training data), we see lots of dropping is actually happening (mostly images, but also some text tokens).
  }
  \label{fig:token_priority_imagenet_b16}
\end{figure}

The local entropy loss encourages concentrated routing predictions with high $p_\text{max}$ for text.
At the same time, BPR prioritises tokens with high $p_\text{max}$. One might assume that this combination is effectively just ranking all text tokens first.
The following plots give us some insight into how the buffers end up sorting tokens from both modalities.
Figures~\ref{fig:token_priority_argus_b32} and ~\ref{fig:token_priority_argus_b16} show the priority distribution on the training data for the B/32 and B/16 models, respectively.
Under a data shift, Figures~\ref{fig:token_priority_coco_b32} and ~\ref{fig:token_priority_coco_b16} show the same statistics for COCO data, and Figures~\ref{fig:token_priority_imagenet_b32} and ~\ref{fig:token_priority_imagenet_b16} for ImageNet.
In these cases, no extra training was performed (i.e., it is zero-shot).
Overall, we see that while text tokens enjoy by default a much higher priority, this is not always the case, and some (important?) image patches are sometimes processed before other text tokens.

%% file: appendix/7_rocket.tex
\section{\mmoe{}-H/14 Analysis}
\label{app:rocket}

\begin{figure}[h]
  \centering
  \includegraphics[width=0.85\textwidth, bb=0 0 711 494]{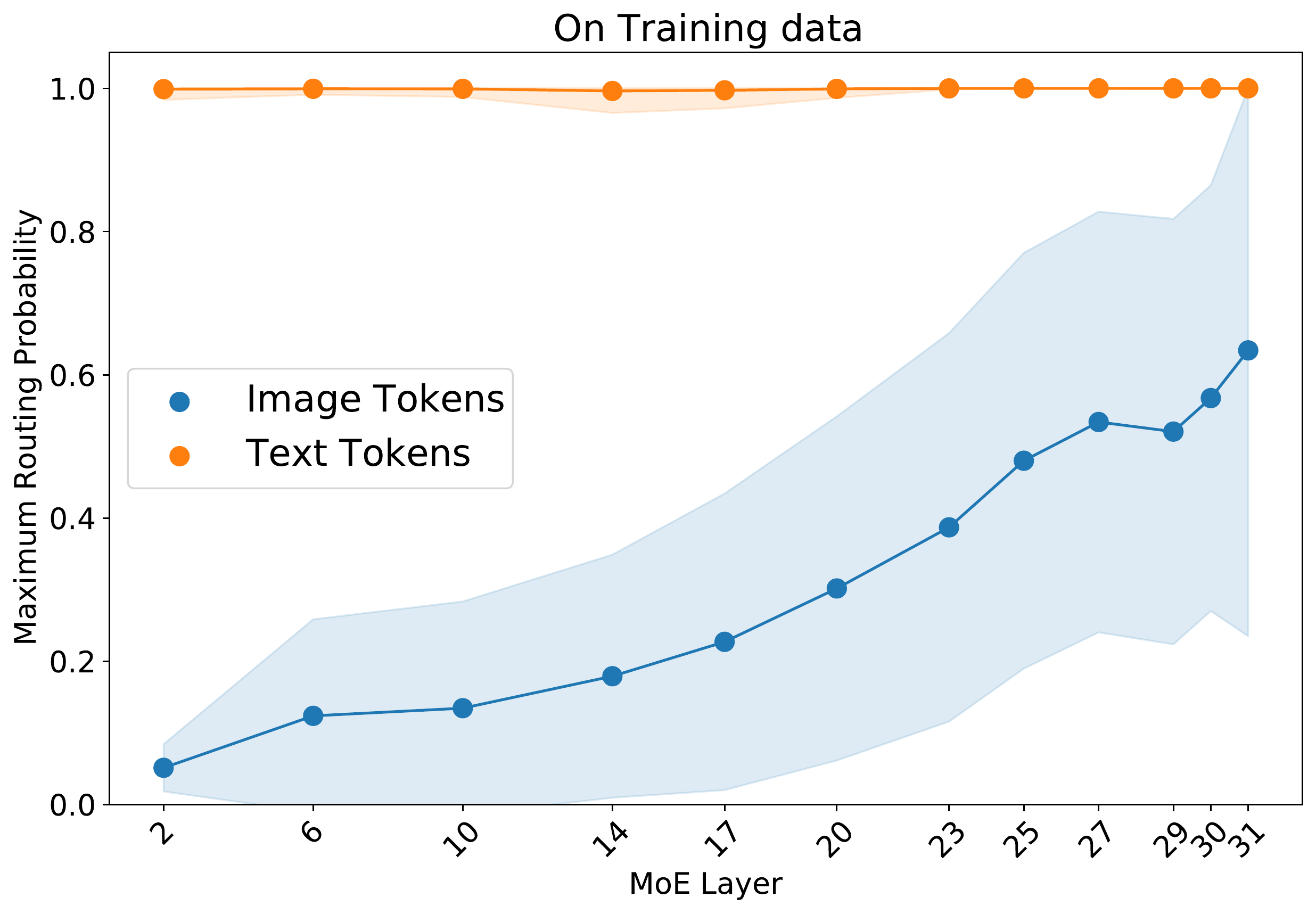}
  \caption{\textbf{Per-token $p_{\max}$ distribution for training data.} For \mmoe{}-H/14 model, we show the average and one standard deviation of the per-token maximum routing probability (corresponding to the selected expert). We see that for image tokens the model is increasingly confident, whereas for text tokens --given the local entropy loss-- most of the predictions are close to one-hot.}
  \label{fig:pmax_h14}
\end{figure}

\begin{figure}[h]
  \centering
  \includegraphics[width=1.0\textwidth, bb=0 0 1431 1071]{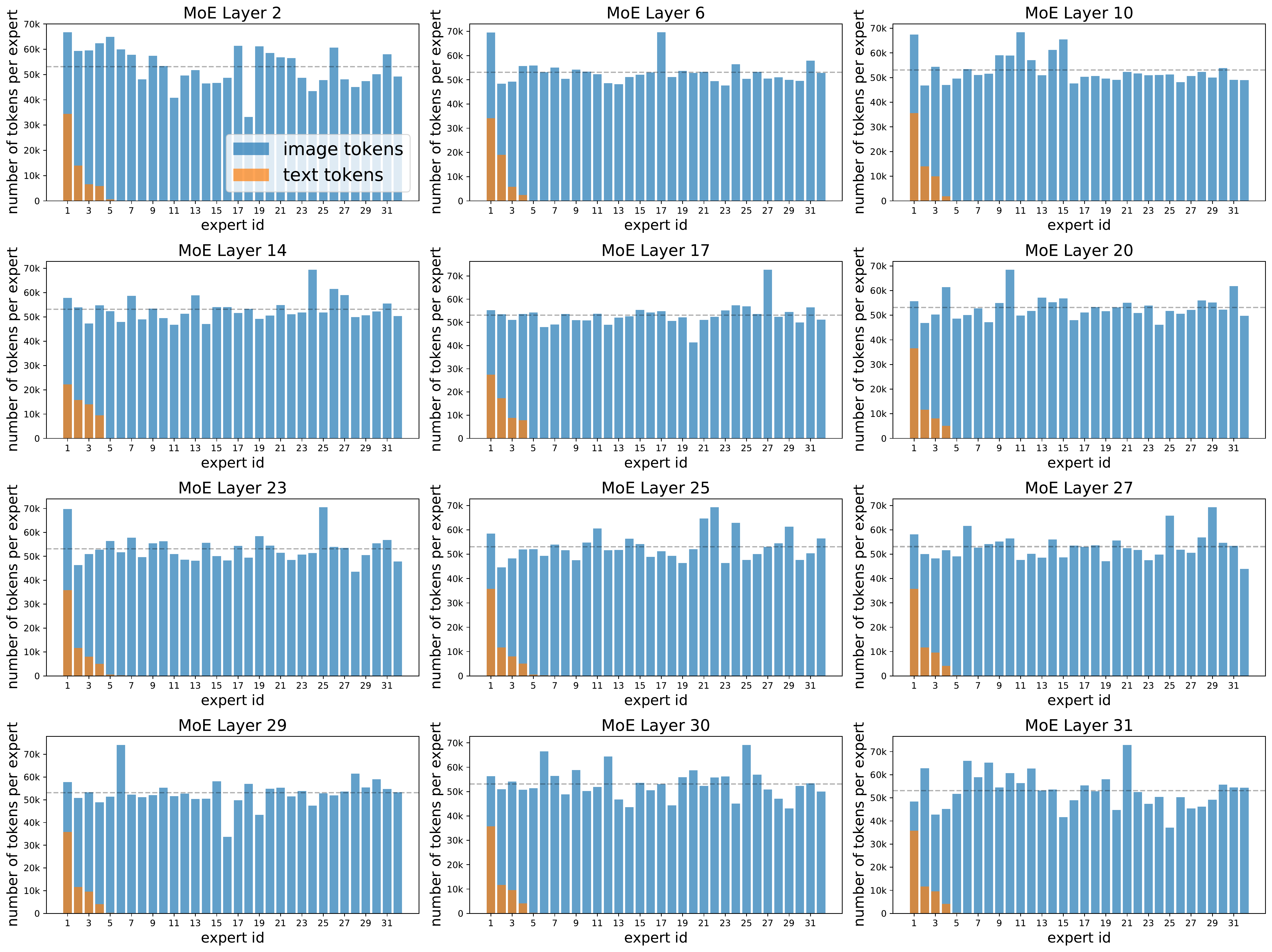}
  \caption{\textbf{Token routing per expert for \mmoe{}-H/14.} We show for each MoE layer and expert, the number of tokens per modality that were routed in a number of forward passes from the training data.
  When above the expert capacity (discontinuous horizontal line), some tokens were dropped -- but not necessarily the image ones; for simplicity, we always show image tokens on top of text ones.
  In this setup, the ratio text:image tokens was close to 1:27.}
  \label{fig:limoe_h14_examples_routing}
\end{figure}

In this section, we share some details and analysis regarding our largest model, the \mmoe{}-H/14.
Figure~\ref{fig:pmax_h14} shows the development of the max routing probability across different MoE layers.
Figure~\ref{fig:limoe_h14_examples_l17_main} shows qualitatively the specialization of image experts. Experts naturally specializing on semantic concepts such as body parts (hands, eyes), textures, fauna, food and doors.
In Figure~\ref{fig:limoe_h14_examples_routing}, we show the distribution of tokens per type and expert for every layer.
Note that we set the entropy loss to approximately require at least 4 text experts, something that seems to agree well with the plot (in this case the ratio text:image tokens was close to 1:27).

\subsection{Preliminary analysis of text routings}
\label{app:text_routing_analysis}
We analyse the routing distributions of text tokens for \mmoe{}-H/14, using NLTK~\cite{nltk} to distinguish between verbs, nouns, adjectives, prepositions and determiners. Note that the SentencePiece tokenizer breaks words into smaller units, which are not necessarily always handled by the same expert, so it is not possible to perfectly parse every token processed by every expert.

\begin{figure}[h]
  \centering
  \includegraphics[width=\textwidth, bb=0 0 1271 263]{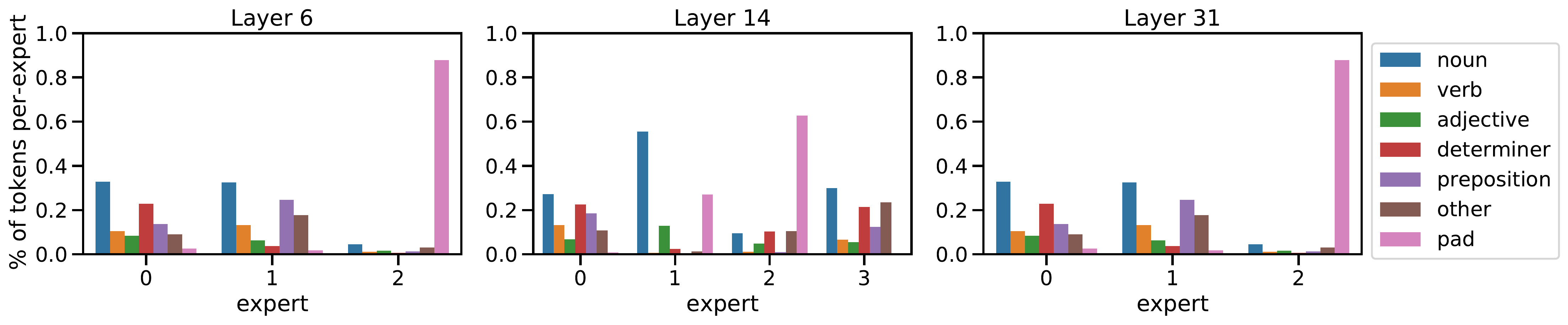}
  \caption{\textbf{Analysis of text routing for \mmoe{}-H/14}
  }
  \label{fig:text_h14}
\end{figure}

The majority of tokens are from images, so only 3-4 experts handle text in this scenario. Figure~\ref{fig:text_h14} contains preliminary analysis, showing for each expert the breakdown of tokens it handles. Though some experts process a bit of everything (e.g. experts 0 and 1 in layer 6 and 31), there are signs of some semantic specialization. There are often experts which process mostly padding tokens. In Layer 14, expert 1 processes no prepositions, determiners or verbs, focussing on nouns and adjectives (and some padding); similarly expert 1 processes very few nouns or adjectives, instead handling padding tokens.

